\documentclass{article} 
\usepackage{iclr2026_conference,times}

\usepackage{amsmath,amsfonts,bm}

\def\eqref#1{equation~\ref{#1}}

\def\1{\bm{1}}

\def\vb{{\bm{b}}}

\def\vw{{\bm{w}}}
\def\vx{{\bm{x}}}

\def\vz{{\bm{z}}}

\def\mA{{\bm{A}}}

\def\mH{{\bm{H}}}
\def\mI{{\bm{I}}}

\def\mM{{\bm{M}}}

\def\mW{{\bm{W}}}
\def\mX{{\bm{X}}}

\DeclareMathAlphabet{\mathsfit}{\encodingdefault}{\sfdefault}{m}{sl}
\SetMathAlphabet{\mathsfit}{bold}{\encodingdefault}{\sfdefault}{bx}{n}

\def\gG{{\mathcal{G}}}

\def\gT{{\mathcal{T}}}

\def\sV{{\mathbb{V}}}

\newcommand{\E}{\mathbb{E}}

\newcommand{\R}{\mathbb{R}}

\usepackage{hyperref}
\usepackage{url}

\usepackage{graphicx}
\usepackage{caption}
\usepackage{subcaption}

\usepackage{booktabs}
\usepackage{placeins}
\usepackage{multirow} 
\usepackage{adjustbox}
\usepackage{enumitem}
\usepackage{comment}
\usepackage[dvipsnames]{xcolor}

\title{\raggedright The Unreasonable Effectiveness of Randomized Representations in Online Continual Graph Learning}

\author{Giovanni Donghi\textsuperscript{1,2} \thanks{Corresponding author email: \textit{giovanni.donghi@phd.unipd.it}}, Daniele Zambon\textsuperscript{2}, Luca Pasa\textsuperscript{1}, Cesare Alippi\textsuperscript{2,3}, Nicolò Navarin\textsuperscript{1} \\
\textsuperscript{1}Università di Padova, \textsuperscript{2}Università della Svizzera italiana, IDSIA, \textsuperscript{3}Politecnico di Milano 
}

\iclrfinalcopy 
\begin{document}

\maketitle

\begin{abstract}
Catastrophic forgetting is one of the main obstacles for Online Continual Graph Learning (OCGL), where nodes arrive one by one, distribution drifts may occur at any time and offline training on task-specific subgraphs is not feasible.
In this work, we explore a surprisingly simple yet highly effective approach for OCGL: we use a fixed, randomly initialized encoder to generate robust and expressive node embeddings by aggregating neighborhood information, training online only a lightweight classifier.
By freezing the encoder, we eliminate drifts of the representation parameters, a key source of forgetting, obtaining embeddings that are both expressive and stable.
When evaluated across several OCGL benchmarks, despite its simplicity and lack of memory buffer, this approach yields consistent gains over state-of-the-art methods, with surprising improvements of up to 30\% and performance often approaching that of the joint offline-training upper bound.
These results suggest that in OCGL, catastrophic forgetting can be minimized without complex replay or regularization by embracing architectural simplicity and stability.
\end{abstract}

\section{Introduction}

In Online Continual Graph Learning (OCGL), nodes arrive sequentially, are observed only once, and undergo drifts in both distribution and task to solve \citep{ocgl}. Therefore, models must adapt to new operating conditions on the fly, while making anytime predictions and retaining past knowledge from limited observations, and under strict memory and latency constraints. This makes OCGL one of the most challenging continual learning scenarios. 
This setting poses several additional requirements to the traditional Continual Learning (CL) \citep{parisi_continual_2019,de_lange_continual_2022,van_de_ven_three_2022}, Online CL \citep{mai_online_2022}, and Continual Graph Learning (CGL) \citep{zhang_cglb_2022}, enabling applications that require fast adaptations and anytime predictions \citep{koh_online_2021} such as in healthcare \citep{LeBaher_2023_PatientElectronicHealth}, IoT intrusion detection \citep{Lin_2024_EGRACLIoTIntrusion} and financial markets \citep{Weber_2019_AntiMoneyLaunderingBitcoin}, other than modeling citation networks \citep{liu_overcoming_2021,zhou_overcoming_2021}, transportation networks \citep{ijcai2021p498} and recommender systems \citep{Caroprese2025}.

In this paper, we introduce a simple, yet surprisingly effective approach for node-level OCGL, decoupling node representations from the predictive model, and yielding excellent prediction accuracy while taming the forgetting of previously learned concepts.
Our approach leverages the effectiveness of randomized and over-parametrized architectures to provide rich, untrained representations, combined with a lightweight trained classifier on top \citep{rahimi2008weighted,rudi2017generalization,Scardapane_2017_RandomnessNeuralNetworks}.
As demonstrated in related literature, fixed representations from pre-trained models help prevent forgetting at the feature extraction level even when used in CL for image classification \citep{Hayes_2020_LifelongMachineLearning,Pelosin_2022_SimplerBetterOfftheshelf,Mehta_2023_EmpiricalInvestigationRole}. The beneficial effect appears to be even more striking in graphs, where stable neighborhood embeddings eliminate the need to retain the entire topological information for replay strategies \citep{Zhang_2024}. Additionally, the literature on randomized models provides us results that ensure good approximation and generalization properties for this family of models, given that we use a sufficiently large embedding dimension \citep{Scardapane_2017_RandomnessNeuralNetworks}.
Coupled with a Streaming Linear Discriminant Analysis (SLDA) classifier \citep{Hayes_2020_LifelongMachineLearning}, this approach generally outperforms state-of-the-art OCGL methods, without need for a memory buffer.

Our main contributions are the following.
\begin{enumerate}[nolistsep,leftmargin=0.6cm]
    \item We propose a surprisingly strong OCGL method. A simple, yet effective method coupling principled untrained features with a lightweight streaming classifier, granting both expressivity and forgetting-resilience.
    \item We demonstrate how the approach achieves generally best results across seven OCGL benchmarks, in both class-incremental and time-incremental setups.
    \item We theoretically motivate our research findings, opening to the development of new, simpler CL methods without compromising on accuracy.
\end{enumerate}

Beyond providing a strong new method with said advantages, our findings suggest that future OCGL research should rethink model designs, emphasizing architectural simplicity and stability.

\section{Related works}\label{sec:related-work}

In this section, we first outline the specifics of Online Continual Learning for graphs, the setting used for our experiments, distinguishing it from related paradigms. We then discuss randomized neural models and other fixed feature extraction strategies in Continual Learning. These concepts provide context for the problem addressed in the paper and motivate our design choices.

\paragraph{Online continual learning for graphs}

Continual Learning has been studied mainly in domains such as computer vision \citep{rebuffi_icarl_2017,lopez-paz_gradient_2017} or reinforcement learning \citep{kirkpatrick_overcoming_2017,rolnick_experience_2019}, with clearly defined task boundaries and no dependencies between successive tasks. In contrast, working on a node level in a graph domain introduces structural dependencies between data points, complicating the definition of the task. Continual Graph Learning (CGL) \citep{febrinanto_graph_2023,yuan_continual_2023,zhang_continual_2024} extends CL to graph settings, and assumes that the graph is presented in blocks, that is, as subsets of nodes connected in subgraphs \citep{zhang_cglb_2022}.
Several methods for CL on graph data have been proposed \citep{zhou_overcoming_2021,liu_overcoming_2021,liu_cat_2023,hoang_universal_2023,Sun_2023_SelfSupervisedContinualGraph,Cui_2023_LifelongEmbeddingLearning}, yet, similarly to traditional CL, state-of-the-art performance is achieved by replay-based methods which leverage a memory buffer tailored to leverage graph topology \citep{zhang_sparsified_2022,Zhang_2024}.
However, the use of Graph Neural Network (GNN) \citep{scarselli_graph_2009,micheli_neural_2009,kipf2017semisupervised} models poses an issue regarding task separation: nodes form connections also with nodes that were observed in past tasks, thus requiring access to past information due to message passing \citep{gilmer_neural_2017}. This is sometimes addressed by ignoring inter-task edges \citep{zhang_cglb_2022}, yet it is an unrealistic solution since it assumes supervisory information in the form of task identifiers also at test time. On the other hand, due to small-world nature of most real-world graphs, accessing the entire neighborhood information of the nodes would likely mean using the entirety of past data, rendering the use of the CL approach less meaningful. 
In this context, a more principled setting is Online Continual Graph Learning (OCGL) \citep{ocgl}, which bridges the gap between existing research on online CL and CGL. This setting addresses the issue of access to node neighborhoods with the use of neighborhood sampling, to keep a constant computational footprint even as the graph gets denser over time.
In particular, the online setting for CL entails a single pass over the streaming data \citep{chaudhry_efficient_2018,mai_online_2022,soutifcormerais_comprehensive_2023}, in contrast to traditional offline CL where training is done offline in a batched setting on each task \citep{de_lange_continual_2022}. This constraint, together with stricter computational and memory requirements, is motivated by applications where quick model adaptation is necessary, and anytime predictions may be required \citep{koh_online_2021}, thus impeding the task-wise offline training. 

\paragraph{Randomized representations}

In the literature on neural networks, multiple approaches have leveraged some form of randomization, due to training efficiency, and often theoretical guarantees \citep{Scardapane_2017_RandomnessNeuralNetworks}.
Notable examples for vector data are Random Vector Functional-Link networks (RVFL) \citep{Pao_1992,PAO1994163}, Extreme Learning Machines (ELM) \citep{Huang_2006_ExtremeLearningMachine,Huang_2015_TrendsExtremeLearning} and Echo State Networks (ESN) \citep{Jaeger2004}. These fix most network parameters randomly, from an appropriate distribution, training only lightweight readouts. Theoretical results such as universal approximation properties and generalization guarantees have been proved for some these architectures \citep{LIU201258}.
Randomized strategies have also been explored for kernel approximation, such as with Random Fourier Features (RFF) \citep{NIPS2007_013a006f}, a randomized projection that approximates the RBF kernel.
Some works have also explored randomized strategies for graph data, such as with GESN \citep{Gallicchio_2010_GraphEchoState} or MRGNN \citep{pasa2022}. Different instantiations of untrained Graph Convolutional Networks \citep{kipf2017semisupervised} have been proposed in recent years: GCELM \citep{Zhang_2020_GraphConvolutionalExtreme} uses a single GCN layer, with randomly initialized and fixed parameters. GCN-RW \citep{Huang_2023} improves on this approach by considering an increased receptive field, employing the square of the adjacency matrix for aggregation. The more recent UGCN \citep{navarin2023} instead uses multiple GCN layers, followed by non-differentiable pooling, as the network does not need to be trained. Inspired by RFF, GRNF \citep{Zambon_2020_GraphRandomNeural} are derived from expressive GNN architectures, and constitute a family that can separate graphs.

\paragraph{Fixed feature extraction for CL}

Some works in the CL literature for image classification suggest the use of a frozen CNN backbone, pre-trained on a different dataset, and then training continually only a classifier such a simple MLP \citep{van_de_ven_three_2022,Mehta_2023_EmpiricalInvestigationRole}. In the context of graph data, this approach is not currently feasible, as it is not trivial to obtain pre-trained GNNs that can handle graphs with different input feature domains, even though there is ongoing promising work on graph foundation models \citep{Wang_2025_GraphFoundationModels}. Additionally, despite the many randomized approaches for representation extraction, only a very limited number of CL papers rely on them, and only with image classification tasks. CRNet \citep{Li_2023_CRNetFastContinual} uses a randomly initialized network for feature extraction, training the classification head via closed-form solution on each task. RanPAC \citep{McDonnell_2023_RanPACRandomProjections} uses random projections of feature extracted with a pre-trained model with the objective of increasing dimensionality to facilitate class separation. More recently, RanDumb \citep{Prabhu_2025_RanDumbRandomRepresentations} leverages Random Fourier Features \citep{NIPS2007_013a006f} used with nearest class mean classification. On the other hand, there have been some works that investigated theoretically the use of linear models in simple CL setups \citep{Evron_2022,Evron_2023_ContinualLearningLinear,Ding_2024_UnderstandingForgettingContinual}, leading us to adopt these simple classifiers. Furthermore, it has been observed that prototype based classifiers, such as nearest class mean, are particularly suited for online CL, due to being schedule-robust, i.e. independent from stream ordering \citep{Wang_2022_ScheduleRobustOnlineContinual}.

\section{Method}\label{sec:method}

The OCGL setting \citep{ocgl} considers an incremental graph $\gG$, induced by a stream of nodes $v_1, v_2, \ldots, v_t, \ldots$ that are added one by one. At each timestep $t$, the graph is updated with neighborhood and attribute information about the incoming node $(v_t, \mathcal{N}(v_t), \vx^t)$, obtaining an updated graph snapshot $\smash{\gG^{(t)} = (\sV^{(t)}, \E^{(t)}, \mX^{(t)})}$. For example, in the Elliptic dataset \citep{Weber_2019_AntiMoneyLaunderingBitcoin}, when a new transaction is processed, its information is captured as node features, and payment flows indicate connections.
The goal is to learn a model $F_{\Theta}$ to make node-level predictions $\smash{\hat{y}_{v}^{(t)}}$ for given node $v$ at time step $t$ given information coming from its neighboring nodes in $\mathcal G^{(t)}$. Specifically, models are trained with a single pass over the node stream, using information from the associated $l$-hop ego-graph $\smash{\gG^{(t,l)}_{v}}$. 
The online setting requests bounded memory and computational cost at each time steps, even as the graph grows. Accordingly, 
only a subset $\smash{\widetilde\gG^{(t,l)}_{v}}$ of the ego-graph may be allowed to produce prediction 
\begin{equation}
    \hat{y}_{v}^{(t)} = F_{\Theta}\left(\widetilde\gG^{(t,l)}_{v}\right) \,. \label{eq:predictions}
\end{equation}
In the context of CL, the node stream can encompass diverse drift patterns, such as those triggered by a class-incremental setting.
Nonetheless, model predictions may be requested for nodes observed in the past, therefore requesting the model not only to adapt to evolving graph conditions but also to preserve previously learned knowledge. 

\subsection{Proposed solution}

To address forgetting in the OCGL setting, we propose a decoupled approach to node prediction. In this setup, model $F_\Theta = \Phi \circ \Psi$
decomposes as a node feature extractor $\Psi$ followed by a node-level predictor $\Phi$.
Feature extraction is performed by a fixed (randomly initialized) backbone $\Psi$:
\begin{equation}
    \vz_v^{(t)} = \Psi\left(\widetilde\gG^{(t,l)}_{v}\right) \,, \label{eq:extractor}
\end{equation}
where $\smash{\vz_v^{(t)}} \in \R^d$ denotes the node embedding that captures neighborhood information. 
As $\Psi$ is untrained, it does not change over time and is inherently immune to forgetting. Secondly, it offers a clear advantage for experience replay methods, as storing $\vz_v^{(t)}$ is significantly more memory-efficient than storing the entire sampled neighborhood $\smash{\widetilde\gG^{(t,l)}_{v}}$.

Feature extractor $\Psi$ also mitigates other sources of forgetting. Unlike trained models, which tend to produce task-specific representations that remove task-irrelevant information, a well-chosen $\Psi$ can produce rich node embeddings suitable for multiple tasks. Notably, theoretical results \citep{NIPS2017_61b1fb3f} show that for some families of model architectures, a random initialization of $\Psi$ yields linearly separable embeddings, allowing to effectively solve the downstream task by simple linear predictive models
\begin{equation}
    \hat{y}_{v}^{(t)} = \Phi_{\mW,\vb}\left(\vz_v^{(t)}\right) = \mW \vz_v^{(t)} + \vb  \,, \label{eq:linear}
\end{equation}
which are less prone to forgetting than deep networks \citep{Mirzadeh_2022_WideNeuralNetworks} and can be learned efficiently.

\subsection{Randomized feature extraction}

We consider two different types of graph feature extractors $\Psi$, or backbones, for our experiments, untrained GCN \citep{navarin2023} and Graph Random Neural Features \citep{Zambon_2020_GraphRandomNeural} adapted to the node-level, which we discuss here. In general, any randomized graph network with sufficient expressivity and robustness to limited structural shifts can be adopted for our decoupled approach.

\paragraph{UGCN} 
As first feature extractor, wee use the recent UGCN \citep{navarin2023}, which uses multiple GCN layers $\mH^{(i)} = tanh(\tilde{\mA} \mH^{(i-1)} \Theta)$, where $\tilde{\mA}$ is the normalized adjacency matrix, with $\mH^{(0)}=\mX$ and with weights initialized with a Glorot uniform approach \citep{pmlr-v9-glorot10a} regulated by a gain hyperparameter, and successively left untrained. 
In our case, since we adopt it for node-level predictions instead of graph-level ones, we do not use the entire adjacency matrix, but the efficient forward propagation limited to the node ego-graph of \citep{hamilton_inductive_2017}.
The output of the different layers is then concatenated to obtain an embedding that reflects information at multiple resolutions, i.e. 
\begin{equation}
    \Psi_{\text{UGCN}} \left(\widetilde\gG^{(t,l)}_{v}\right) = \text{concat}\left(\mH^{(1)}_v, \ldots, \mH^{(l)}_v\right) \,.
\end{equation}
In an over-parameterized regime, this model performs competitively with trained GCN counterparts on graph classification \citep{Navarin_2023_EmpiricalStudyOverParameterized,Donghi_2024_InvestigatingOverparameterizedRandomized}.

\paragraph{GRNF} Graph Random Neural Features \citep{Zambon_2020_GraphRandomNeural}
define expressive embeddings for attributed graphs, derived from a GNN model which is a universal approximator of graph functions \citep{Maron_2019,Keriven_2019}. Each GRNF is a parametric maps $\psi(\cdot, \vw): \gT^2 \to \R$ defined by the composition of equivariant and invariant affine maps, interleaved with non-linearity:
\begin{equation}
    \psi_{\omega}(\cdot) = \sigma \circ H_k(\cdot, \theta_{H}) \circ \sigma \circ F_{2,k}(\cdot, \theta_F) \,,
\end{equation}
where $\omega = (k, \theta_F, \theta_H)$, and $k$ represent tensor order (formal definitions of $F_{2,k}$ and $H_k$, and more details can be found in \citep{Zambon_2020_GraphRandomNeural}). Interestingly, the family of GRNF can separate graphs, meaning that for any non-isomorphic graphs $\gG_1, \gG_2$, there exists $\omega$ such that $\psi_{\omega}(\gG_1) \neq \psi_{\omega}(\gG_2)$. Thus, under suitable assumptions on the distribution of $\vw$, the family of GRNF induces a metric distance on graphs, with also results that hold in probability for a fixed number of features \citep{Zambon_2020_GraphRandomNeural}. To adapt GRNF to make node level predictions, we use them on $\smash{\widetilde\gG^{(t,l)}_{v}}$, and since the the output is permutation invariant over the entire ego-graph, we concatenate it to the component relative to $v$ of the output of the equivariant map, using $d/2$ independent draws of $\omega$, resulting in a $d$-dimensional embedding (assuming even $d$).

\subsection{Linear classifier}

As indicated in \eqref{eq:linear}, the 
extracted features are fed to a linear layer, which can be trained continually on the stream. Effectively, thanks to this strategy we have simplified the Continual Graph Learning problem by eliminating the graph specificity, relegating it to the fixed feature extractor, and reducing the problem to Continual Learning on vectors. A linear layer may thus be trained with gradient descent, supported by any standard CL learning strategy, such as experience replay which can now leverage informative node embeddings. 

Furthermore, we can also address forgetting in the linear classifier by adopting a specific form which does not require gradients: we use Streaming Linear Discriminant Analysis (SLDA) \citep{Hayes_2020_LifelongMachineLearning}, which learns by accumulating class means, independently for each class.
This choice makes training more efficient and provides further robustness against catastrophic forgetting risk, as further discussed in Section~\ref{sec:theory}. 
Specifically, during training, SLDA keeps a cumulative mean $\mu_y$ for the features of each class $y$, and a shared covariance matrix $\Sigma$ with streaming updates. To make predictions, using precision matrix $\Lambda = [(1-\epsilon)\Sigma + \epsilon\mI]^{-1}$ with $\epsilon=10^{-4}$, the parameters $\vw_y$ (rows of $\mW$) and $\vb$ of \eqref{eq:linear} are computed as  
\begin{equation}
    \vw_y = \Lambda \mu_y \,, \qquad
    \vb_y = - \frac{1}{2}(\mu_y^T \Lambda \mu_y) \,.
\end{equation}

\subsection{Effectiveness of the approach}\label{sec:theory}

As the graph evolves, multiple sources of forgetting can impair the model's prediction accuracy. This section elaborates on our method's effectiveness in mitigating forgetting while yielding high anytime classification accuracy.

One is a common challenge in all CL setups and stems from the continual update of the model's parameters based on recent training signals, both for the feature extraction backbone (if trained) and for the classifier. 
A second, graph-specific issue arises from the evolving nature of the graphs and associated structural shifts \citep{Su_2023_RobustGraphIncremental,Su_2024_LimitationExperienceReplay}: predictions made for the same node at different time steps may rely on different and potentially inconsistent neighborhoods.
To put our intuitions more formally, we consider time interval $\delta t$ and define the forgetting risk for node $v$ at time $t$ as the increase in the loss from time $t_0=t-\delta t$ to time $t$:
\begin{equation}
    \Delta \mathcal{R}_v = \Delta \mathcal{R}_{v,t}(\delta t) = \ell\left(y,\Phi_{W_t}\left(\Psi_{\Theta_t}\left(\gG_v^t\right)\right)\right) - \ell\left(y,\Phi_{W_{t_0}}\left(\Psi_{\Theta_{t_0}}\left(\gG_v^{t_0}\right)\right)\right) \,,
\end{equation}
with loss function $\ell$ and where we made explicit the decomposition into feature extraction ($\Psi_{\Theta}$) and linear classifier  ($\Phi_{\mathbf W}$); to avoid overwhelming notations, we omit the bias terms in $\Phi_{\mathbf W}$ and use $\smash{\gG^t_{v}}$ to indicate $\smash{\widetilde\gG^{(t,l)}_{v}}$. Then, assuming that the loss is Lipschitz continuous with constant $L_\ell$, and using the triangle inequality, we can isolate three components of forgetting risk:
\begin{align}
    \Delta \mathcal{R}_v \leq L_\ell & \left( \left\lVert \Phi_{W_t}\left(\Psi_{\Theta_t}\left(\gG_v^t\right)\right) - \Phi_{W_t}\left(\Psi_{\Theta_t}\left(\gG_v^{t_0}\right)\right)\right\rVert\right.  &\textit{structural drift}\\
    &+ \left. \left\lVert\Phi_{W_t}\left(\Psi_{\Theta_t}\left(\gG_v^{t_0}\right)\right) - \Phi_{W_t}\left(\Psi_{\Theta_{t_0}}\left(\gG_v^{t_0}\right)\right)\right\rVert\right. &\textit{backbone parameters drift} \\
    &+ \left. \left\lVert\Phi_{W_t}\left(\Psi_{\Theta_{t_0}}\left(\gG_v^{t_0}\right)\right) - \Phi_{W_{t_0}}\left(\Psi_{\Theta_{t_0}}\left(\gG_v^{t_0}\right)\right)\right\rVert \right). &\textit{classifier parameters drift}
\end{align}
While the first term is inherent in the data-generating process, and therefore irreducible, 
a crucial advantage of using a fixed feature extractor is that it eliminates the second term, as $\Theta_t = \Theta_{t_0}$. Furthermore, if we assume the norm of the embeddings to be bounded by $B_z$, the third term is bounded by $B_z \left\lVert \mW_t - \mW_{t_0} \right\rVert$. For SLDA, the term is expected to decrease as the number of observed examples of a class increases, due to the update scheme through separate, class-specific cumulative means;
this is in contrast with SGD-based training, where weight updates are less separated by class.
Therefore, SLDA with fixed feature extraction provides high stability, with decreasing forgetting risk as the node stream progresses. Higher stability compared to SGD-trained methods can be empirically observed in Figures~\ref{fig:anytime_evaluation_ugcn}-\ref{fig:anytime_evaluation_grnf} of Appendix \ref{app:plots}. 

This analysis illustrates the robustness against forgetting of untrained feature extraction with SLDA. However, stability is not sufficient for good CL performance: the model must also remain plastic enough to acquire new knowledge as the node stream evolves.
For this, we can rely on the over-parameterized randomized feature extractors, which give us expressive and general topological embedding independently of task, allowing for the training of just a simple linear classifier on top \citep{Scardapane_2017_RandomnessNeuralNetworks}.

\section{Experimental setting}\label{sec:setting}

For our experiments we adopt the OCGL setting described in \citep{ocgl}. 
In particular, we follow the requirement of neighborhood sampling, and we consider small mini-batches of nodes instead of individual ones. We compare the use of randomized feature extractors, UGCN and GRNF, with linear classifier, either SLDA or coupled with CL strategies, across multiple benchmarks.

\paragraph{Benchmarks} 
We use the same six node-classification graph datasets of \citep{ocgl}: CoraFull \citep{bojchevski_deep_2018}, Arxiv \citep{hu_open_2021}, Reddit \citep{hamilton_inductive_2017}, Amazon Computer \citep{shchur_pitfalls_2019}, Roman Empire \citep{Platonov_2022_CriticalLookEvaluation} and Elliptic \citep{Weber_2019_AntiMoneyLaunderingBitcoin}. On all except Elliptic (as it only has two classes) we consider a class-incremental stream: nodes in the graph arrive one by one in blocks consisting of two classes (each segment can be identified with a task in the context of CL, even though the models in this setting are agnostic to task boundaries). On Elliptic and Arxiv we consider a time-incremental stream: since real node timestamps are available, we use them for a realistic node stream (dividing the stream into 10 blocks simply for evaluation). We split the nodes in each graph into 60\% for training, 20\% for validation and 20\% for testing, and we use a transductive setting.
We consider the same mini-batch sizes as \citep{ocgl}: 10 for the smaller CoraFull, Amazon Computer and Roman Empire, 50 for Arxiv, Reddit and Elliptic.

\paragraph{Metrics} 
To evaluate model predictions in the considered setting, we use three metrics: \textit{Average Performance (AP)}, \textit{Average Forgetting (AF)} \citep{lopez-paz_gradient_2017}, and \textit{Average Anytime Performance (AAP)} \citep{caccia_new_2021}.
The performance metric is accuracy for all datasets except for Elliptic, as it is highly unbalanced with only two classes, and therefore F1 score of the minority class is used.
For anytime predictions, we obtain AAP by evaluating the model on the validation nodes after each training mini-batch. The metrics are described in detail in Appendix \ref{app:Metrics}.

\paragraph{Baselines}
In addition to the described SLDA, we couple the linear classifier with some popular CL strategies: we consider A-GEM \citep{chaudhry_efficient_2018}, ER \citep{chaudhry_tiny_2019}, EWC \citep{kirkpatrick_overcoming_2017}, LwF \citep{li_learning_2018} and MAS \citep{aljundi_memory_2018}. Furthermore, we consider the \textit{bare} baseline which consists of simply finetuning the linear layer on the stream without any CL method. We also provide the \textit{joint} baseline consisting of jointly training the linear layer offline on the embeddings from all the nodes in the entire final graph. We do not consider graph-specific methods, as once the features are extracted with the untrained backbone they are no longer relevant. However, we provide a comparison with recent state-of-the-art graph methods for OCGL \citep{ocgl} in Appendix \ref{app:trainedgcn} and in summary in Table \ref{tab:trained_results}.

\paragraph{Implementation details} 
Following \citep{ocgl}, we consider sparsified 2-hop node neighborhoods, sampling recursively 10 neighbors per layer for each node. For UGCN, we therefore use a 2-layer network with 1024 units per layer, resulting in a 2048-dimensional node embedding due to layer concatenation. For GRNF, we use 1024 features, which amount to a 2048-dimensional node embedding since we concatenate the equivariant and invariant components. In both cases, magnitude of weight initialization is regulated by a tunable gain hyperparameter. We use Adam optimizer \citep{kingma_adam_2017} without weight decay nor dropout, tuning the learning rate as an hyperparameter. Another hyperparameter is the number of passes on a mini-batch before passing to the next, as suggested by Aljundi et~al. \citep{aljundi_online_2019}. Hyperparameters are tuned following the protocol outlined by Chaudhry et~al. \citep{chaudhry_efficient_2018}: they are selected according to validation performance (AP) only on a small section of the data stream. All training and method specific hyperparameters are reported in Appendix \ref{app:hyperparameters}. All experiments are performed with 5 different random initializations, and results are reported as average and standard deviation over them.

\section{Results}\label{sec:results}

\renewcommand{\arraystretch}{0.9}
\setlength{\tabcolsep}{4pt}
\begin{table*}[!t]
\centering
\begin{small}
\begin{sc}
\begin{tabular}{llccclccc}
\toprule
                  &        & \multicolumn{3}{c}{UGCN}                                                                             &  & \multicolumn{3}{c}{GRNF}                                                                             \\ \cmidrule{3-5} \cmidrule{7-9} 
                  & Method & AP\% $\uparrow$                 & AAP$_{val}$\% $\uparrow$        & AF\% $\uparrow$                  &  & AP\% $\uparrow$                 & AAP$_{val}$\% $\uparrow$        & AF\% $\uparrow$                  \\ \midrule
\multirow{8}{*}{\rotatebox[origin=c]{90}{CoraFull}} & A-GEM  & $32.99 {\scriptstyle \pm 1.02}$ & $52.35 {\scriptstyle \pm 0.90}$ & $-50.53 {\scriptstyle \pm 1.19}$ &  & $33.33 {\scriptstyle \pm 0.42}$ & $48.04 {\scriptstyle \pm 0.89}$ & $-29.93 {\scriptstyle \pm 0.67}$ \\
                  & ER     & $55.67 {\scriptstyle \pm 0.47}$ & $65.01 {\scriptstyle \pm 0.60}$ & $-24.90 {\scriptstyle \pm 0.71}$ &  & $52.53 {\scriptstyle \pm 0.60}$ & $59.79 {\scriptstyle \pm 0.56}$ & $-27.40 {\scriptstyle \pm 0.70}$ \\
                  & EWC    & $34.46 {\scriptstyle \pm 1.28}$ & $52.28 {\scriptstyle \pm 0.35}$ & $-49.31 {\scriptstyle \pm 1.82}$ &  & $30.53 {\scriptstyle \pm 0.52}$ & $46.66 {\scriptstyle \pm 0.92}$ & $-50.44 {\scriptstyle \pm 0.75}$ \\
                  & LwF    & $35.30 {\scriptstyle \pm 0.50}$ & $52.94 {\scriptstyle \pm 1.18}$ & $-46.41 {\scriptstyle \pm 0.74}$ &  & $30.70 {\scriptstyle \pm 0.85}$ & $45.87 {\scriptstyle \pm 0.98}$ & $-35.06 {\scriptstyle \pm 0.50}$ \\
                  & MAS    & $34.44 {\scriptstyle \pm 1.30}$ & $52.59 {\scriptstyle \pm 0.36}$ & $-49.44 {\scriptstyle \pm 1.79}$ &  & $30.12 {\scriptstyle \pm 0.63}$ & $46.58 {\scriptstyle \pm 0.87}$ & $-51.28 {\scriptstyle \pm 0.73}$ \\
                  & SLDA   & $64.03 {\scriptstyle \pm 0.50}$ & $74.65 {\scriptstyle \pm 0.44}$ & $-14.48 {\scriptstyle \pm 0.39}$ &  & $62.05 {\scriptstyle \pm 0.27}$ & $72.42 {\scriptstyle \pm 0.31}$ & $-16.87 {\scriptstyle \pm 0.50}$ \\ \cmidrule{2-9} 
                  & bare   & $32.10 {\scriptstyle \pm 1.50}$ & $51.48 {\scriptstyle \pm 0.55}$ & $-51.85 {\scriptstyle \pm 1.62}$ &  & $28.18 {\scriptstyle \pm 0.71}$ & $44.82 {\scriptstyle \pm 1.11}$ & $-35.56 {\scriptstyle \pm 0.65}$ \\
                  & Joint  & $66.10 {\scriptstyle \pm 0.28}$ & -                               & $-8.91 {\scriptstyle \pm 0.36}$  &  & $65.23 {\scriptstyle \pm 0.78}$ & -                               & $-8.05 {\scriptstyle \pm 0.54}$  \\ \midrule
\multirow{8}{*}{\rotatebox[origin=c]{90}{A. Computer}} & A-GEM  & $60.35 {\scriptstyle \pm 4.86}$  & $58.12 {\scriptstyle \pm 3.18}$ & $-31.60 {\scriptstyle \pm 7.88}$  &  & $53.69 {\scriptstyle \pm 5.85}$  & $62.51 {\scriptstyle \pm 3.59}$ & $-39.10 {\scriptstyle \pm 8.55}$  \\
                 & ER     & $80.71 {\scriptstyle \pm 2.83}$  & $83.30 {\scriptstyle \pm 0.51}$ & $-13.32 {\scriptstyle \pm 3.99}$  &  & $62.62 {\scriptstyle \pm 10.93}$ & $75.25 {\scriptstyle \pm 2.93}$ & $-20.04 {\scriptstyle \pm 11.26}$ \\
                 & EWC    & $48.02 {\scriptstyle \pm 1.30}$  & $59.01 {\scriptstyle \pm 1.07}$ & $-30.29 {\scriptstyle \pm 1.55}$  &  & $30.80 {\scriptstyle \pm 5.50}$  & $49.10 {\scriptstyle \pm 1.05}$ & $-44.21 {\scriptstyle \pm 2.34}$  \\
                 & LwF    & $47.31 {\scriptstyle \pm 11.89}$ & $60.84 {\scriptstyle \pm 3.24}$ & $-30.40 {\scriptstyle \pm 13.68}$ &  & $40.59 {\scriptstyle \pm 5.37}$  & $53.53 {\scriptstyle \pm 1.97}$ & $-50.96 {\scriptstyle \pm 2.41}$  \\
                 & MAS    & $41.81 {\scriptstyle \pm 2.06}$  & $62.90 {\scriptstyle \pm 0.87}$ & $-45.72 {\scriptstyle \pm 3.10}$  &  & $43.14 {\scriptstyle \pm 3.89}$  & $57.19 {\scriptstyle \pm 2.01}$ & $-44.81 {\scriptstyle \pm 3.95}$  \\
                 & SLDA   & $86.65 {\scriptstyle \pm 0.48}$  & $90.64 {\scriptstyle \pm 0.13}$ & $-7.72 {\scriptstyle \pm 0.47}$   &  & $84.32 {\scriptstyle \pm 0.42}$  & $89.95 {\scriptstyle \pm 0.23}$ & $-10.94 {\scriptstyle \pm 0.39}$  \\ \cmidrule{2-9} 
                 & bare   & $43.98 {\scriptstyle \pm 3.39}$  & $50.21 {\scriptstyle \pm 1.26}$ & $-42.56 {\scriptstyle \pm 8.59}$  &  & $42.53 {\scriptstyle \pm 6.22}$  & $52.62 {\scriptstyle \pm 1.53}$ & $-49.47 {\scriptstyle \pm 2.49}$  \\
                 & Joint  & $86.84 {\scriptstyle \pm 0.47}$  & -                               & $-7.30 {\scriptstyle \pm 0.47}$   &  & $87.43 {\scriptstyle \pm 0.34}$  & -                               & $-6.80 {\scriptstyle \pm 0.33}$   \\ \midrule
\multirow{8}{*}{\rotatebox[origin=c]{90}{Arxiv}} & A-GEM  & $22.68 {\scriptstyle \pm 3.24}$ & $30.81 {\scriptstyle \pm 2.18}$ & $-59.57 {\scriptstyle \pm 1.86}$ &  & $14.32 {\scriptstyle \pm 1.82}$ & $32.39 {\scriptstyle \pm 1.80}$ & $-64.85 {\scriptstyle \pm 1.27}$ \\
                & ER     & $15.12 {\scriptstyle \pm 4.06}$ & $32.36 {\scriptstyle \pm 3.98}$ & $-49.55 {\scriptstyle \pm 4.96}$ &  & $14.63 {\scriptstyle \pm 1.78}$ & $31.84 {\scriptstyle \pm 2.05}$ & $-57.71 {\scriptstyle \pm 1.41}$ \\
                & EWC    & $17.26 {\scriptstyle \pm 2.12}$ & $27.71 {\scriptstyle \pm 1.02}$ & $-56.43 {\scriptstyle \pm 2.11}$ &  & $17.31 {\scriptstyle \pm 1.06}$ & $26.33 {\scriptstyle \pm 0.66}$ & $-67.39 {\scriptstyle \pm 0.52}$ \\
                & LwF    & $19.52 {\scriptstyle \pm 1.49}$ & $27.04 {\scriptstyle \pm 0.88}$ & $-56.52 {\scriptstyle \pm 0.92}$ &  & $21.72 {\scriptstyle \pm 0.98}$ & $29.84 {\scriptstyle \pm 0.40}$ & $-60.04 {\scriptstyle \pm 1.71}$ \\
                & MAS    & $18.77 {\scriptstyle \pm 2.44}$ & $28.92 {\scriptstyle \pm 1.27}$ & $-53.20 {\scriptstyle \pm 3.47}$ &  & $16.45 {\scriptstyle \pm 1.35}$ & $29.10 {\scriptstyle \pm 0.82}$ & $-67.50 {\scriptstyle \pm 1.20}$ \\
                & SLDA   & $55.71 {\scriptstyle \pm 0.08}$ & $64.55 {\scriptstyle \pm 0.03}$ & $-18.47 {\scriptstyle \pm 0.08}$ &  & $52.69 {\scriptstyle \pm 0.19}$ & $62.69 {\scriptstyle \pm 0.07}$ & $-27.67 {\scriptstyle \pm 0.16}$ \\ \cmidrule{2-9} 
                & bare   & $21.38 {\scriptstyle \pm 3.37}$ & $24.28 {\scriptstyle \pm 1.69}$ & $-41.42 {\scriptstyle \pm 2.17}$ &  & $14.13 {\scriptstyle \pm 1.14}$ & $24.67 {\scriptstyle \pm 0.33}$ & $-75.06 {\scriptstyle \pm 1.31}$ \\
                & Joint  & $59.06 {\scriptstyle \pm 0.15}$ & -                               & $-16.09 {\scriptstyle \pm 0.30}$ &  & $57.87 {\scriptstyle \pm 0.28}$ & -                               & $-16.72 {\scriptstyle \pm 0.19}$ \\ \midrule
\multirow{8}{*}{\rotatebox[origin=c]{90}{Reddit}} & A-GEM  & $60.60 {\scriptstyle \pm 0.88}$ & $78.29 {\scriptstyle \pm 0.31}$ & $-36.77 {\scriptstyle \pm 0.89}$ &  & $46.51 {\scriptstyle \pm 0.91}$ & $61.06 {\scriptstyle \pm 0.20}$ & $-25.63 {\scriptstyle \pm 0.96}$ \\
                  & ER     & $80.40 {\scriptstyle \pm 1.04}$ & $89.75 {\scriptstyle \pm 0.05}$ & $-16.60 {\scriptstyle \pm 1.11}$ &  & $83.46 {\scriptstyle \pm 0.43}$ & $89.22 {\scriptstyle \pm 0.12}$ & $-14.44 {\scriptstyle \pm 0.40}$ \\
                  & EWC    & $40.91 {\scriptstyle \pm 1.22}$ & $61.13 {\scriptstyle \pm 1.07}$ & $-49.58 {\scriptstyle \pm 1.37}$ &  & $44.26 {\scriptstyle \pm 1.12}$ & $60.49 {\scriptstyle \pm 0.45}$ & $-29.80 {\scriptstyle \pm 1.18}$ \\
                  & LwF    & $13.60 {\scriptstyle \pm 0.43}$ & $49.00 {\scriptstyle \pm 0.85}$ & $-84.37 {\scriptstyle \pm 0.43}$ &  & $39.09 {\scriptstyle \pm 0.77}$ & $58.00 {\scriptstyle \pm 1.07}$ & $-35.99 {\scriptstyle \pm 0.87}$ \\
                  & MAS    & $11.93 {\scriptstyle \pm 0.66}$ & $50.54 {\scriptstyle \pm 0.48}$ & $-86.50 {\scriptstyle \pm 0.66}$ &  & $42.99 {\scriptstyle \pm 0.99}$ & $58.22 {\scriptstyle \pm 0.50}$ & $-32.62 {\scriptstyle \pm 0.97}$ \\
                  & SLDA   & $89.31 {\scriptstyle \pm 0.03}$ & $95.17 {\scriptstyle \pm 0.03}$ & $-5.56 {\scriptstyle \pm 0.09}$  &  & $91.42 {\scriptstyle \pm 0.14}$ & $95.84 {\scriptstyle \pm 0.03}$ & $-3.81 {\scriptstyle \pm 0.15}$  \\ \cmidrule{2-9} 
                  & bare   & $41.35 {\scriptstyle \pm 1.27}$ & $60.55 {\scriptstyle \pm 1.05}$ & $-48.97 {\scriptstyle \pm 1.26}$ &  & $42.67 {\scriptstyle \pm 0.75}$ & $57.80 {\scriptstyle \pm 0.36}$ & $-33.13 {\scriptstyle \pm 0.41}$ \\
                  & Joint  & $88.07 {\scriptstyle \pm 0.20}$ & -                               & $-3.57 {\scriptstyle \pm 0.19}$  &  & $90.90 {\scriptstyle \pm 0.17}$ & -                               & $-2.77 {\scriptstyle \pm 0.17}$  \\ \midrule
\multirow{8}{*}{\rotatebox[origin=c]{90}{Roman E.}} & A-GEM  & $17.18 {\scriptstyle \pm 0.46}$ & $43.99 {\scriptstyle \pm 0.20}$ & $-69.65 {\scriptstyle \pm 0.42}$ &  & $31.19 {\scriptstyle \pm 0.98}$ & $37.77 {\scriptstyle \pm 1.42}$ & $-23.28 {\scriptstyle \pm 4.54}$ \\
                  & ER     & $24.80 {\scriptstyle \pm 1.21}$ & $46.35 {\scriptstyle \pm 0.50}$ & $-32.56 {\scriptstyle \pm 0.84}$ &  & $40.45 {\scriptstyle \pm 1.18}$ & $46.83 {\scriptstyle \pm 1.46}$ & $-15.68 {\scriptstyle \pm 2.20}$ \\
                  & EWC    & $15.82 {\scriptstyle \pm 1.00}$ & $37.30 {\scriptstyle \pm 0.50}$ & $-43.62 {\scriptstyle \pm 0.95}$ &  & $21.54 {\scriptstyle \pm 0.85}$ & $40.22 {\scriptstyle \pm 0.91}$ & $-22.71 {\scriptstyle \pm 2.53}$ \\
                  & LwF    & $21.51 {\scriptstyle \pm 0.65}$ & $45.51 {\scriptstyle \pm 0.37}$ & $-42.38 {\scriptstyle \pm 0.83}$ &  & $20.29 {\scriptstyle \pm 1.84}$ & $37.25 {\scriptstyle \pm 0.76}$ & $-46.63 {\scriptstyle \pm 2.02}$ \\
                  & MAS    & $16.33 {\scriptstyle \pm 0.66}$ & $35.12 {\scriptstyle \pm 0.81}$ & $-28.71 {\scriptstyle \pm 1.77}$ &  & $22.47 {\scriptstyle \pm 0.58}$ & $40.52 {\scriptstyle \pm 0.42}$ & $-26.38 {\scriptstyle \pm 0.60}$ \\
                  & SLDA   & $34.61 {\scriptstyle \pm 0.02}$ & $57.57 {\scriptstyle \pm 0.06}$ & $-34.38 {\scriptstyle \pm 0.20}$ &  & $52.71 {\scriptstyle \pm 0.09}$ & $72.07 {\scriptstyle \pm 0.03}$ & $-29.43 {\scriptstyle \pm 0.21}$ \\ \cmidrule{2-9} 
                  & bare   & $7.09 {\scriptstyle \pm 0.03}$  & $34.74 {\scriptstyle \pm 0.69}$ & $-77.07 {\scriptstyle \pm 0.93}$ &  & $16.82 {\scriptstyle \pm 0.29}$ & $42.44 {\scriptstyle \pm 0.66}$ & $-68.56 {\scriptstyle \pm 1.54}$ \\
                  & Joint  & $49.12 {\scriptstyle \pm 0.51}$ & -                               & $-5.22 {\scriptstyle \pm 0.48}$  &  & $71.50 {\scriptstyle \pm 0.10}$ & -                               & $-6.54 {\scriptstyle \pm 0.37}$  \\ 
\bottomrule
\end{tabular}
\end{sc}
\end{small}
\caption{Results for class-incremental node stream.}
\label{tab:results_class_incremental}
\end{table*}

\begin{table*}[t]
\centering
\begin{small}
\begin{sc}
\begin{tabular}{llccclccc}
\toprule
                  &        & \multicolumn{3}{c}{UGCN}                                                                             &  & \multicolumn{3}{c}{GRNF}                                                                             \\ \cmidrule{3-5} \cmidrule{7-9} 
                  & Method & AP\% $\uparrow$                 & AAP$_{val}$\% $\uparrow$        & AF\% $\uparrow$                  &  & AP\% $\uparrow$                 & AAP$_{val}$\% $\uparrow$        & AF\% $\uparrow$                  \\ \midrule
\multirow{8}{*}{\rotatebox[origin=c]{90}{Elliptic}} & A-GEM  & $42.43 {\scriptstyle \pm 0.99}$ & $40.72 {\scriptstyle \pm 0.57}$ & $-13.51 {\scriptstyle \pm 1.54}$ &  & $53.65 {\scriptstyle \pm 0.73}$ & $50.10 {\scriptstyle \pm 0.33}$ & $-12.08 {\scriptstyle \pm 1.16}$  \\
                  & ER     & $44.61 {\scriptstyle \pm 1.29}$ & $45.54 {\scriptstyle \pm 0.40}$ & $-9.17 {\scriptstyle \pm 1.82}$ &  & $57.30 {\scriptstyle \pm 1.69}$ & $53.36 {\scriptstyle \pm 1.29}$ & $-8.90 {\scriptstyle \pm 1.65}$  \\
                  & EWC    & $36.62 {\scriptstyle \pm 1.23}$ & $34.64 {\scriptstyle \pm 0.20}$ & $-13.36 {\scriptstyle \pm 1.59}$ &  & $50.44 {\scriptstyle \pm 1.45}$ & $48.81 {\scriptstyle \pm 0.36}$ & $-13.77 {\scriptstyle \pm 1.16}$ \\
                  & LwF    & $37.88 {\scriptstyle \pm 1.52}$ & $34.63 {\scriptstyle \pm 0.30}$ & $-11.43 {\scriptstyle \pm 1.39}$ &  & $55.16 {\scriptstyle \pm 1.26}$ & $50.15 {\scriptstyle \pm 0.59}$ & $-10.45 {\scriptstyle \pm 0.88}$  \\
                  & MAS    & $36.80 {\scriptstyle \pm 0.84}$ & $34.47 {\scriptstyle \pm 0.21}$ & $-13.21 {\scriptstyle \pm 1.30}$ &  & $50.26 {\scriptstyle \pm 1.19}$ & $48.30 {\scriptstyle \pm 0.36}$ & $-14.29 {\scriptstyle \pm 0.97}$ \\
                  & SLDA   & $55.49 {\scriptstyle \pm 0.64}$ & $54.13 {\scriptstyle \pm 0.28}$ & $-1.85 {\scriptstyle \pm 1.17}$ &  & $65.92 {\scriptstyle \pm 0.71}$ & $65.43 {\scriptstyle \pm 0.53}$ & $-3.21 {\scriptstyle \pm 1.90}$  \\ \cmidrule{2-9} 
                  & bare   & $37.33 {\scriptstyle \pm 0.68}$ & $34.97 {\scriptstyle \pm 0.32}$ & $-12.21 {\scriptstyle \pm 1.01}$ &  & $50.14 {\scriptstyle \pm 1.44}$ & $48.81 {\scriptstyle \pm 0.45}$ & $-13.23 {\scriptstyle \pm 1.11}$ \\
                  & Joint  & $59.29 {\scriptstyle \pm 0.82}$ & -                               & $-4.94 {\scriptstyle \pm 0.62}$ &  & $70.23 {\scriptstyle \pm 0.76}$ & -                               & $-3.59 {\scriptstyle \pm 1.00}$  \\ \midrule
\multirow{8}{*}{\rotatebox[origin=c]{90}{Arxiv}} & A-GEM  & $67.66 {\scriptstyle \pm 0.12}$ & $64.97 {\scriptstyle \pm 0.07}$ & $0.98 {\scriptstyle \pm 0.17}$  &  & $67.73 {\scriptstyle \pm 0.06}$ & $63.07 {\scriptstyle \pm 0.11}$ & $2.32 {\scriptstyle \pm 0.09}$ \\
                  & ER     & $68.67 {\scriptstyle \pm 0.18}$ & $65.26 {\scriptstyle \pm 0.05}$ & $2.12 {\scriptstyle \pm 0.18}$ &  & $68.30 {\scriptstyle \pm 0.04}$ & $63.56 {\scriptstyle \pm 0.09}$ & $2.93 {\scriptstyle \pm 0.09}$   \\
                  & EWC    & $67.65 {\scriptstyle \pm 0.07}$ & $64.96 {\scriptstyle \pm 0.03}$ & $0.94 {\scriptstyle \pm 0.08}$ &  & $67.65 {\scriptstyle \pm 0.08}$ & $63.93 {\scriptstyle \pm 0.12}$ & $1.48 {\scriptstyle \pm 0.10}$  \\
                  & LwF    & $68.21 {\scriptstyle \pm 0.09}$ & $65.22 {\scriptstyle \pm 0.08}$ & $1.37 {\scriptstyle \pm 0.09}$ &  & $67.56 {\scriptstyle \pm 0.12}$ & $62.94 {\scriptstyle \pm 0.12}$ & $2.22 {\scriptstyle \pm 0.11}$ \\
                  & MAS    & $67.65 {\scriptstyle \pm 0.07}$ & $64.96 {\scriptstyle \pm 0.03}$ & $0.94 {\scriptstyle \pm 0.08}$ &  & $67.64 {\scriptstyle \pm 0.09}$ & $63.93 {\scriptstyle \pm 0.12}$ & $1.47 {\scriptstyle \pm 0.10}$  \\
                  & SLDA   & $66.69 {\scriptstyle \pm 0.07}$ & $61.97 {\scriptstyle \pm 0.04}$ & $2.29 {\scriptstyle \pm 0.17}$   &  & $65.09 {\scriptstyle \pm 0.21}$ & $60.14 {\scriptstyle \pm 0.10}$ & $2.42 {\scriptstyle \pm 0.20}$   \\ \cmidrule{2-9} 
                  & bare   & $67.59 {\scriptstyle \pm 0.20}$ & $64.98 {\scriptstyle \pm 0.04}$ & $0.97 {\scriptstyle \pm 0.18}$  &  & $67.68 {\scriptstyle \pm 0.09}$ & $63.89 {\scriptstyle \pm 0.04}$ & $1.57 {\scriptstyle \pm 0.10}$ \\
                  & Joint  & $70.35 {\scriptstyle \pm 0.16}$ & -                               & $2.67 {\scriptstyle \pm 0.30}$   &  & $69.53 {\scriptstyle \pm 0.12}$ & -                               & $2.89 {\scriptstyle \pm 0.09}$   \\
\bottomrule
\end{tabular}
\end{sc}
\end{small}
\caption{Results for time-incremental node stream.}
\label{tab:results_time_incremental}
\end{table*}

\begin{table*}[t]
\centering
\begin{small}
\begin{sc}
\begin{tabular}{lcccccc}
\toprule
Method    & CoraFull                        & A. Computer                     & Arxiv {\scriptsize(cl.-incr.)}          & Reddit                          & Roman E.                        & Elliptic                        \\ \midrule
Best OCGL    & $40.45 {\scriptstyle \pm 0.77}$ & $70.45 {\scriptstyle \pm 3.66}$ & $35.86 {\scriptstyle \pm 1.20}$ & $58.08 {\scriptstyle \pm 8.04}$ & $14.20 {\scriptstyle \pm 0.87}$ & $51.13 {\scriptstyle \pm 1.74}$ \\
Joint    & $67.55 {\scriptstyle \pm 0.05}$ & $83.07 {\scriptstyle \pm 1.30}$ & $58.58 {\scriptstyle \pm 0.28}$ & $90.02 {\scriptstyle \pm 0.12}$ & $39.47 {\scriptstyle \pm 0.33}$ & $71.97 {\scriptstyle \pm 0.83}$ \\ \midrule
ugcn+slda & \textcolor{Green}{+23.6}       & \textcolor{Green}{+16.2}      & \textcolor{Green}{+19.9}       & \textcolor{Green}{+31.2}       & \textcolor{Green}{+20.4}       & \textcolor{Green}{+4.4}        \\
grnf+slda & \textcolor{Green}{+21.6}       & \textcolor{Green}{+13.9}       & \textcolor{Green}{+16.8}       & \textcolor{Green}{+33.3}       & \textcolor{Green}{+38.5}       & \textcolor{Green}{+14.8}       \\ \bottomrule
\end{tabular}
\end{sc}
\end{small}
\caption{Top rows: best AP results from Table \ref{tab:full_trained_results} of Appendix \ref{app:trainedgcn} for CL strategies with trained GCN and joint offline training upper bound. Bottom rows: increase in AP of the proposed approach (Tables \ref{tab:results_class_incremental}-\ref{tab:results_time_incremental}) compared to the best performing method with trained GCN. All the increases in performance show statistical significance with p-value $\leq 0.005$ with a one-sided Mann–Whitney U test.}
\label{tab:trained_results}
\end{table*}

The results of our experiments in the OCGL setting are reported in Table \ref{tab:results_class_incremental} for benchmarks with class-incremental node stream and Table \ref{tab:results_time_incremental} for those on which a time-incremental stream is defined. Additionally, for comparison with results obtainable by a model trained end-to-end, we report in Table \ref{tab:trained_results} the best AP reported in \citep{ocgl} (Arxiv is reported only with class-incremental stream as the time-incremental one is not considered in \citep{ocgl}) by any OCGL strategy, in most cases SSM \citep{zhang_sparsified_2022} and PDGNN \citep{Zhang_2024}, together with the respective joint upper bound, and we highlight the increase in performance using the proposed decoupled approach of randomized feature extraction with SLDA.

\paragraph{Class-incremental benchmarks}

We first observe from Table \ref{tab:results_class_incremental} that the upper bounds with randomized features are comparable to those obtained with a trained GCN model, as reported in Table \ref{tab:trained_results}. This proves that the extracted features are expressive enough for these tasks, confirming our approach's viability. On Roman Empire specifically, which is highly heterophilic, we see a much higher upper bound with GRNF: this is due to the implicit bias of the GCN design, which smooths node embeddings with neighborhood information, while GRNF are more expressive, as they are derived from a universal approximator of graph functions. We also note how AF for the joint baseline represents the increasing difficulty of the classification task as new classes are added.

From the results the superior performance of SLDA also emerges clearly, as it outperforms all considered CL methods on a linear layer trained with gradient descent, and in most cases with a wide margin. 
Additionally, the AP results of SLDA with both UGCN and GRNF closely approach their respective upper bounds. 
The results of the randomized feature extractors coupled with SLDA also significantly outperform the best results obtained with state-of-the-art replay methods tailored for graphs, as seen in Table \ref{tab:trained_results}, despite SLDA not using any memory buffer. The differences between UGCN and GRNF are in most case limited, especially for the SLDA classifier, with UGCN showing a slight advantage on CoraFull, Amazon Computer and Arxiv benchmarks, while GRNF appears superior on Reddit and more significantly on Roman Empire, due to the heterophily of the graph as discussed above.

In general, compared to the full results in Appendix \ref{app:trainedgcn}, also most other CL strategies in most benchmarks report performance improvements compared to the use of a trained GCN, confirming the benefits of keeping a frozen feature extractor immune to forgetting. ER specifically is significantly better when coupled with randomized features compared to a trained GCN, in some cases approaching SLDA, due to the fact that in this setting the memory buffer is much more informative, as stored examples contain neighborhood information, albeit subject to structural shift.

\paragraph{Time-incremental benchmarks}

The time-incremental setting naturally possesses more docile shifts in distribution, compared to the abrupt class changes of the class-incremental benchmark.
Nonetheless, on Elliptic CL methods are still beneficial, with SLDA still outperforming all baselines, also compared to the trained GCN state-of-the-art (Table \ref{tab:trained_results}).
Importantly, our results on Elliptic, which is a dataset of real Bitcoin transactions, highlight the feasibility and benefits of the proposed approach on realistic data streams, beyond the academic class-incremental setting.
On the other hand, for the Arxiv time-incremental benchmark, we see little difference in the results of the various strategies, with SLDA no longer over-performing. In fact, we even see positive AF values for all methods, indicating that the classification becomes easier, rather than harder, as the node stream goes on. This is because Arxiv does not present a significant drift in class distribution throughout time.
Therefore, a CL learning approach here is less meaningful, as the bare baseline is already close to the upper bound. Nevertheless, ER proves beneficial, and these high results are further proof that the feature extractors are robust to structural shifts that still remain.

\paragraph{Impact of number of features}

\begin{figure*}[t]
\centering
\begin{subfigure}[b]{0.24\textwidth}
    \includegraphics[width=\textwidth]{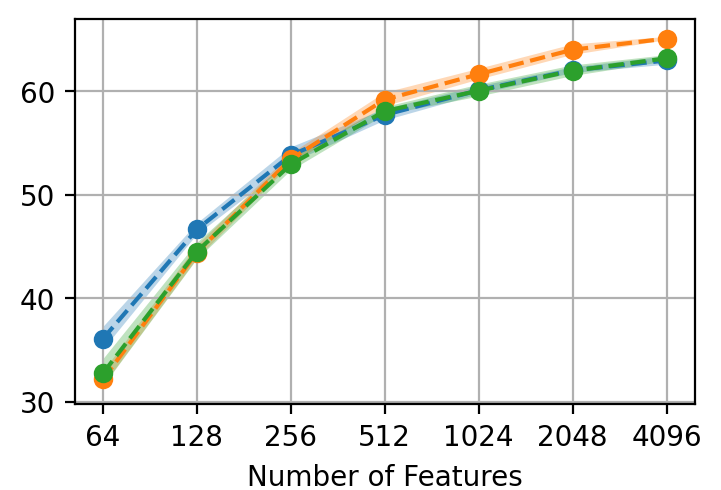}
    \caption{CoraFull}
    \label{fig:n_features_cora}
\end{subfigure}
\hfill
\begin{subfigure}[b]{0.24\textwidth}
    \includegraphics[width=\textwidth]{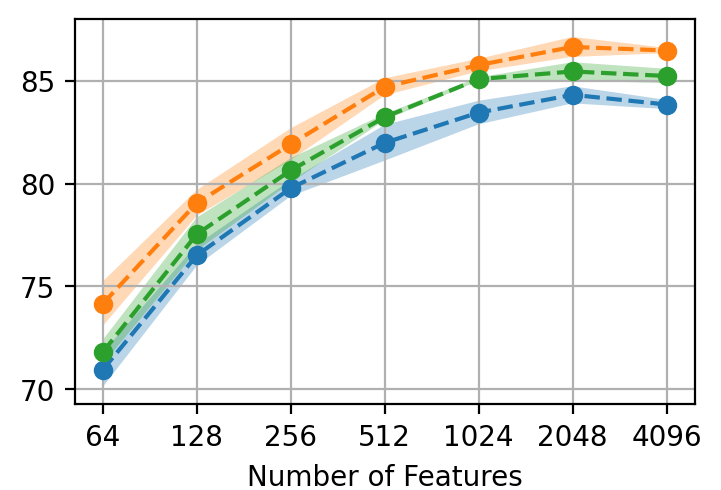}
    \caption{Amazon Computer}
    \label{fig:n_features_computer}
\end{subfigure}
\hfill
\begin{subfigure}[b]{0.24\textwidth}
    \includegraphics[width=\textwidth]{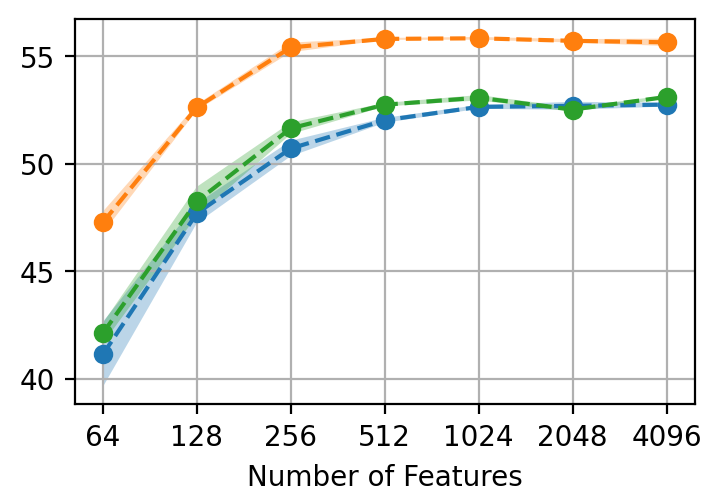}
    \caption{Arxiv (class-incr.)}
    \label{fig:n_features_arxiv}
\end{subfigure}
\hfill
\begin{subfigure}[b]{0.24\textwidth}
    \includegraphics[width=\textwidth]{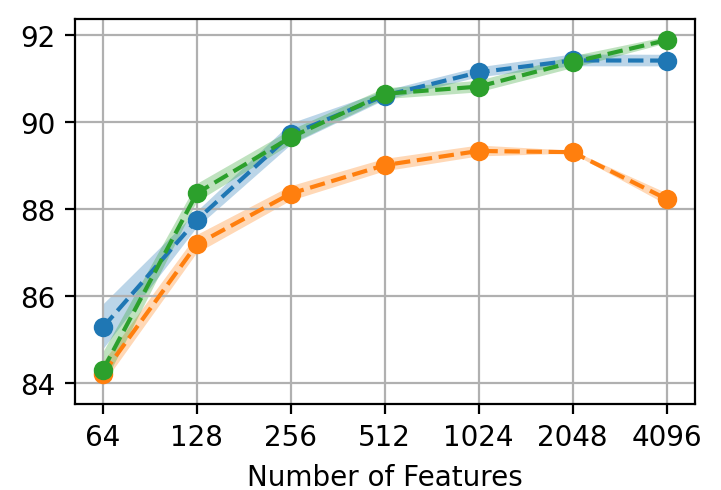}
    \caption{Reddit}
    \label{fig:n_features_reddit}
\end{subfigure}
\\
\vspace{3mm}
\begin{subfigure}[b]{0.24\textwidth}
    \includegraphics[width=\textwidth]{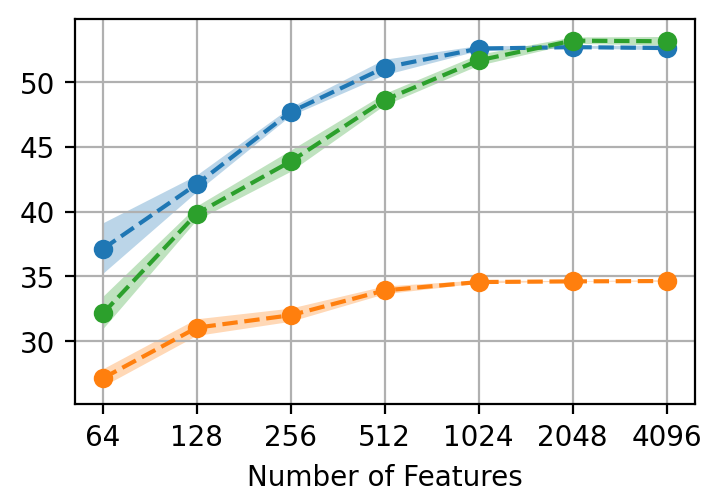}
    \caption{Roman Empire}
    \label{fig:n_features_roman}
\end{subfigure}
\hspace{2mm}
\begin{subfigure}[b]{0.24\textwidth}
    \includegraphics[width=\textwidth]{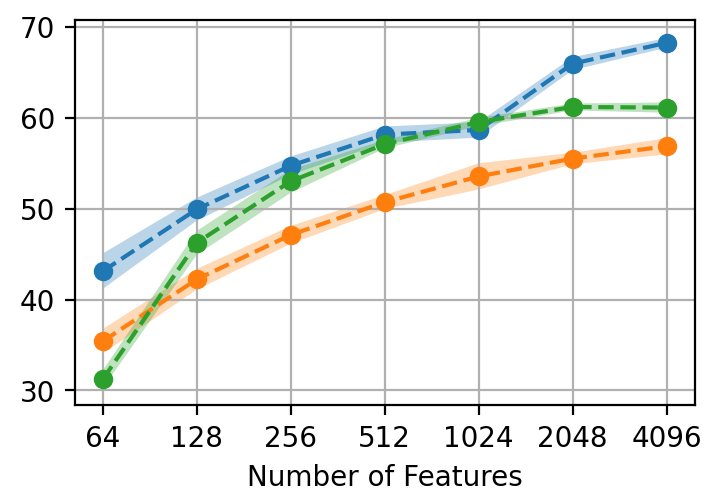}
    \caption{Elliptic}
    \label{fig:n_features_elliptic}
\end{subfigure}
\hspace{2mm}
\begin{subfigure}[b]{0.24\textwidth}
    \includegraphics[width=\textwidth]{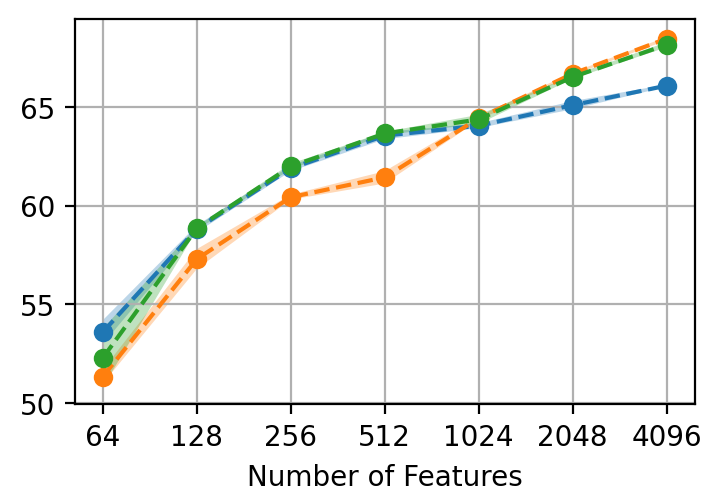}
    \caption{Arxiv (time-incr.)}
    \label{fig:n_features_arxivTI}
\end{subfigure}
\begin{subfigure}[t]{0.10\textwidth}
\adjustbox{raise=1.35cm,margin=3mm 0mm 0mm 0mm}{
    \includegraphics[width=\textwidth]{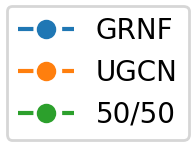}
    }
\end{subfigure}
\caption{Comparison of AP of SLDA with different number of features extracted with GRNF, UGCN, and a 50/50 mix of the two. The shaded area covers one standard deviation.}
\label{fig:n_features}
\end{figure*}

Given the overwhelming over-performance of randomized feature extraction with SLDA, we investigate the impact of the number of extracted features on model performance. In Figure \ref{fig:n_features}, we see that, with a lower number of features AP decreases as well, even though on most benchmarks even as few as 64 randomized features are sufficient to obtain results on par with the state-of-the-art CL methods on trained GCN of Table \ref{tab:trained_results}. Also, for many benchmarks performance seems to not have reached saturation even at 4096 features, indicating that further gains could be achieved with a larger feature extractor.

Finally, since UGCN and GRNF appear to have different strengths over the multiple benchmarks, we consider a hybrid feature extractor that extracts half of the features with UGCN, and half with GRNF. This mixed feature extractor shows a generally more robust performance over the benchmarks, with performance that is never lower than the individual two, except for one single point. Generally the hybrid extractor also does not improve over the best single one, indicating that the two types of feature do not benefit from integration. However, this strategy could be used to obtain a reliable feature extractor without the need to evaluate the two strategies.

\section{Conclusion}

This work introduces a simple, yet surprisingly effective approach for Online Continual Graph Learning, addressing forgetting by decoupling representation learning from classification. We use randomized, fixed node feature extractors that encode neighborhood information, coupled with a lightweight linear classifier trained incrementally on the node stream. By leveraging two types of untrained feature extractors -- UGCN and GRNF -- the proposed method provides robust and expressive node embeddings, resistant to catastrophic forgetting. Extensive experiments in the OCGL setting demonstrate that when paired with SLDA, this approach significantly outperforms other Continual Learning strategies, including state-of-the-art replay-based methods tailored for graph data. The method achieves performance close to joint offline training across various benchmarks. Beyond strong performance, its efficient streaming updates and no reliance on memory buffers make it a scalable and practical approach to deal with real-time classification on graph node streams.

\paragraph{Limitations and future directions} 
While we provide motivation and empirical evidence of the improved resistance to forgetting and higher prediction accuracy of our approach, a thorough theoretical analysis is yet to be developed.
Secondly, we highlight how our results are specific for the challenging OCGL scenario, while for different, offline settings other methods can perform more favorably. 
Finally, we focus on node-level classification, as graph-level is less interesting for OCGL, and leave regression and edge-level tasks as future research.

\section*{Acknowledgments}
This research was supported by the European Union - NextGenerationEU as part of the Italian National Recovery and Resilience Plan (PNRR), the project ``Lifelong Learning on large-scale and structured data'' funded by the EC-funded OCRE project, and the Swiss National Science Foundation project FNS 204061: ``HORD GNN: Higher-Order Relations and Dynamics in Graph Neural Networks''.

\newpage
\appendix

\section{Benchmarks}\label{app:benchmarks}

The benchmarks for out experiments are obtained from six node-level classification graph datasets. The CoraFull dataset \cite{bojchevski_deep_2018} is a citation network where nodes represent research papers and edges denote citation links between them, with labels corresponding to paper topics. Amazon Computer \cite{shchur_pitfalls_2019} is a co-purchase graph, with nodes representing products and edges indicating frequent co-purchases in the computer category on Amazon. Arxiv \cite{hu_open_2021} is a larger citation network based on arXiv submissions in the Computer Science domain. The Reddit dataset \cite{hamilton_inductive_2017} comprises posts from various Reddit communities, where each node represents a post, and edges connect posts that were commented on by the same user, capturing user interaction patterns. Roman Empire \cite{Platonov_2022_CriticalLookEvaluation} is an heterophilous dataset constructed from the corresponding Wikipedia page, where nodes are words linked through syntactic relationships or adjacency in the text. Lastly, the Elliptic dataset \cite{Weber_2019_AntiMoneyLaunderingBitcoin} is a graph of Bitcoin transactions, with edges representing the flow of funds. Only a subset of nodes are labeled as either \textit{licit} (42,019 nodes) or \textit{illicit} (4,545 nodes) transactions. Summary statistics for the six datasets are provided in Table \ref{tab:datasets}.

\begin{table*}[htbp]
\begin{center}
\begin{tabular}{lcccccc}
\toprule
Dataset    & CoraFull & Amazon Computer & Arxiv     & Reddit      & Roman Empire & Elliptic \\ \midrule
\# nodes   & 19,793   & 13,752          & 169,343   & 227,853     & 22,662       & 203,769  \\
\# edges   & 130,622  & 491,722         & 1,166,243 & 114,615,892 & 32,927       & 234,355  \\
\# classes & 70       & 10              & 40        & 40          & 18           & 2        \\ \bottomrule
\end{tabular}
\end{center}
\caption{Dataset statistics.}
\label{tab:datasets}
\end{table*}

\section{Metrics} 
\label{app:Metrics}

Due to the way the node stream is built, with a definition of task boundaries, we can make use of two commonly adopted continual learning (CL) metrics: \textit{Average Performance (AP)} and \textit{Average Forgetting (AF)} \cite{lopez-paz_gradient_2017}. These metrics are both derived from the more general performance matrix $\mM \in \R^{T \times T}$, where $T$ denotes the total number of tasks, and each element $M_{i,j}$ corresponds to the test performance on task $j$ after training on task $i$.

The \textit{Average Performance} is given by $\text{AP} = \frac{1}{T} \sum_{i=1}^T M_{T,i}$, representing the model’s performance on all tasks after completing the full training stream. The \textit{Average Forgetting} is computed as $\text{AF} = \frac{1}{T-1} \sum_{i=1}^{T-1} M_{T,i} - M_{i,i}$, and quantifies how much the model's performance on each task has deteriorated between its initial learning and the end of training. For evaluating performance, we rely on classification accuracy across all datasets, except for Elliptic, which is significantly imbalanced. For this dataset, we instead report the F1 score specific to the \textit{illicit} class.

To track model behavior throughout the node stream, we also employ anytime evaluation: the model is evaluated on validation nodes after each mini-batch update \cite{koh_online_2021}. This provides a fine-grained view of model performance over time, revealing its adaptability to distributional shifts. We quantify this using the \textit{Average Anytime Performance (AAP)} metric \cite{caccia_new_2021}, a generalization of average incremental accuracy for the online scenario. Letting AP$t$ denote the average accuracy after processing the $t$-th mini-batch, and $n$ be the total number of mini-batches, AAP is defined as $\text{AAP} = \frac{1}{n}\sum{t=1}^n \text{AP}_t$. This metric can be interpreted as the area under the accuracy curve across the training process \cite{koh_online_2021}.

\section{Online feature centering trick}\label{app:centering}

As in our experiments we consider also using a standard linear layer trained continually with gradient descent instead of SLDA, in this case it is beneficial to have featured centered in the origin. This is especially true due to the online setting, as having centered features can make the learning of bias parameters for newly observed classes faster, since we initialize the weights symmetrically around zero. Therefore, we adopt an online centering procedure, which allows us to keep the features centered at any point during the stream. Specifically, we maintain an updated global cumulative mean $m$ of node embeddings, so that
\begin{equation}
    m^{(t)} = \frac{(t-1) m^{(t-1)} + \vz^{(t)}}{t} \,, \label{eq:centering}
\end{equation}
where for ease of notation we use $\vz^{(t)} = \vz_{v_t}^{(t)}$. With this, we then feed the centered embeddings $\smash{\vz_v^{(t)} - m^{(t)}}$ in equation (3) of the paper.
To ensure consistency of model predictions with online feature centering under updates of the embedding mean as in \eqref{eq:centering}, we want to correct the bias term $\vb \to \vb'$ to ensure that for any $\vz \in \R^d$ predictions don't change, that is 
\begin{equation}
    \mW (\vz - m^{(t)}) + \vb' = \mW (\vz - m^{(t-1)}) + \vb  \,.
\end{equation}
Simplifying and using the definition of the update of $m^{(t)}$ in \eqref{eq:centering}, we obtain:
\begin{align}
   - \mW m^{(t)} + \vb' &= - \mW m^{(t-1)} + \vb\,, \\
   - \frac{(t-1)}{t} \mW m^{(t-1)} - \frac{1}{t} \mW \vz^{(t)} + \vb' &= - \mW m^{(t-1)} + \vb \,, \\
   \vb' &= -  \frac{1}{t} \mW m^{(t-1)} +  \frac{1}{t} \mW \vz^{(t)} + \vb \,, \\
   \vb' &= \frac{1}{t} \mW (\vz^{(t)} - m^{(t-1)}) + \vb \,.
\end{align}
This is the bias update formula, and a similar one can be derived for mini-batch updates. We highlight how this bias correction is performed only for seen classes, as we grow the classification head when new classes are encountered. Also, this online centering procedure is not performed with SLDA, since it is a prototype-based classifier and is thus indifferent to feature centering. 
Empirically, we have observed that this trick greatly improves performance when the node embeddings are not centered, which is often the case with GRNF.

\section{Hyperparameters}\label{app:hyperparameters}

We perform model selection using a limited section of the node stream, approximately 20\% of the tasks. Therefore, for class-incremental stream we validate over 7 out 35 tasks for CoraFull, 2 out of 5 for Amazon Computer (as considering 20\% of the tasks would mean using only 1, thus without any CL aspect), 4 out of 20 for Arxiv and Reddit, and 2 out of 9 for Roman Empire. 
For the time-incremental stream, we validate over the first 20\% of the nodes (i.e., 2 out of 10 tasks). The hyperparameters are selected by running a standard grid search, over the search space that we illustrate here. For all experiments and both backbones, we consider the gain hyperparameter for weight initialization in $\{0.1, 1, 10\}$. For all methods, except SLDA, we select the learning rate from the set $\{0.01, 0.001, 0.0001, 0.00001\}$, and the number of passes on each batch before going to the next one between 1 and 5. For ER and A-GEM, we consider the proportion of memories to use with respect to each training batch in $\{1,2,3\}$. Additionally, we set the same memory buffer size as \cite{ocgl}: 4000 for CoraFull, Amazon Computer, Roman Empire and Elliptic, 16000 for Arxiv and Reddit. The regularization hyperparameter for EWC and MAS is selected in $\{10^0, 10^2, 10^4, 10^6, 10^8, 10^{10}\}$. For LwF, we consider lambda\_dist in $\{1,10\}$, T in $\{0.2,2\}$ and the number of mini-batches after which to update the teacher model in $\{10,100\}$.

We highlight how the only hyperparameter considered for SLDA is the gain of the backbone, making it even easier to use than other methods, avoiding expensive hyperparameter search.

\section{Results with trained GCN}\label{app:trainedgcn}

In Table \ref{tab:full_trained_results} we report the AP of CL strategies when used with a trained GCN in the OCGL setting \cite{ocgl}. These results are obtained in the same configurations used for the experiments in the main paper, only with a trained 2-layer GCN with 256 hidden units instead of an untrained feature extractor. The results on Arxiv are provided only for the class-incremental stream. For the full results with the other metrics (AAP and AF), we point the reader to the original paper introducing the setting.

\begin{table*}[ht]
\centering
\begin{small}
\begin{sc}
\begin{tabular}{lcccccc}
\toprule
Method    & CoraFull                                    & A. Computer                                 & Arxiv               & Reddit                                      & Roman E.                                    & Elliptic                                    \\ \midrule
EWC       & $28.10 {\scriptstyle \pm 2.76}$             & $14.86 {\scriptstyle \pm 6.00}$             & $4.81 {\scriptstyle \pm 0.08}$              & $4.33 {\scriptstyle \pm 2.77}$              & $8.85 {\scriptstyle \pm 0.05}$              & $51.08 {\scriptstyle \pm 1.10}$ \\
LwF       & $15.74 {\scriptstyle \pm 1.56}$             & $19.33 {\scriptstyle \pm 0.14}$             & $4.79 {\scriptstyle \pm 0.08}$              & $13.13 {\scriptstyle \pm 1.92}$             & $8.81 {\scriptstyle \pm 0.01}$              & $50.79 {\scriptstyle \pm 1.36}$             \\
MAS       & $8.40 {\scriptstyle \pm 0.62}$              & $12.68 {\scriptstyle \pm 8.28}$             & $3.35 {\scriptstyle \pm 0.99}$              & $10.21 {\scriptstyle \pm 1.03}$             & $11.02 {\scriptstyle \pm 1.36}$             & $51.08 {\scriptstyle \pm 1.10}$ \\
TWP       & $13.98 {\scriptstyle \pm 1.66}$             & $19.80 {\scriptstyle \pm 3.41}$             & $4.74 {\scriptstyle \pm 0.05}$              & $12.79 {\scriptstyle \pm 2.51}$             & $8.99 {\scriptstyle \pm 0.06}$              & $\mathbf{51.13 {\scriptstyle \pm 1.74}}$    \\
ER        & $20.65 {\scriptstyle \pm 2.33}$             & $38.53 {\scriptstyle \pm 2.93}$             & $23.43 {\scriptstyle \pm 1.65}$             & $22.34 {\scriptstyle \pm 2.46}$             & $10.43 {\scriptstyle \pm 0.20}$             & $43.94 {\scriptstyle \pm 0.52}$             \\
A-GEM     & $\mathbf{40.45 {\scriptstyle \pm 0.77}}$    & $38.49 {\scriptstyle \pm 2.80}$             & $17.16 {\scriptstyle \pm 1.45}$             & $\mathbf{58.08 {\scriptstyle \pm 8.04}}$    & $9.07 {\scriptstyle \pm 0.15}$              & $47.06 {\scriptstyle \pm 1.18}$             \\
PDGNN     & $38.48 {\scriptstyle \pm 1.15}$ & $68.91 {\scriptstyle \pm 0.33}$ & $\mathbf{35.86 {\scriptstyle \pm 1.20}}$    & $53.98 {\scriptstyle \pm 0.44}$ & $\mathbf{14.20 {\scriptstyle \pm 0.87}}$    & $49.91 {\scriptstyle \pm 1.44}$             \\
SSM-ER    & $23.23 {\scriptstyle \pm 2.69}$             & $\mathbf{70.45 {\scriptstyle \pm 3.66}}$    & $31.28 {\scriptstyle \pm 1.91}$ & $50.48 {\scriptstyle \pm 1.69}$             & $11.71 {\scriptstyle \pm 0.51}$ & $24.45 {\scriptstyle \pm 2.97}$             \\
SSM-A-GEM & $36.63 {\scriptstyle \pm 4.63}$             & $50.01 {\scriptstyle \pm 9.12}$             & $13.12 {\scriptstyle \pm 2.46}$             & $22.54 {\scriptstyle \pm 4.51}$             & $9.00 {\scriptstyle \pm 0.09}$              & $40.48 {\scriptstyle \pm 0.82}$             \\ \midrule
bare      & $12.59 {\scriptstyle \pm 0.82}$             & $20.51 {\scriptstyle \pm 4.26}$             & $4.74 {\scriptstyle \pm 0.08}$              & $12.59 {\scriptstyle \pm 2.59}$             & $8.78 {\scriptstyle \pm 0.10}$              & $51.28 {\scriptstyle \pm 2.37}$             \\
Joint     & $67.55 {\scriptstyle \pm 0.05}$             & $83.07 {\scriptstyle \pm 1.30}$             & $58.58 {\scriptstyle \pm 0.28}$             & $90.02 {\scriptstyle \pm 0.12}$             & $39.47 {\scriptstyle \pm 0.33}$             & $71.97 {\scriptstyle \pm 0.83}$             \\ \bottomrule
\end{tabular}
\end{sc}
\end{small}
\caption{AP results in the OCGL setting from \citep{ocgl} of CL strategies with trained GCN and joint offline training upper bound. In particular, TWP \citep{liu_overcoming_2021}, PDGNN \citep{Zhang_2024}, SSM-ER and SSM-A-GEM \citep{zhang_sparsified_2022} are graph-specific CL baselines. Best performing method for each dataset is highlighted in bold. For Arxiv class-incremental stream is considered.}
\label{tab:full_trained_results}
\end{table*}

\section{Performance plots}\label{app:plots}

\begin{figure*}[h]
\centering
\begin{subfigure}[b]{0.44\textwidth}
    \includegraphics[width=\textwidth]{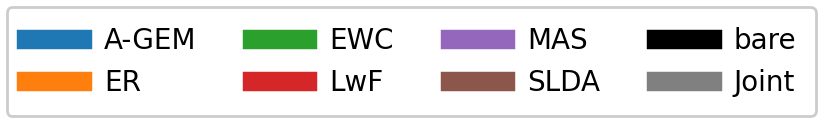}
    \label{fig:legend}
\end{subfigure}

\begin{subfigure}[b]{0.44\textwidth}
    \includegraphics[width=\textwidth]{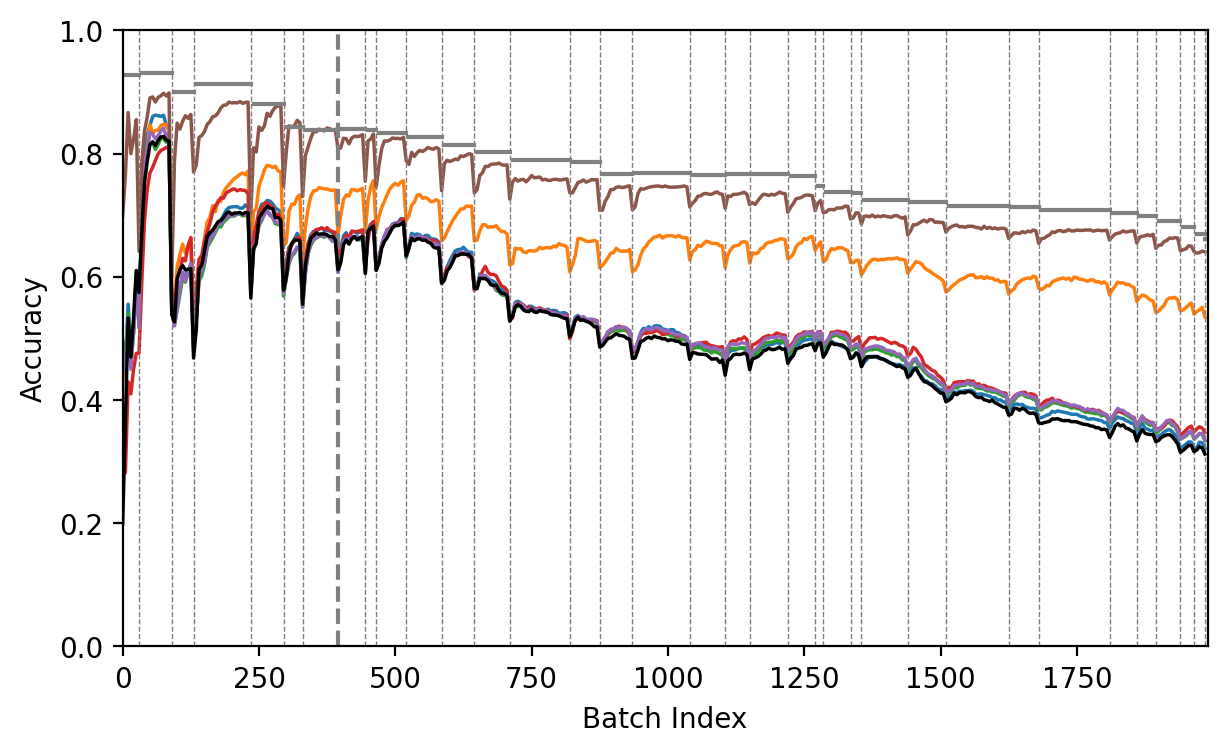}
    \caption{CoraFull}
    \label{fig:cora_ugcn}
\end{subfigure}
\hfill
\begin{subfigure}[b]{0.44\textwidth}
    \includegraphics[width=\textwidth]{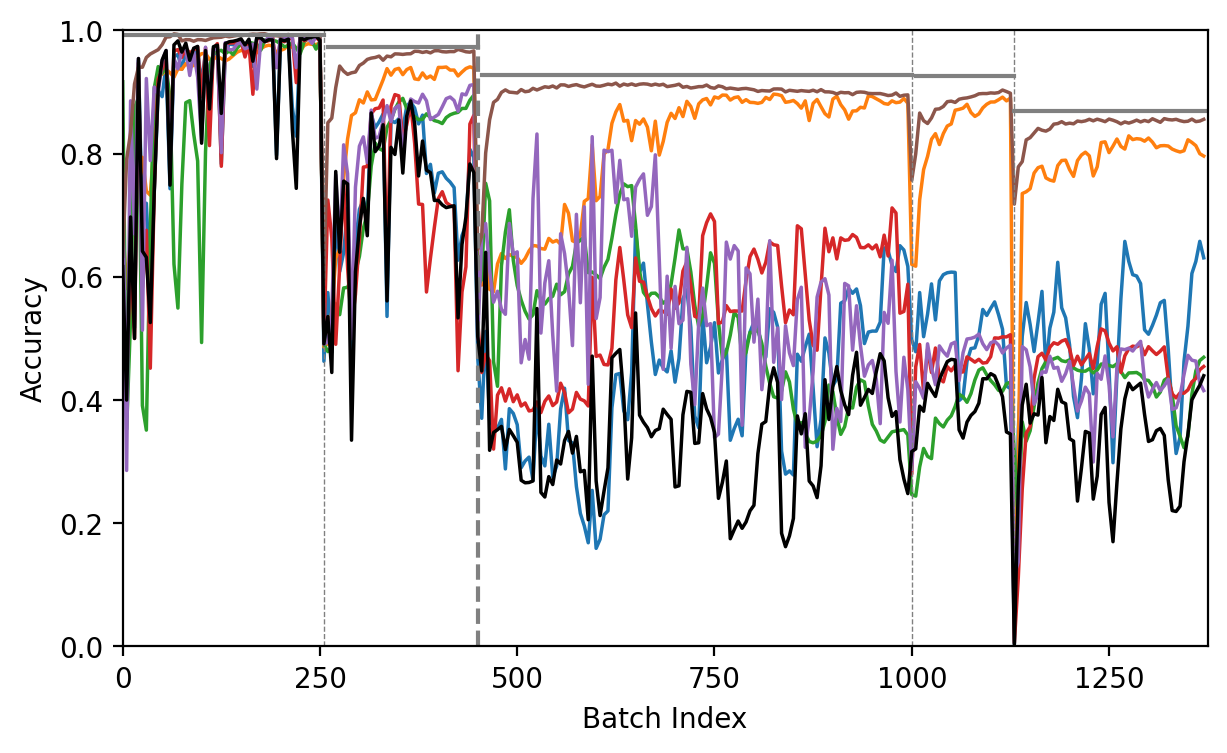}
    \caption{Amazon Computer}
    \label{fig:computer_ugcn}
\end{subfigure}

\begin{subfigure}[b]{0.44\textwidth}
    \includegraphics[width=\textwidth]{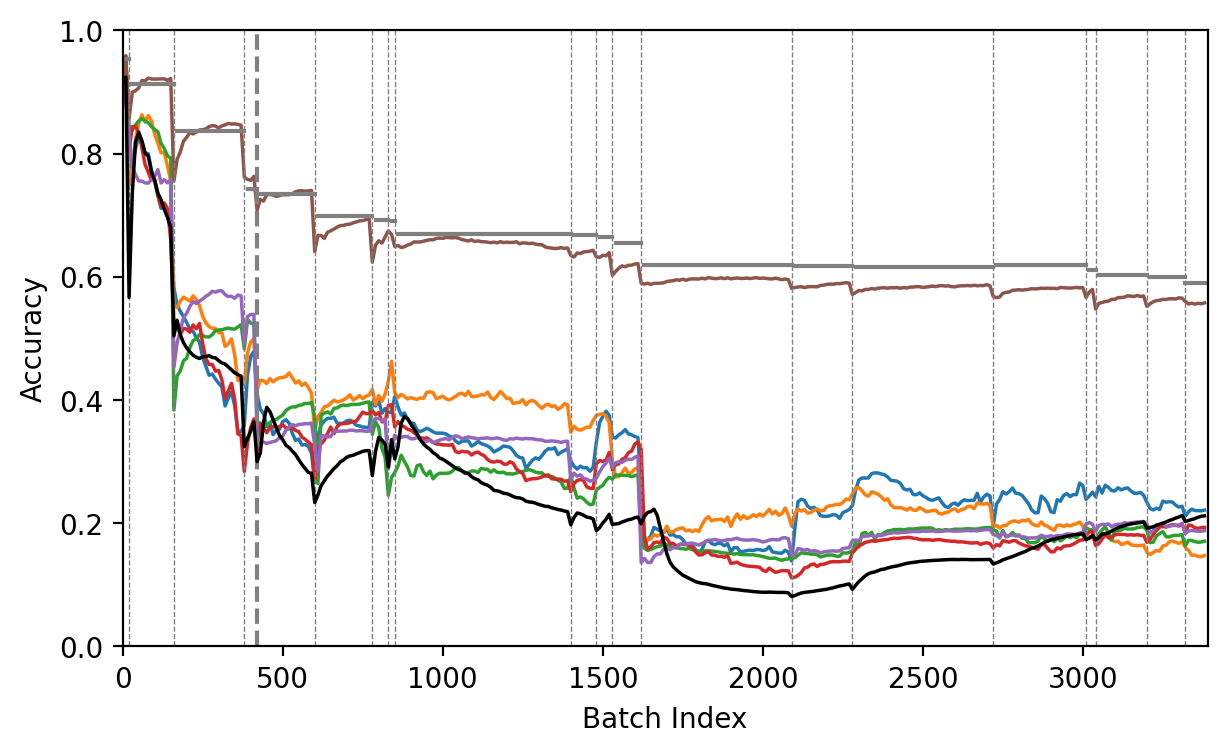}
    \caption{Arxiv}
    \label{fig:arxiv_ugcn}
\end{subfigure}
\hfill
\begin{subfigure}[b]{0.44\textwidth}
    \includegraphics[width=\textwidth]{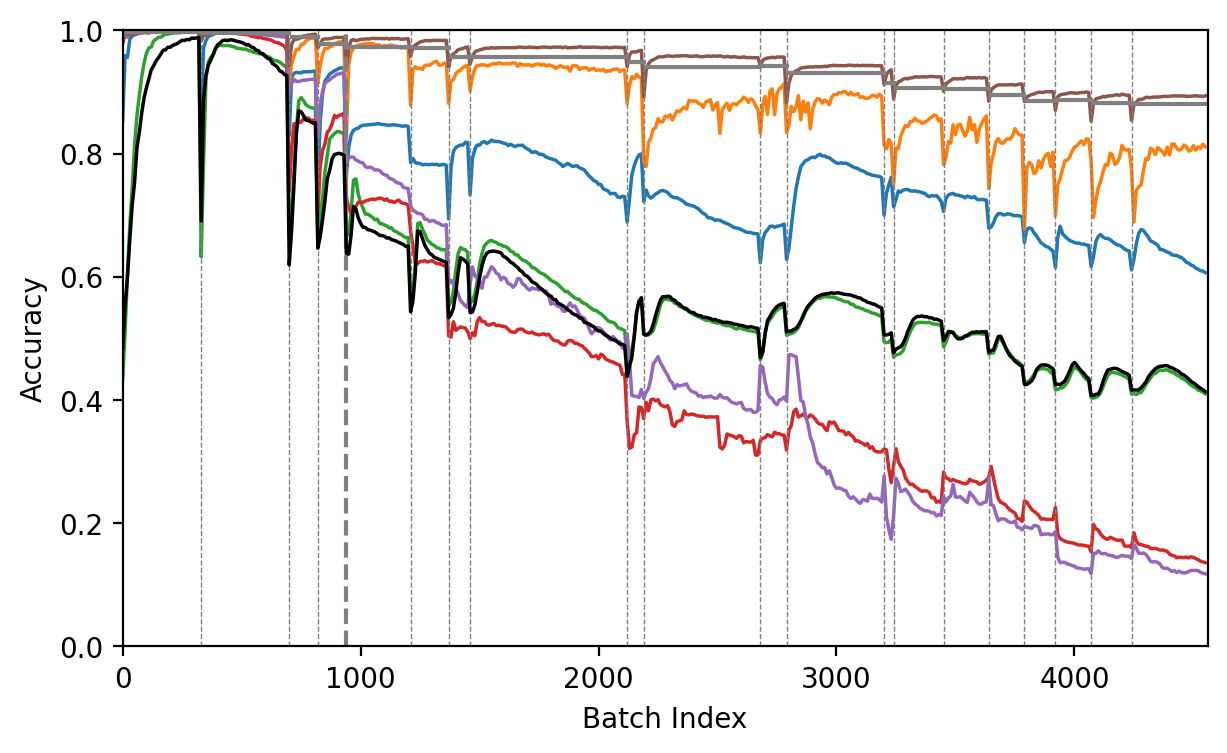}
    \caption{Reddit}
    \label{fig:reddit_ugcn}
\end{subfigure}

\begin{subfigure}[b]{0.44\textwidth}
    \includegraphics[width=\textwidth]{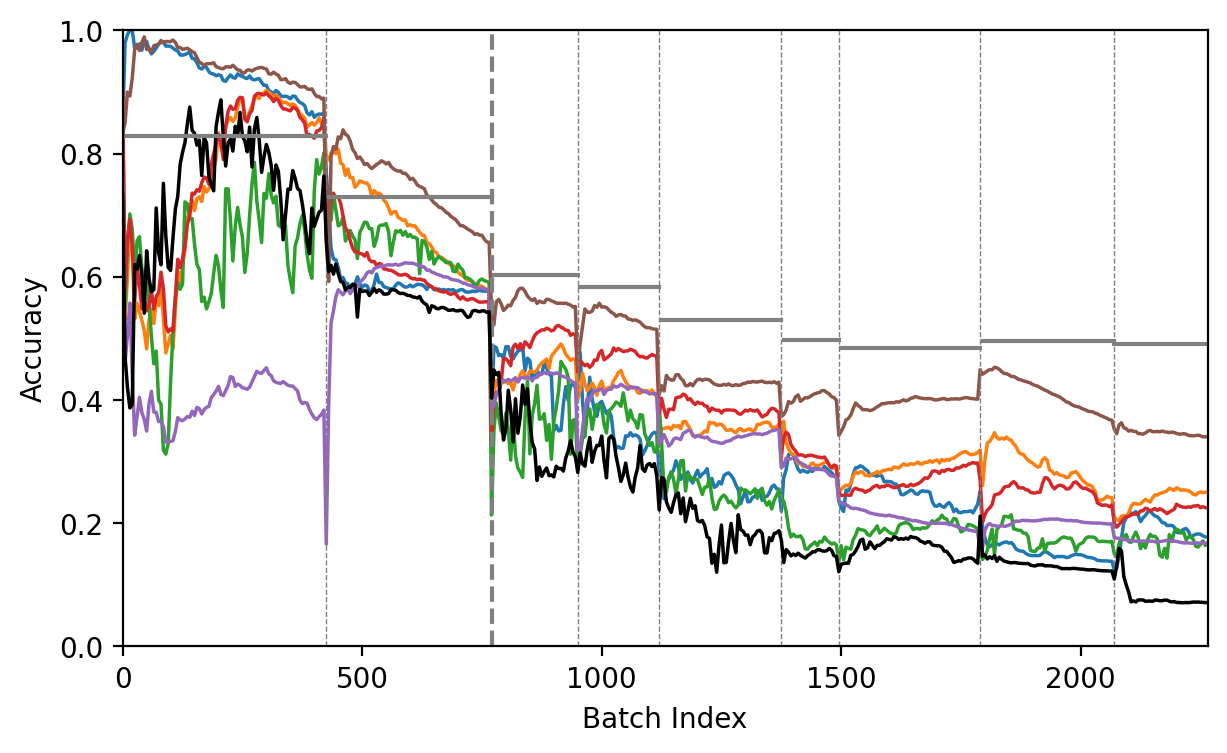}
    \caption{Roman Empire}
    \label{fig:roman_ugcn}
\end{subfigure}
\hfill
\begin{subfigure}[b]{0.44\textwidth}
    \includegraphics[width=\textwidth]{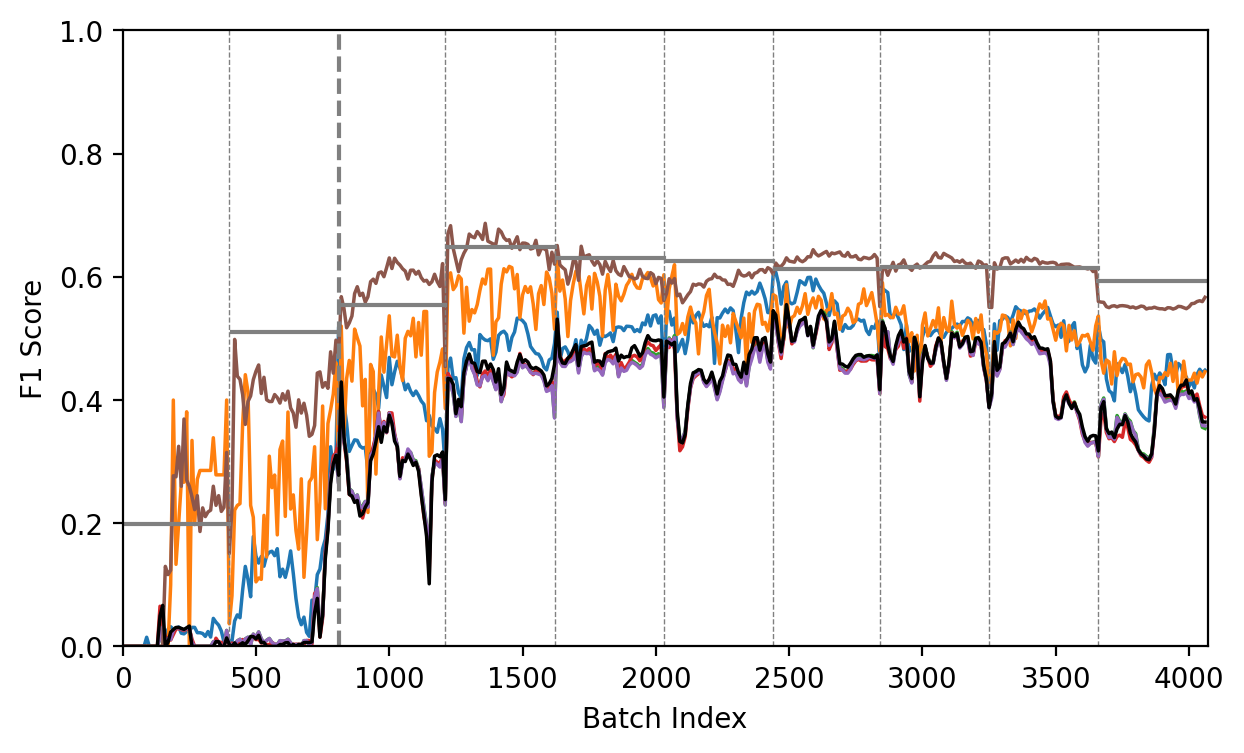}
    \caption{Elliptic (time-incremental)}
    \label{fig:elliptic_ugcn}
\end{subfigure}
\begin{subfigure}[b]{0.44\textwidth}
    \includegraphics[width=\textwidth]{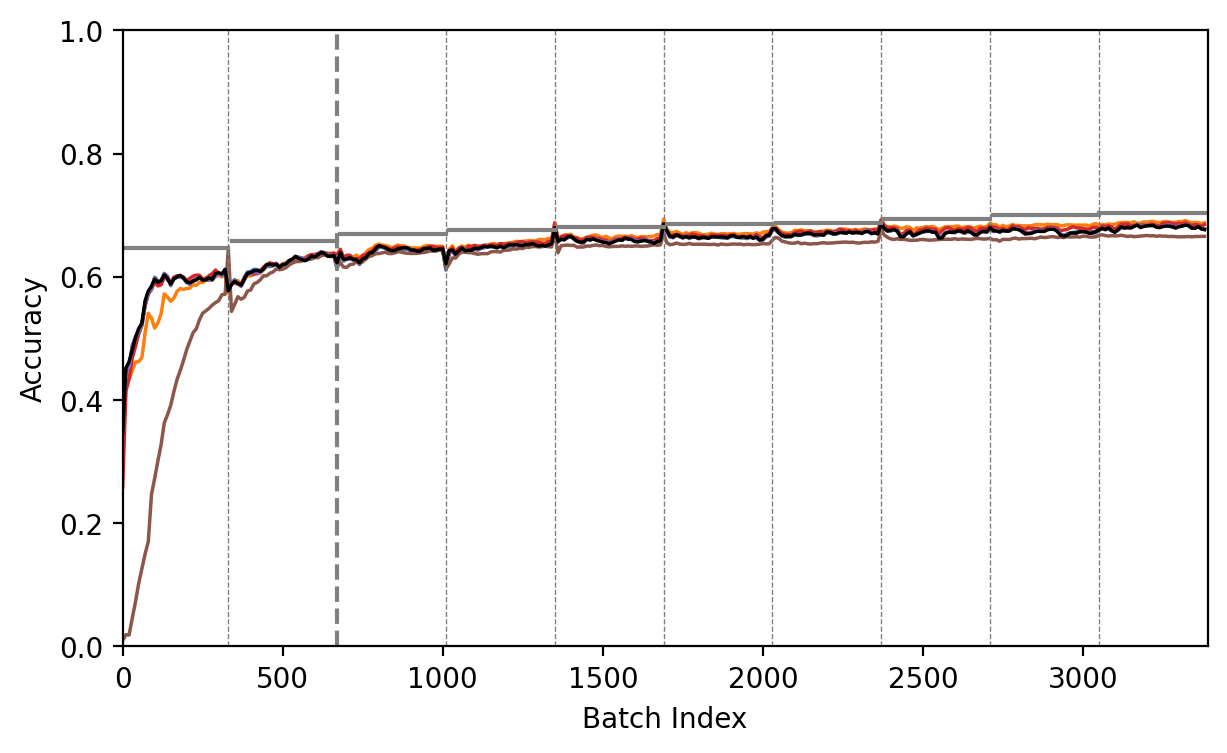}
    \caption{Arxiv (time-incremental)}
    \label{fig:arxivTI_ugcn}
\end{subfigure}
\caption{Anytime evaluation performance for the different datasets with UGCN Backbone. We highlight with vertical lines the task boundaries and the hyperparameter selection threshold.
}
\label{fig:anytime_evaluation_ugcn}
\end{figure*}

\begin{figure*}[h]
\centering
\begin{subfigure}[b]{0.4\textwidth}
    \includegraphics[width=\textwidth]{figures/legend.png}
    \label{fig:legend2}
\end{subfigure}

\begin{subfigure}[b]{0.44\textwidth}
    \includegraphics[width=\textwidth]{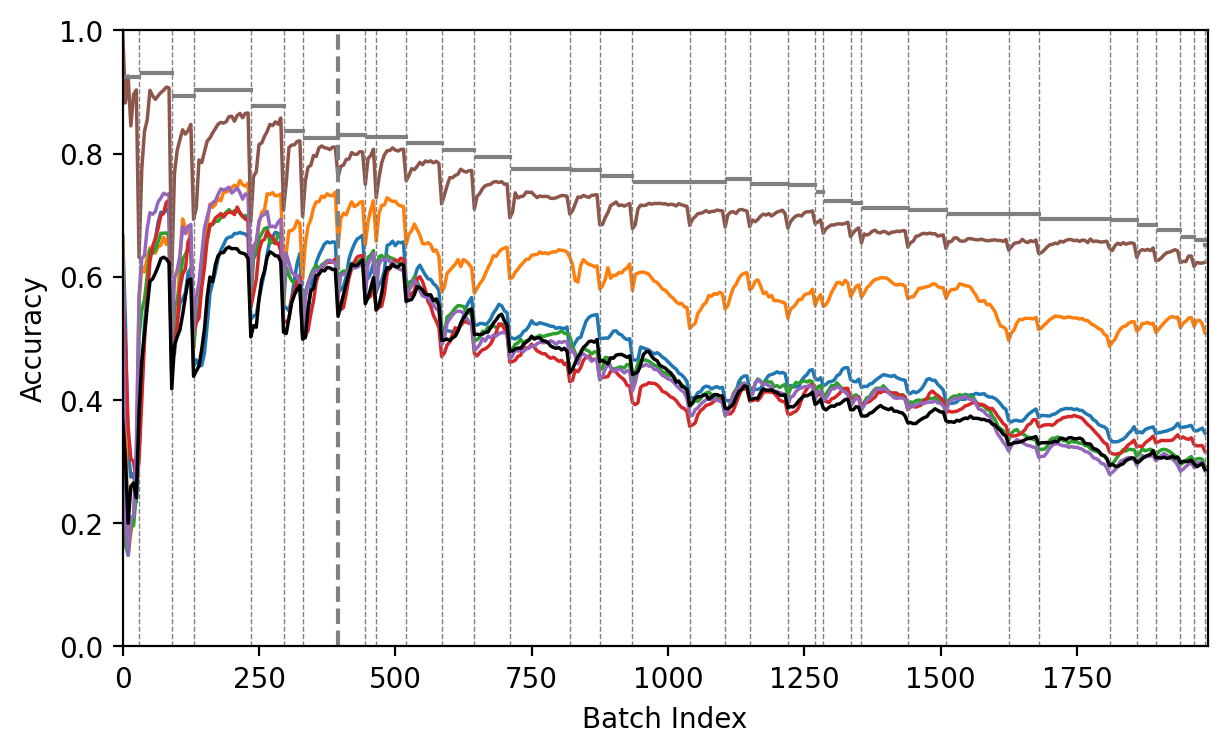}
    \caption{CoraFull}
    \label{fig:cora_grnf}
\end{subfigure}
\hfill
\begin{subfigure}[b]{0.44\textwidth}
    \includegraphics[width=\textwidth]{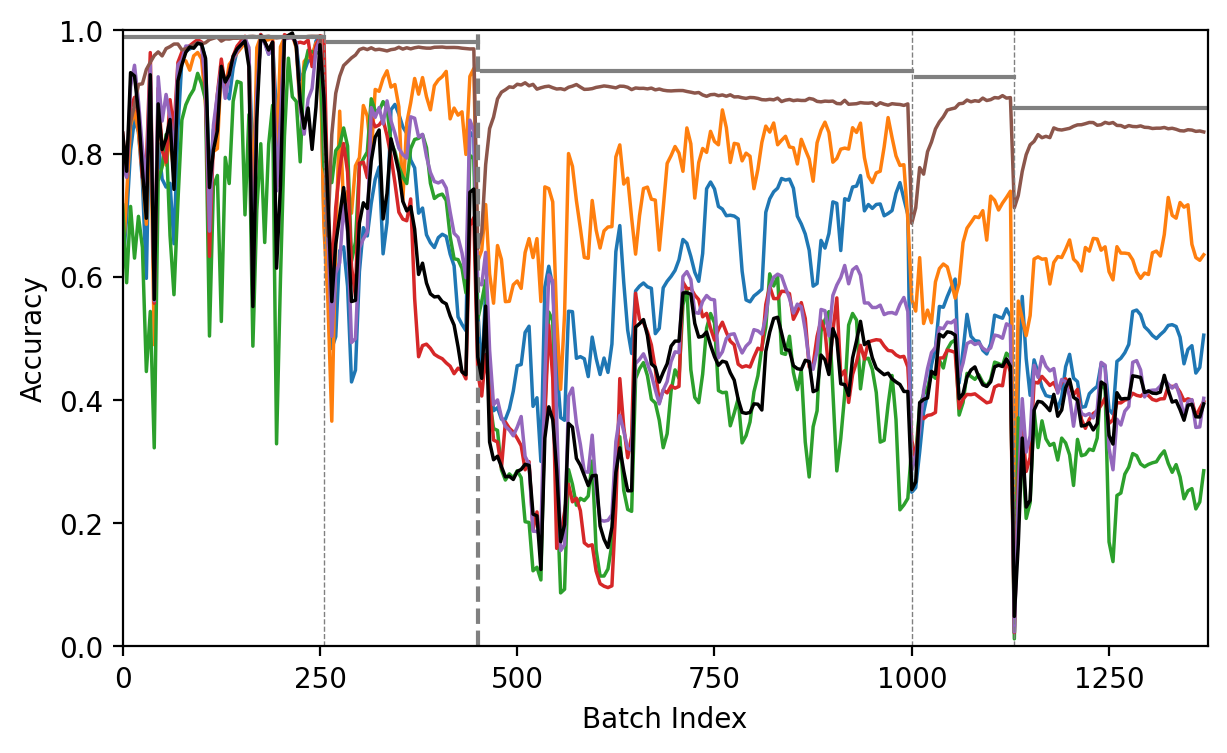}
    \caption{Amazon Computer}
    \label{fig:computer_grnf}
\end{subfigure}

\begin{subfigure}[b]{0.44\textwidth}
    \includegraphics[width=\textwidth]{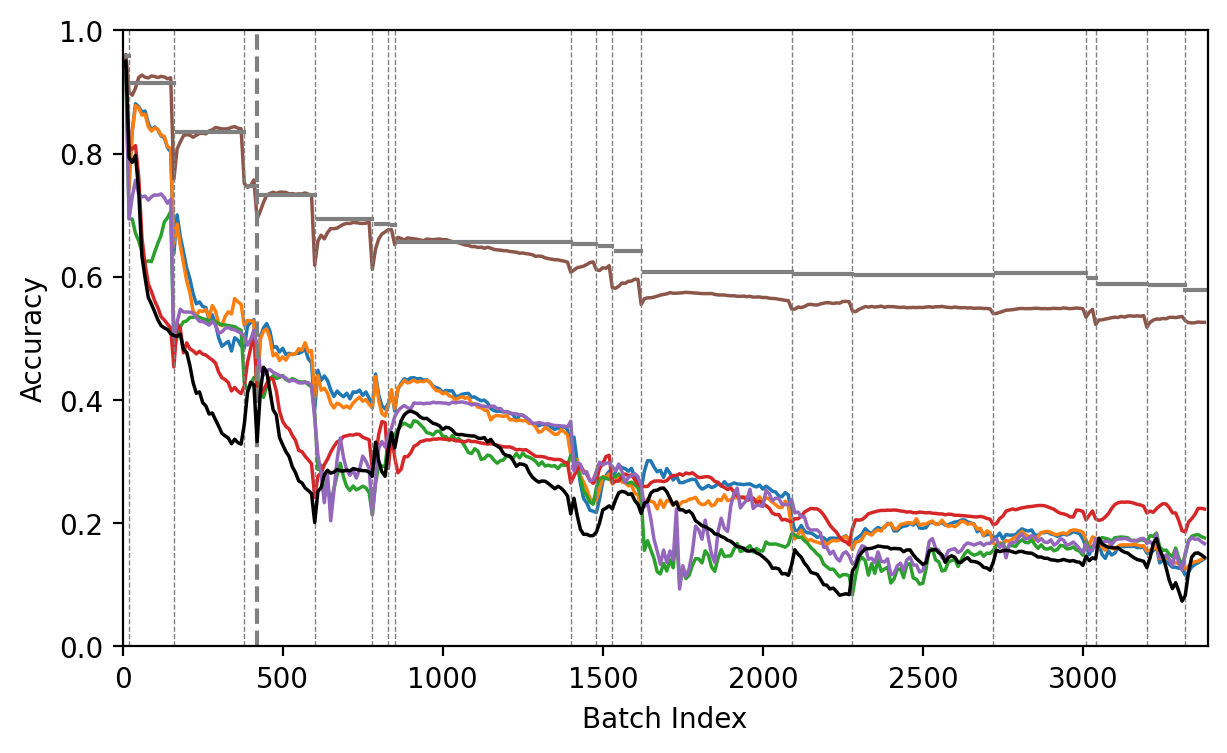}
    \caption{Arxiv}
    \label{fig:arxiv_grnf}
\end{subfigure}
\hfill
\begin{subfigure}[b]{0.44\textwidth}
    \includegraphics[width=\textwidth]{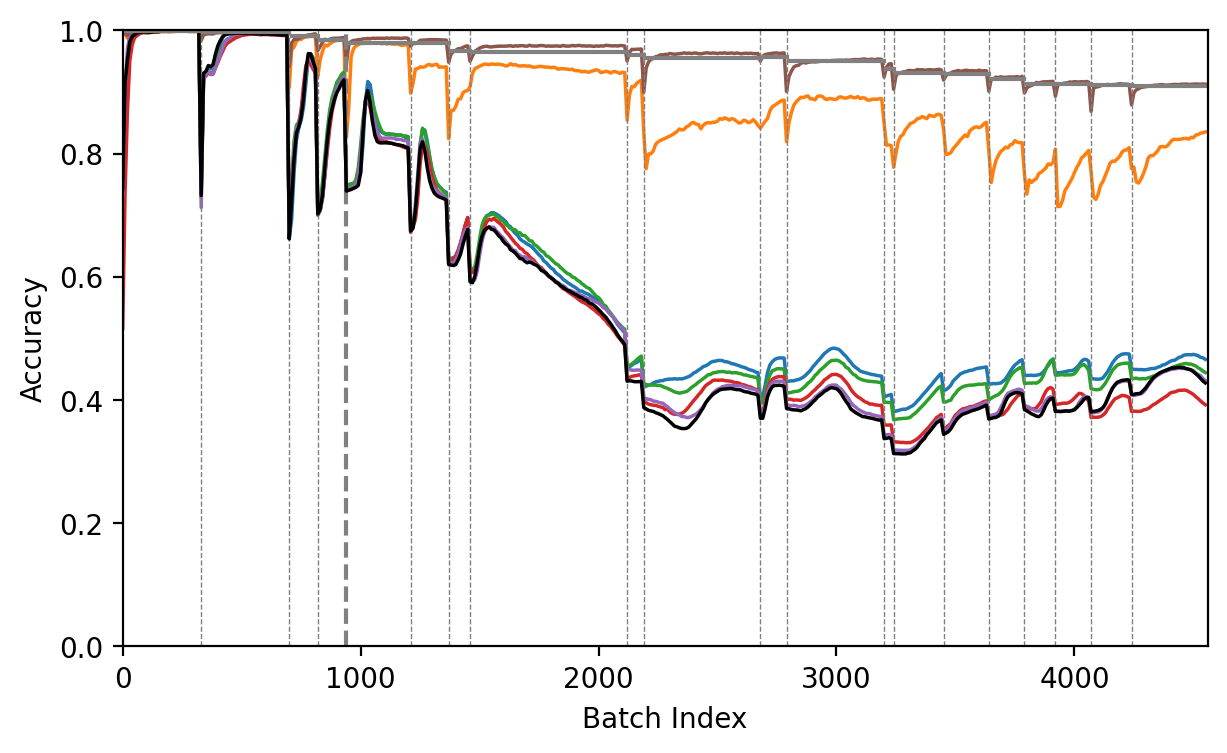}
    \caption{Reddit}
    \label{fig:reddit_grnf}
\end{subfigure}

\begin{subfigure}[b]{0.44\textwidth}
    \includegraphics[width=\textwidth]{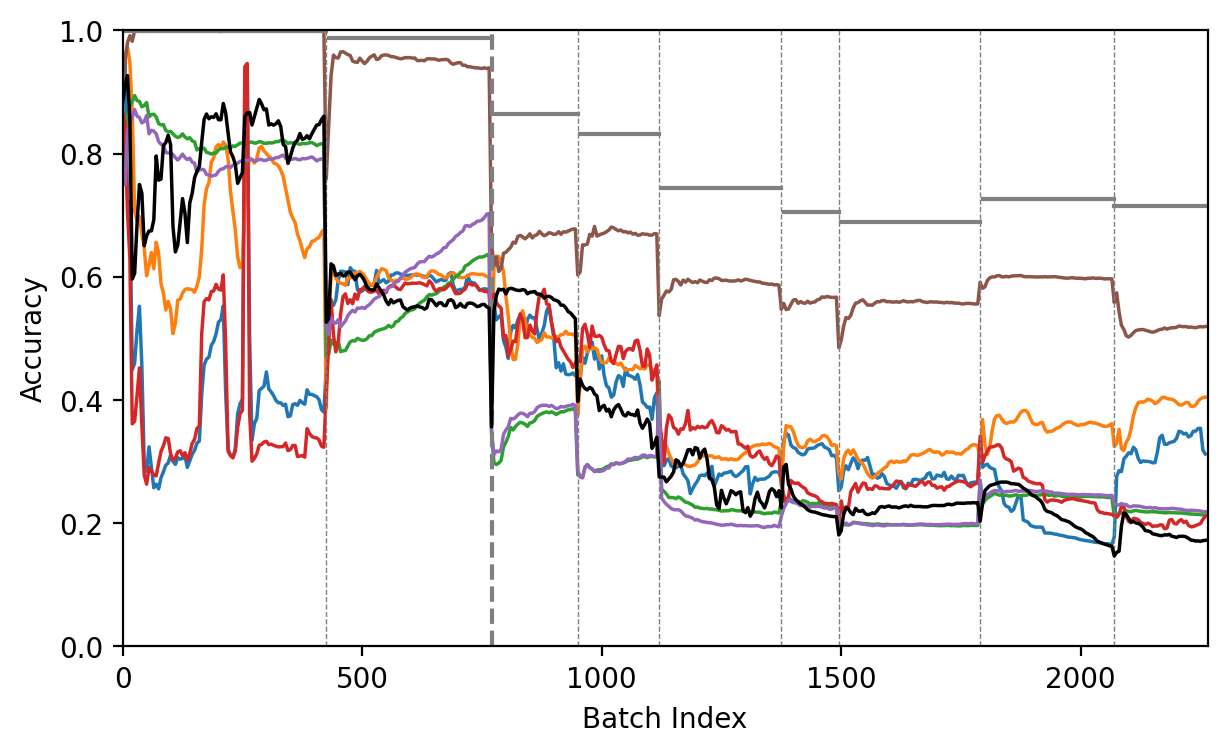}
    \caption{Roman Empire}
    \label{fig:roman_grnf}
\end{subfigure}
\hfill
\begin{subfigure}[b]{0.44\textwidth}
    \includegraphics[width=\textwidth]{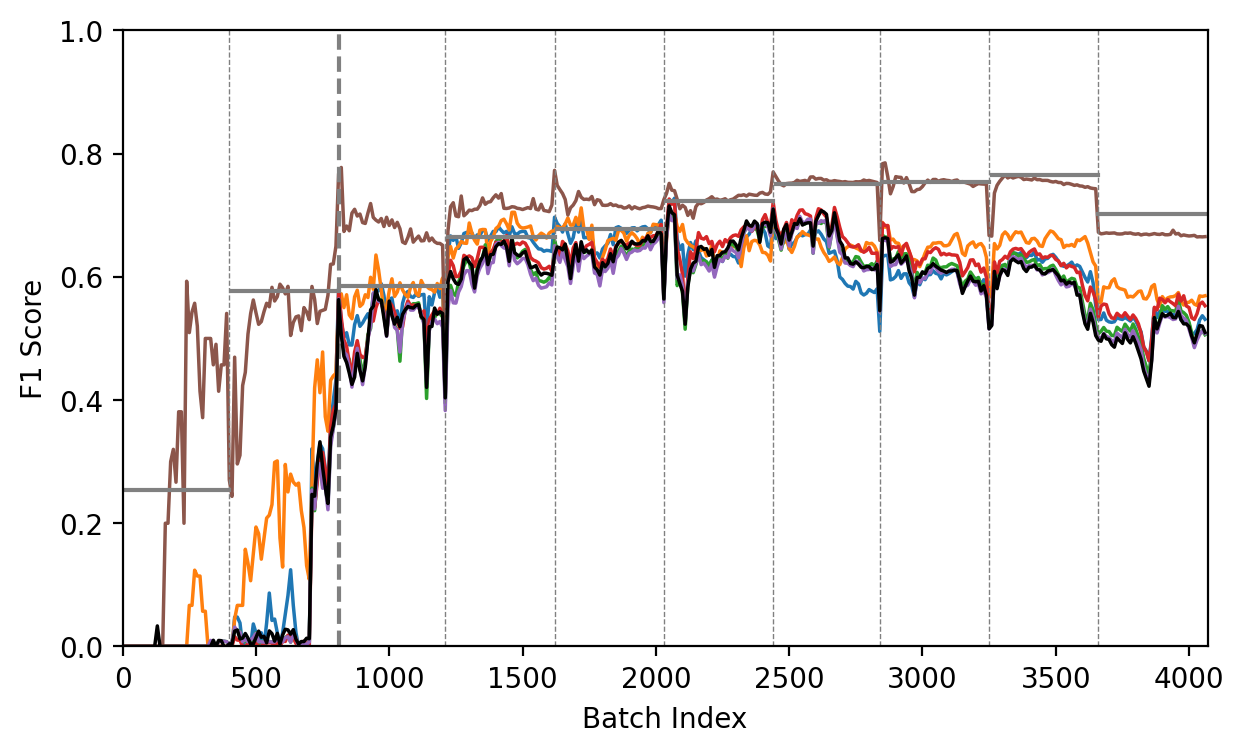}
    \caption{Elliptic (time-incremental)}
    \label{fig:elliptic_grnf}
\end{subfigure}
\begin{subfigure}[b]{0.44\textwidth}
    \includegraphics[width=\textwidth]{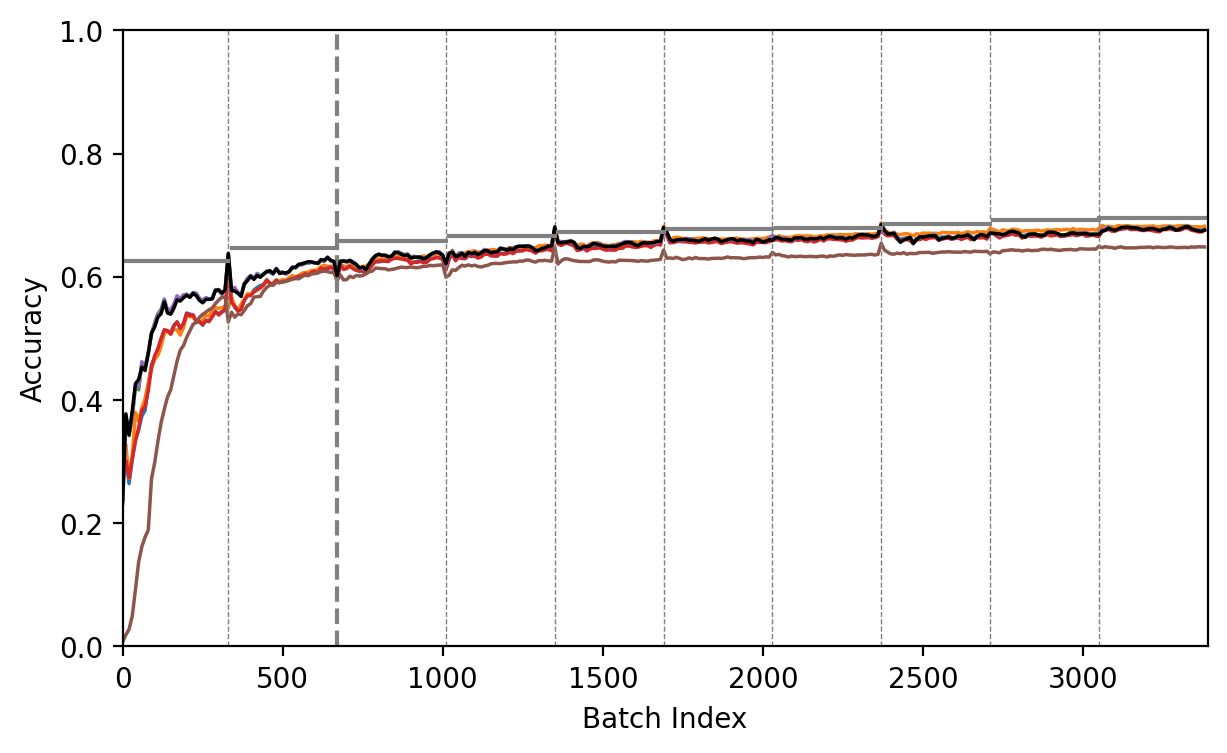}
    \caption{Arxiv (time-incremental)}
    \label{fig:arxivTI_grnf}
\end{subfigure}
\caption{Anytime evaluation performance for the different datasets with GRNF backbone. We highlight with vertical lines the task boundaries and the hyperparameter selection threshold.
}
\label{fig:anytime_evaluation_grnf}
\end{figure*}

We report here plots that show model performance along the node stream, to provide a more detailed understanding of the dynamics of training and forgetting. In Figures \ref{fig:anytime_evaluation_ugcn} and \ref{fig:anytime_evaluation_grnf} we plot for each benchmark a comparison of the performance using the considered methods, for UGCN and GRNF backbones respectively. We highlight the task boundaries with dotted vertical lines, with the thicker dashed one indicating the threshold at which hyperparameter selection is performed. The upper bound of joint training on data up to the present task is represented as an horizontal line over the batches of each task.

\begin{figure*}[]
\centering
\begin{subfigure}[b]{0.3\textwidth}
    \includegraphics[width=\textwidth]{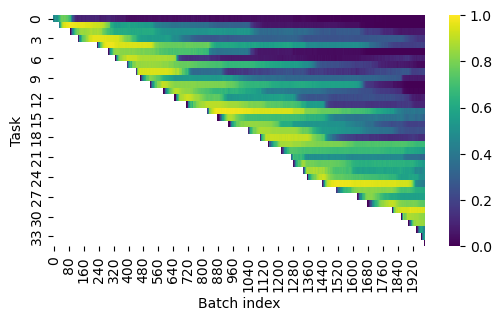}
    \caption{A-GEM}
    \label{fig:cora_ugcn_agem}
\end{subfigure}
\hfill
\begin{subfigure}[b]{0.3\textwidth}
    \includegraphics[width=\textwidth]{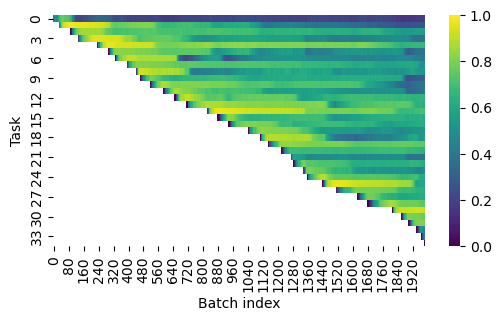}
    \caption{ER}
    \label{fig:cora_ugcn_er}
\end{subfigure}
\hfill
\begin{subfigure}[b]{0.3\textwidth}
    \includegraphics[width=\textwidth]{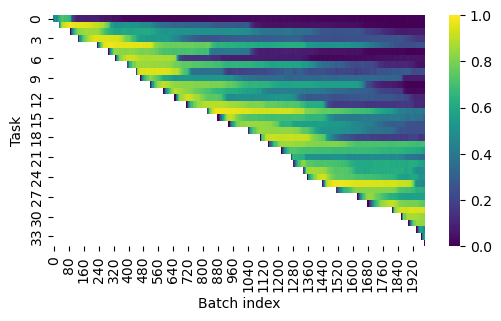}
    \caption{EWC}
    \label{fig:cora_ugcn_ewc}
\end{subfigure}
\begin{subfigure}[b]{0.3\textwidth}
    \includegraphics[width=\textwidth]{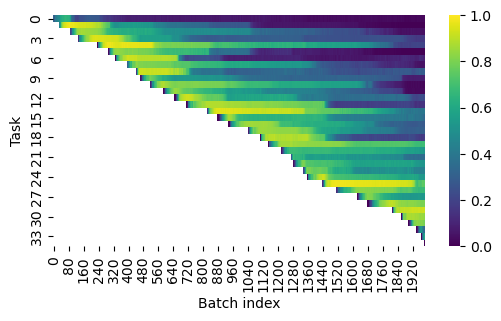}
    \caption{LwF}
    \label{fig:cora_ugcn_lwf}
\end{subfigure}
\hfill
\begin{subfigure}[b]{0.3\textwidth}
    \includegraphics[width=\textwidth]{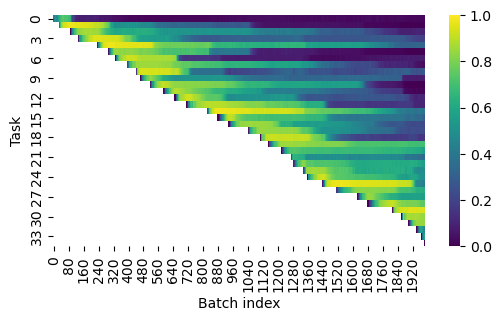}
    \caption{MAS}
    \label{fig:cora_ugcn_mas}
\end{subfigure}
\hfill
\begin{subfigure}[b]{0.3\textwidth}
    \includegraphics[width=\textwidth]{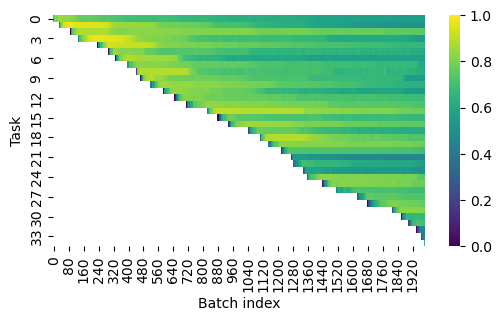}
    \caption{SLDA}
    \label{fig:cora_ugcn_slda}
\end{subfigure}
\caption{Anytime evaluation by task for the CoraFull dataset with UGCN backbone.}
\label{fig:heatmaps_cora_ugcn}
\end{figure*}

\begin{figure*}[]
\centering
\begin{subfigure}[b]{0.3\textwidth}
    \includegraphics[width=\textwidth]{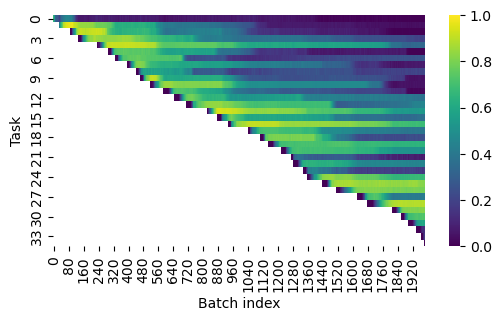}
    \caption{A-GEM}
    \label{fig:cora_grnf_agem}
\end{subfigure}
\hfill
\begin{subfigure}[b]{0.3\textwidth}
    \includegraphics[width=\textwidth]{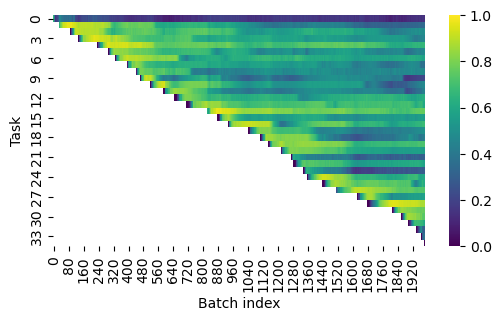}
    \caption{ER}
    \label{fig:cora_grnf_er}
\end{subfigure}
\hfill
\begin{subfigure}[b]{0.3\textwidth}
    \includegraphics[width=\textwidth]{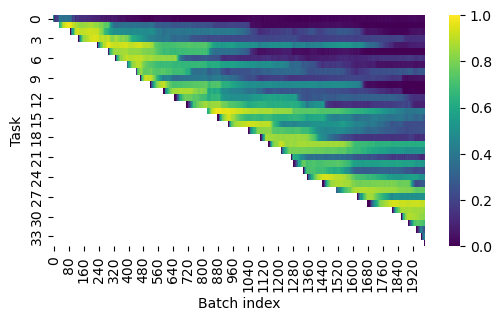}
    \caption{EWC}
    \label{fig:cora_grnf_ewc}
\end{subfigure}
\begin{subfigure}[b]{0.3\textwidth}
    \includegraphics[width=\textwidth]{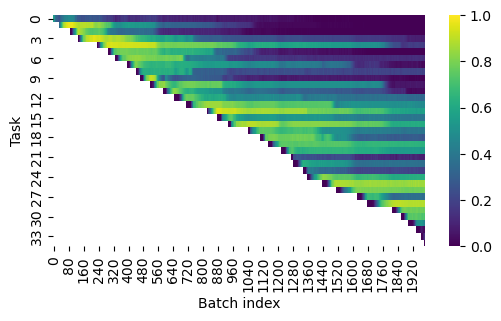}
    \caption{LwF}
    \label{fig:cora_grnf_lwf}
\end{subfigure}
\hfill
\begin{subfigure}[b]{0.3\textwidth}
    \includegraphics[width=\textwidth]{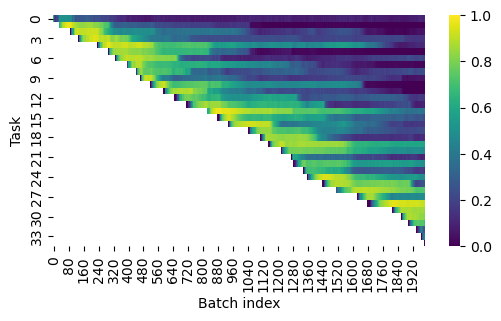}
    \caption{MAS}
    \label{fig:cora_grnf_mas}
\end{subfigure}
\hfill
\begin{subfigure}[b]{0.3\textwidth}
    \includegraphics[width=\textwidth]{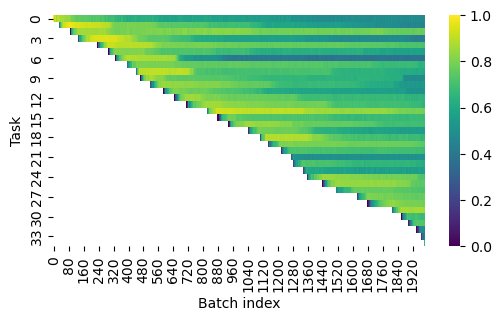}
    \caption{SLDA}
    \label{fig:cora_grnf_slda}
\end{subfigure}
\caption{Anytime evaluation by task for the CoraFull dataset with GRNF backbone.}
\label{fig:heatmaps_cora_grnf}
\end{figure*}

\begin{figure*}[]
\centering
\begin{subfigure}[b]{0.3\textwidth}
    \includegraphics[width=\textwidth]{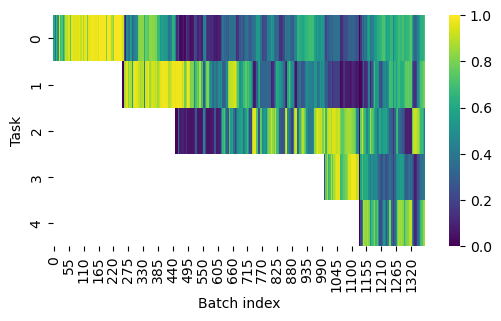}
    \caption{A-GEM}
    \label{fig:computer_ugcn_agem}
\end{subfigure}
\hfill
\begin{subfigure}[b]{0.3\textwidth}
    \includegraphics[width=\textwidth]{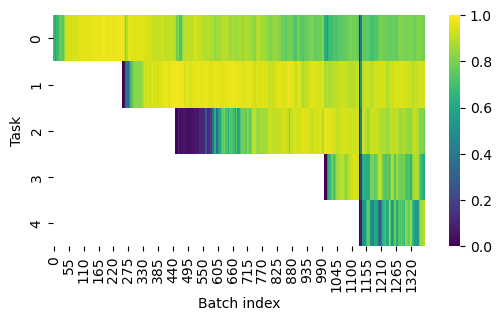}
    \caption{ER}
    \label{fig:computer_ugcn_er}
\end{subfigure}
\hfill
\begin{subfigure}[b]{0.3\textwidth}
    \includegraphics[width=\textwidth]{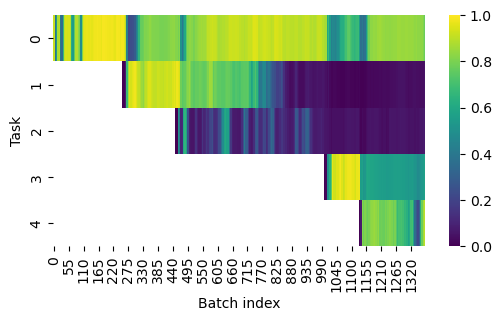}
    \caption{EWC}
    \label{fig:computer_ugcn_ewc}
\end{subfigure}
\begin{subfigure}[b]{0.3\textwidth}
    \includegraphics[width=\textwidth]{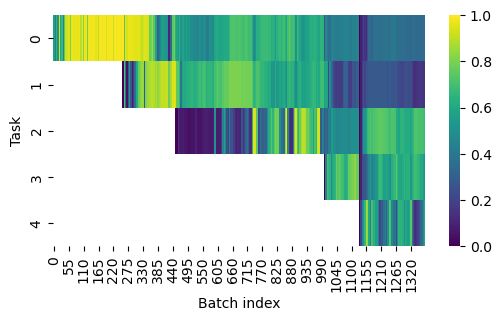}
    \caption{LwF}
    \label{fig:computer_ugcn_lwf}
\end{subfigure}
\hfill
\begin{subfigure}[b]{0.3\textwidth}
    \includegraphics[width=\textwidth]{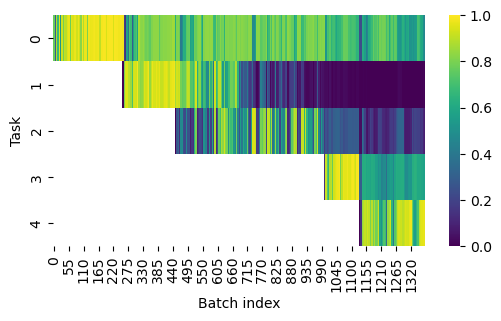}
    \caption{MAS}
    \label{fig:computer_ugcn_mas}
\end{subfigure}
\hfill
\begin{subfigure}[b]{0.3\textwidth}
    \includegraphics[width=\textwidth]{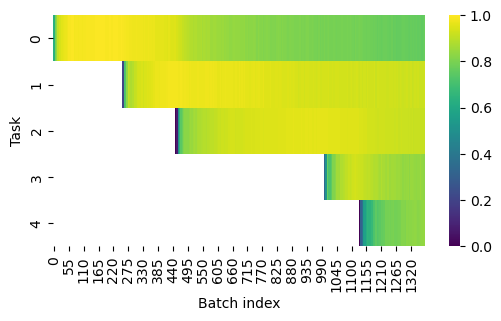}
    \caption{SLDA}
    \label{fig:computer_ugcn_slda}
\end{subfigure}
\caption{Anytime evaluation by task for the Amazon Computer dataset with UGCN backbone.}
\label{fig:heatmaps_computer_ugcn}
\end{figure*}

\begin{figure*}[]
\centering
\begin{subfigure}[b]{0.3\textwidth}
    \includegraphics[width=\textwidth]{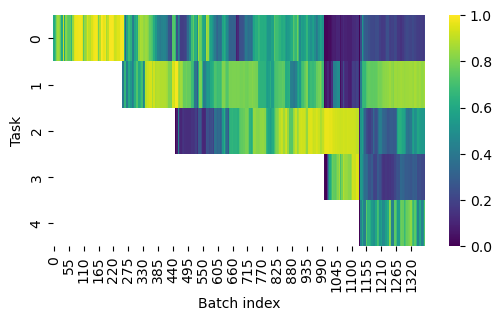}
    \caption{A-GEM}
    \label{fig:computer_grnf_agem}
\end{subfigure}
\hfill
\begin{subfigure}[b]{0.3\textwidth}
    \includegraphics[width=\textwidth]{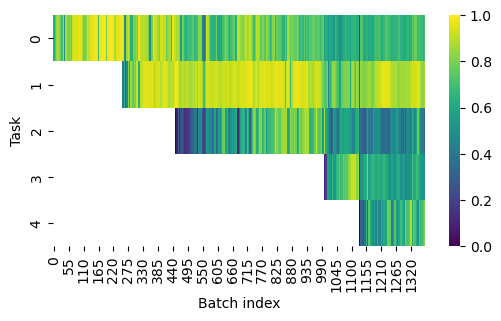}
    \caption{ER}
    \label{fig:computer_grnf_er}
\end{subfigure}
\hfill
\begin{subfigure}[b]{0.3\textwidth}
    \includegraphics[width=\textwidth]{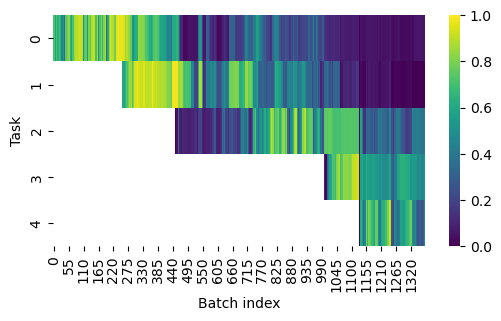}
    \caption{EWC}
    \label{fig:computer_grnf_ewc}
\end{subfigure}
\begin{subfigure}[b]{0.3\textwidth}
    \includegraphics[width=\textwidth]{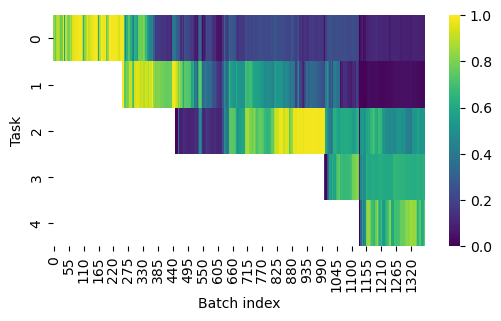}
    \caption{LwF}
    \label{fig:computer_grnf_lwf}
\end{subfigure}
\hfill
\begin{subfigure}[b]{0.3\textwidth}
    \includegraphics[width=\textwidth]{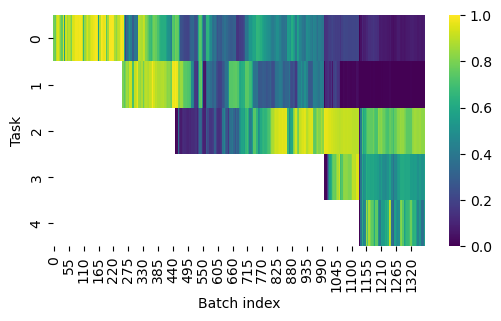}
    \caption{MAS}
    \label{fig:computer_grnf_mas}
\end{subfigure}
\hfill
\begin{subfigure}[b]{0.3\textwidth}
    \includegraphics[width=\textwidth]{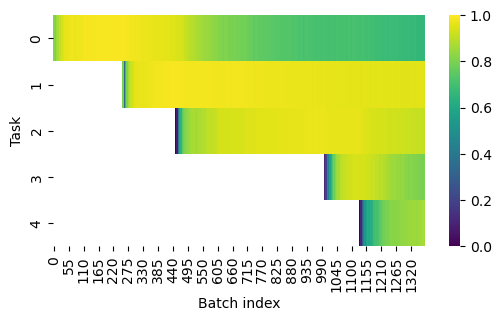}
    \caption{SLDA}
    \label{fig:computer_grnf_slda}
\end{subfigure}
\caption{Anytime evaluation by task for the Amazon Computer dataset with GRNF backbone.}
\label{fig:heatmaps_computer_grnf}
\end{figure*}

\begin{figure*}[]
\centering
\begin{subfigure}[b]{0.3\textwidth}
    \includegraphics[width=\textwidth]{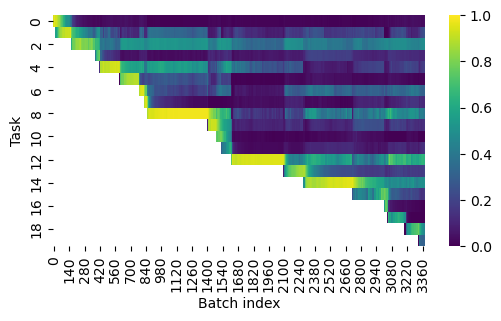}
    \caption{A-GEM}
    \label{fig:arxiv_ugcn_agem}
\end{subfigure}
\hfill
\begin{subfigure}[b]{0.3\textwidth}
    \includegraphics[width=\textwidth]{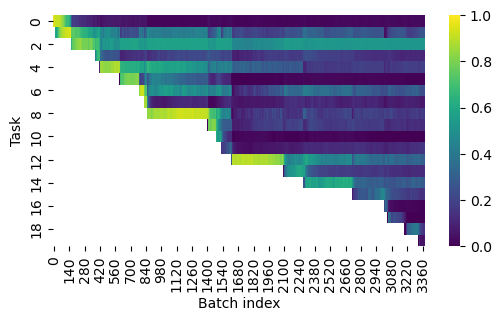}
    \caption{ER}
    \label{fig:arxiv_ugcn_er}
\end{subfigure}
\hfill
\begin{subfigure}[b]{0.3\textwidth}
    \includegraphics[width=\textwidth]{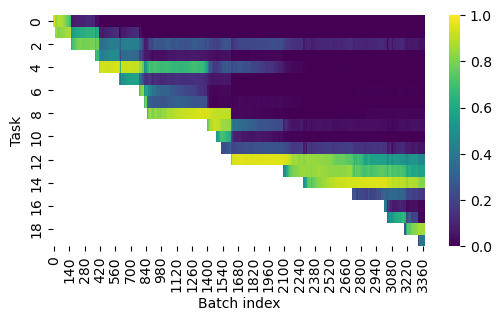}
    \caption{EWC}
    \label{fig:arxiv_ugcn_ewc}
\end{subfigure}
\begin{subfigure}[b]{0.3\textwidth}
    \includegraphics[width=\textwidth]{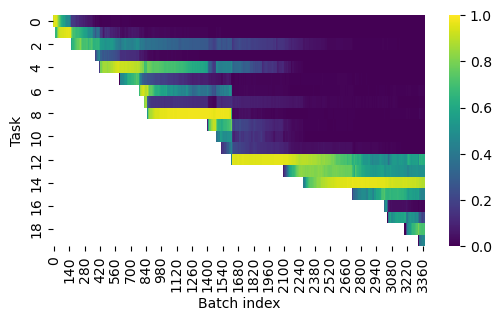}
    \caption{LwF}
    \label{fig:arxiv_ugcn_lwf}
\end{subfigure}
\hfill
\begin{subfigure}[b]{0.3\textwidth}
    \includegraphics[width=\textwidth]{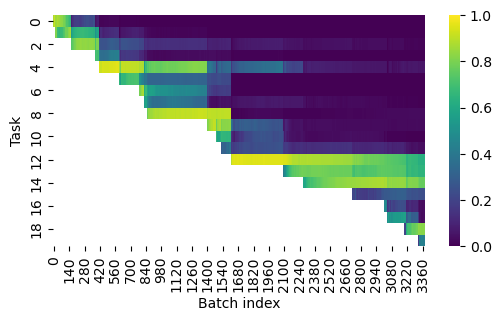}
    \caption{MAS}
    \label{fig:arxiv_ugcn_mas}
\end{subfigure}
\hfill
\begin{subfigure}[b]{0.3\textwidth}
    \includegraphics[width=\textwidth]{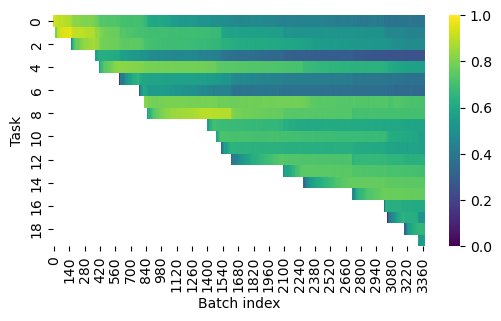}
    \caption{SLDA}
    \label{fig:arxiv_ugcn_slda}
\end{subfigure}
\caption{Anytime evaluation by task for the Arxiv dataset with UGCN backbone.}
\label{fig:heatmaps_arxiv_ugcn}
\end{figure*}

\begin{figure*}[]
\centering
\begin{subfigure}[b]{0.3\textwidth}
    \includegraphics[width=\textwidth]{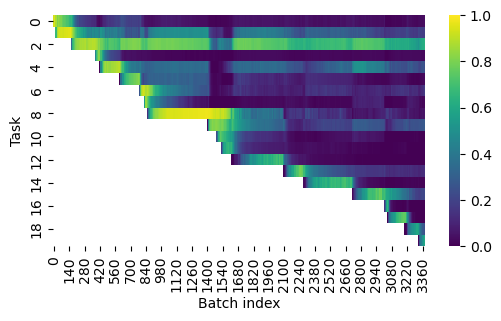}
    \caption{A-GEM}
    \label{fig:arxiv_grnf_agem}
\end{subfigure}
\hfill
\begin{subfigure}[b]{0.3\textwidth}
    \includegraphics[width=\textwidth]{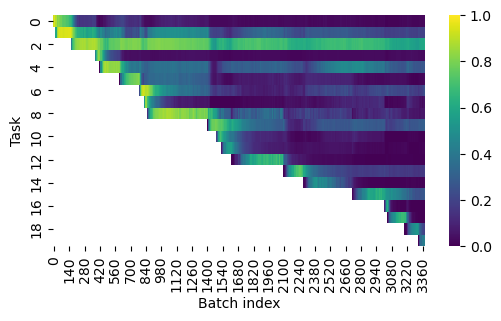}
    \caption{ER}
    \label{fig:arxiv_grnf_er}
\end{subfigure}
\hfill
\begin{subfigure}[b]{0.3\textwidth}
    \includegraphics[width=\textwidth]{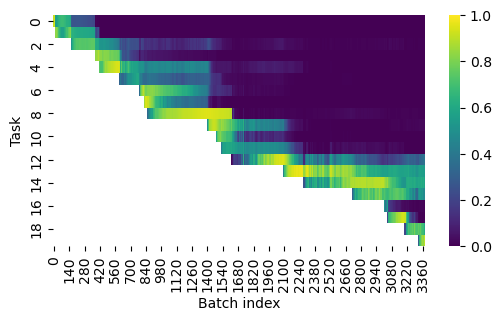}
    \caption{EWC}
    \label{fig:arxiv_grnf_ewc}
\end{subfigure}
\begin{subfigure}[b]{0.3\textwidth}
    \includegraphics[width=\textwidth]{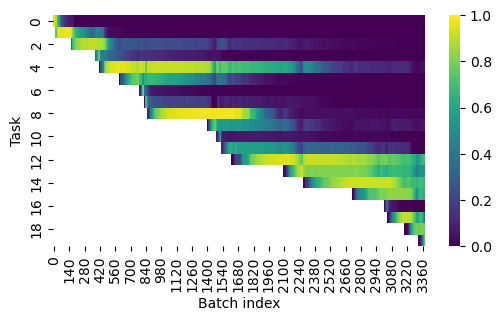}
    \caption{LwF}
    \label{fig:arxiv_grnf_lwf}
\end{subfigure}
\hfill
\begin{subfigure}[b]{0.3\textwidth}
    \includegraphics[width=\textwidth]{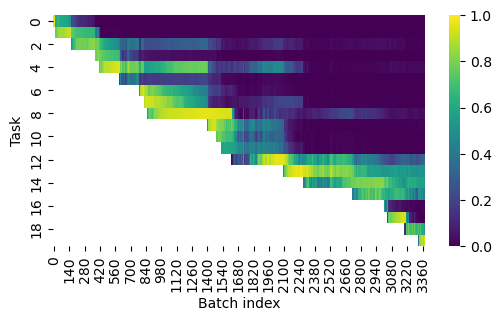}
    \caption{MAS}
    \label{fig:arxiv_grnf_mas}
\end{subfigure}
\hfill
\begin{subfigure}[b]{0.3\textwidth}
    \includegraphics[width=\textwidth]{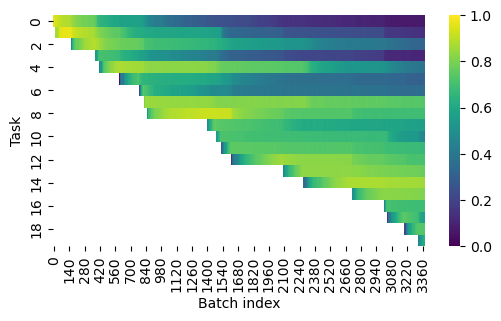}
    \caption{SLDA}
    \label{fig:arxiv_grnf_slda}
\end{subfigure}
\caption{Anytime evaluation by task for the Arxiv dataset with GRNF backbone.}
\label{fig:heatmaps_arxiv_grnf}
\end{figure*}

\begin{figure*}[]
\centering
\begin{subfigure}[b]{0.3\textwidth}
    \includegraphics[width=\textwidth]{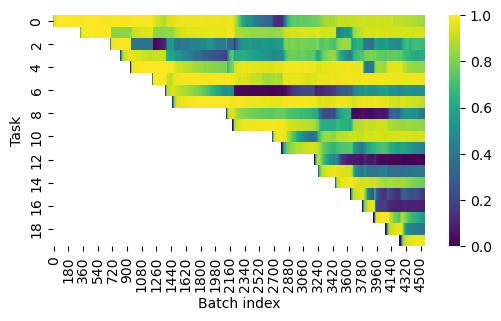}
    \caption{A-GEM}
    \label{fig:reddit_ugcn_agem}
\end{subfigure}
\hfill
\begin{subfigure}[b]{0.3\textwidth}
    \includegraphics[width=\textwidth]{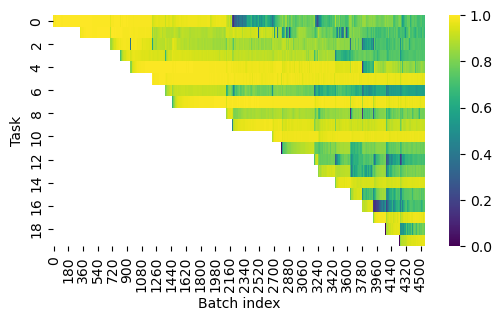}
    \caption{ER}
    \label{fig:reddit_ugcn_er}
\end{subfigure}
\hfill
\begin{subfigure}[b]{0.3\textwidth}
    \includegraphics[width=\textwidth]{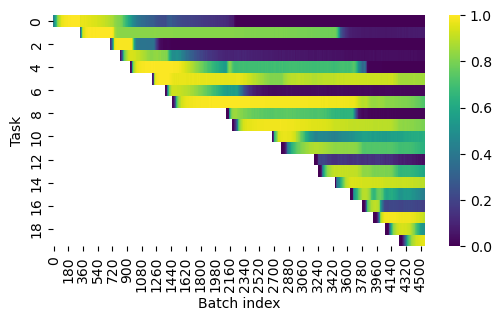}
    \caption{EWC}
    \label{fig:reddit_ugcn_ewc}
\end{subfigure}
\begin{subfigure}[b]{0.3\textwidth}
    \includegraphics[width=\textwidth]{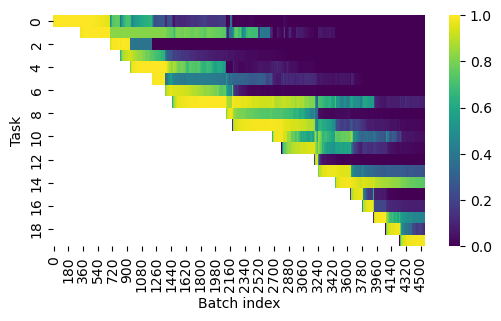}
    \caption{LwF}
    \label{fig:reddit_ugcn_lwf}
\end{subfigure}
\hfill
\begin{subfigure}[b]{0.3\textwidth}
    \includegraphics[width=\textwidth]{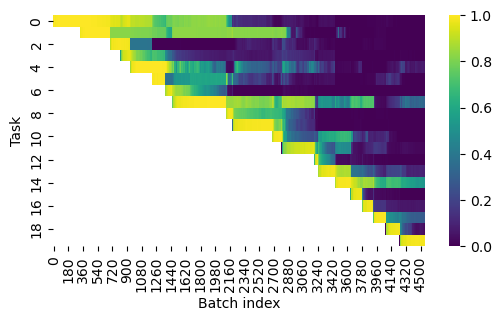}
    \caption{MAS}
    \label{fig:reddit_ugcn_mas}
\end{subfigure}
\hfill
\begin{subfigure}[b]{0.3\textwidth}
    \includegraphics[width=\textwidth]{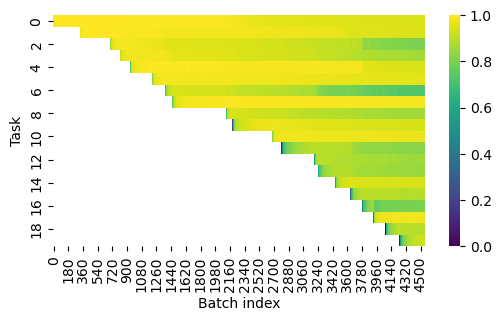}
    \caption{SLDA}
    \label{fig:reddit_ugcn_slda}
\end{subfigure}
\caption{Anytime evaluation by task for the Reddit dataset with UGCN backbone.}
\label{fig:heatmaps_reddit_ugcn}
\end{figure*}

\begin{figure*}[]
\centering
\begin{subfigure}[b]{0.3\textwidth}
    \includegraphics[width=\textwidth]{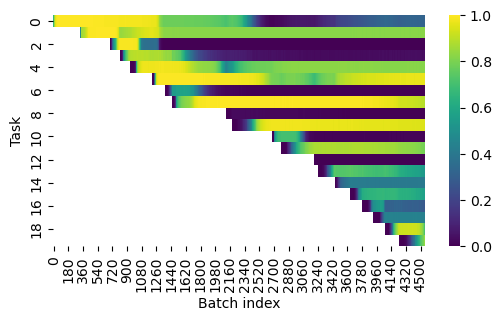}
    \caption{A-GEM}
    \label{fig:reddit_grnf_agem}
\end{subfigure}
\hfill
\begin{subfigure}[b]{0.3\textwidth}
    \includegraphics[width=\textwidth]{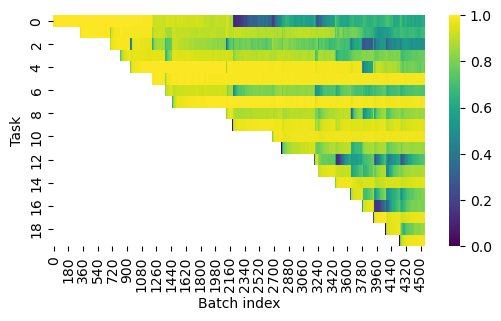}
    \caption{ER}
    \label{fig:reddit_grnf_er}
\end{subfigure}
\hfill
\begin{subfigure}[b]{0.3\textwidth}
    \includegraphics[width=\textwidth]{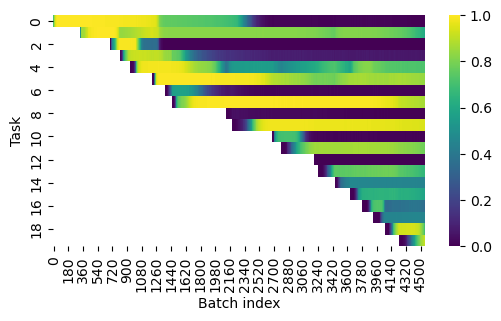}
    \caption{EWC}
    \label{fig:reddit_grnf_ewc}
\end{subfigure}
\begin{subfigure}[b]{0.3\textwidth}
    \includegraphics[width=\textwidth]{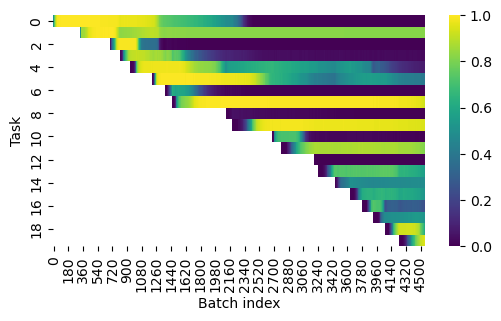}
    \caption{LwF}
    \label{fig:reddit_grnf_lwf}
\end{subfigure}
\hfill
\begin{subfigure}[b]{0.3\textwidth}
    \includegraphics[width=\textwidth]{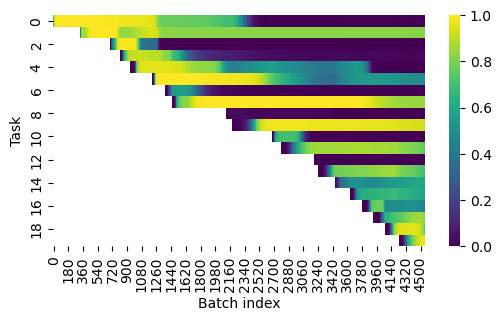}
    \caption{MAS}
    \label{fig:reddit_grnf_mas}
\end{subfigure}
\hfill
\begin{subfigure}[b]{0.3\textwidth}
    \includegraphics[width=\textwidth]{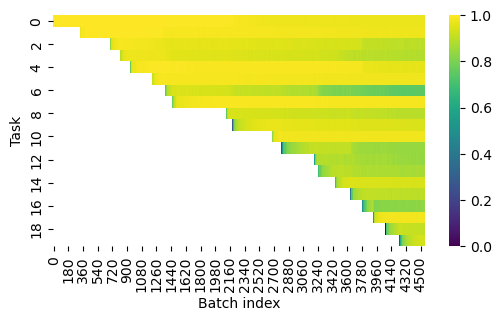}
    \caption{SLDA}
    \label{fig:reddit_grnf_slda}
\end{subfigure}
\caption{Anytime evaluation by task for the Reddit dataset with GRNF backbone.}
\label{fig:heatmaps_reddit_grnf}
\end{figure*}

\begin{figure*}[]
\centering
\begin{subfigure}[b]{0.3\textwidth}
    \includegraphics[width=\textwidth]{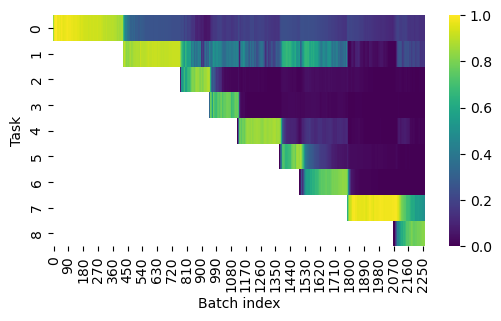}
    \caption{A-GEM}
    \label{fig:romanempire_ugcn_agem}
\end{subfigure}
\hfill
\begin{subfigure}[b]{0.3\textwidth}
    \includegraphics[width=\textwidth]{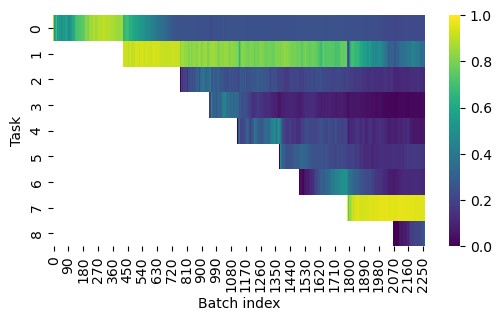}
    \caption{ER}
    \label{fig:romanempire_ugcn_er}
\end{subfigure}
\hfill
\begin{subfigure}[b]{0.3\textwidth}
    \includegraphics[width=\textwidth]{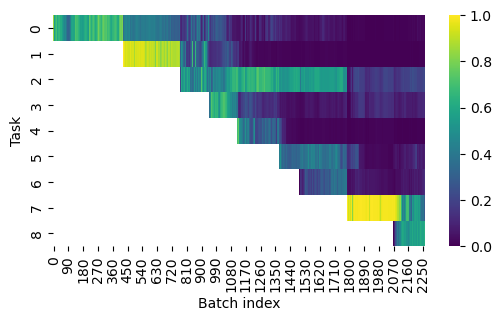}
    \caption{EWC}
    \label{fig:romanempire_ugcn_ewc}
\end{subfigure}
\begin{subfigure}[b]{0.3\textwidth}
    \includegraphics[width=\textwidth]{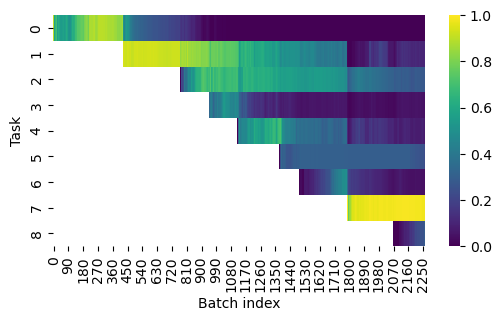}
    \caption{LwF}
    \label{fig:romanempire_ugcn_lwf}
\end{subfigure}
\hfill
\begin{subfigure}[b]{0.3\textwidth}
    \includegraphics[width=\textwidth]{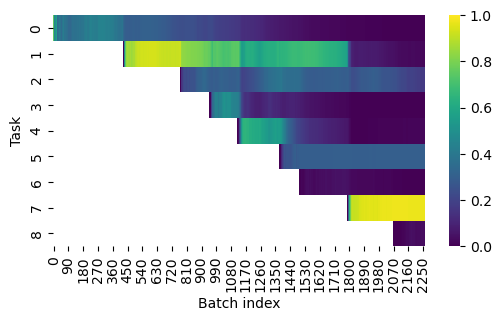}
    \caption{MAS}
    \label{fig:romanempire_ugcn_mas}
\end{subfigure}
\hfill
\begin{subfigure}[b]{0.3\textwidth}
    \includegraphics[width=\textwidth]{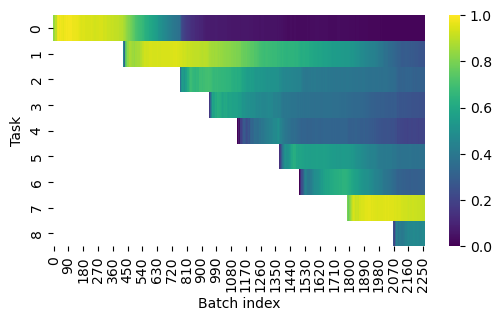}
    \caption{SLDA}
    \label{fig:romanempire_ugcn_slda}
\end{subfigure}
\caption{Anytime evaluation by task for the Roman Empire dataset with UGCN backbone.}
\label{fig:heatmaps_romanempire_ugcn}
\end{figure*}

\begin{figure*}[]
\centering
\begin{subfigure}[b]{0.3\textwidth}
    \includegraphics[width=\textwidth]{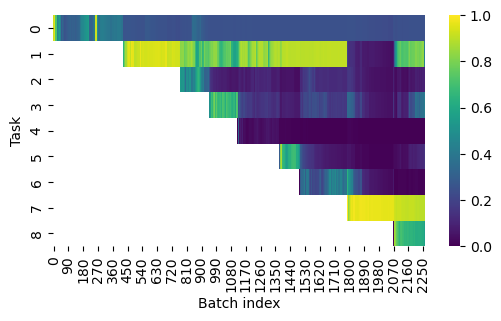}
    \caption{A-GEM}
    \label{fig:romanempire_grnf_agem}
\end{subfigure}
\hfill
\begin{subfigure}[b]{0.3\textwidth}
    \includegraphics[width=\textwidth]{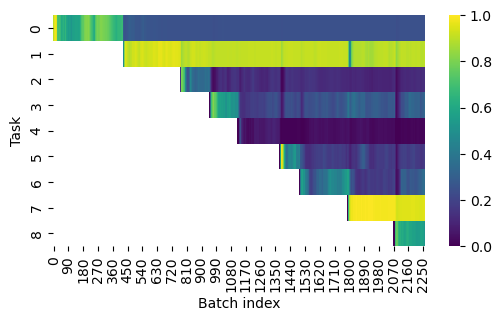}
    \caption{ER}
    \label{fig:romanempire_grnf_er}
\end{subfigure}
\hfill
\begin{subfigure}[b]{0.3\textwidth}
    \includegraphics[width=\textwidth]{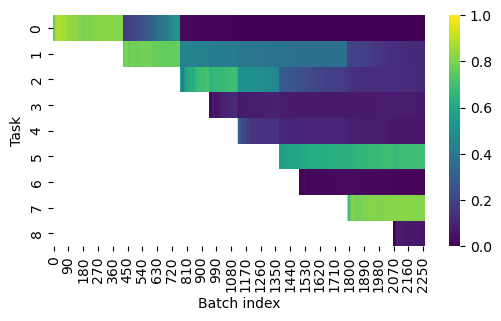}
    \caption{EWC}
    \label{fig:romanempire_grnf_ewc}
\end{subfigure}
\begin{subfigure}[b]{0.3\textwidth}
    \includegraphics[width=\textwidth]{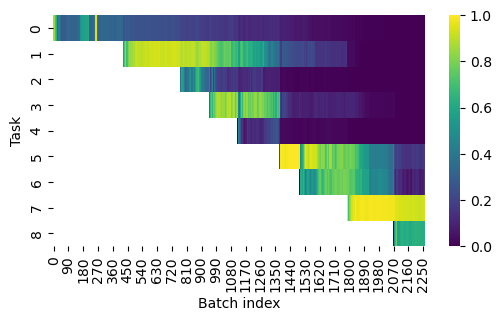}
    \caption{LwF}
    \label{fig:romanempire_grnf_lwf}
\end{subfigure}
\hfill
\begin{subfigure}[b]{0.3\textwidth}
    \includegraphics[width=\textwidth]{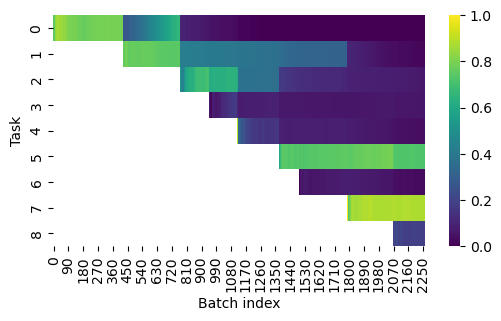}
    \caption{MAS}
    \label{fig:romanempire_grnf_mas}
\end{subfigure}
\hfill
\begin{subfigure}[b]{0.3\textwidth}
    \includegraphics[width=\textwidth]{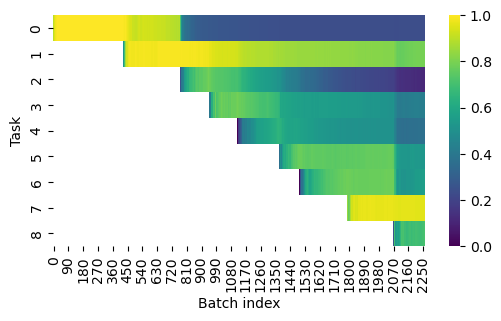}
    \caption{SLDA}
    \label{fig:romanempire_grnf_slda}
\end{subfigure}
\caption{Anytime evaluation by task for the Roman Empire dataset with GRNF backbone.}
\label{fig:heatmaps_romanempire_grnf}
\end{figure*}

\begin{figure*}[]
\centering
\begin{subfigure}[b]{0.3\textwidth}
    \includegraphics[width=\textwidth]{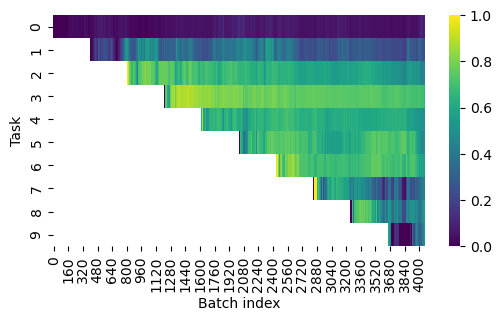}
    \caption{A-GEM}
    \label{fig:elliptic_ugcn_agem}
\end{subfigure}
\hfill
\begin{subfigure}[b]{0.3\textwidth}
    \includegraphics[width=\textwidth]{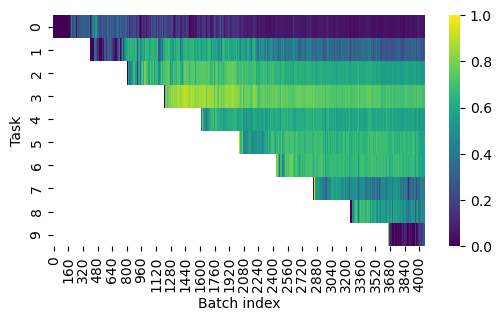}
    \caption{ER}
    \label{fig:elliptic_ugcn_er}
\end{subfigure}
\hfill
\begin{subfigure}[b]{0.3\textwidth}
    \includegraphics[width=\textwidth]{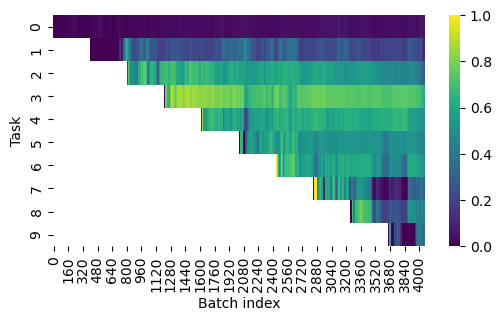}
    \caption{EWC}
    \label{fig:elliptic_ugcn_ewc}
\end{subfigure}
\begin{subfigure}[b]{0.3\textwidth}
    \includegraphics[width=\textwidth]{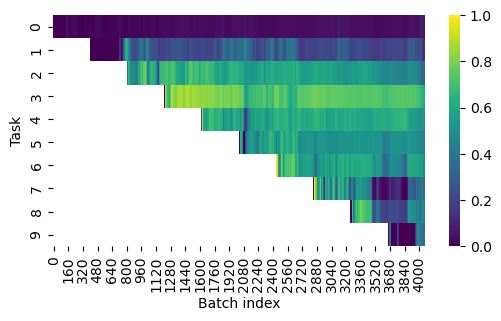}
    \caption{LwF}
    \label{fig:elliptic_ugcn_lwf}
\end{subfigure}
\hfill
\begin{subfigure}[b]{0.3\textwidth}
    \includegraphics[width=\textwidth]{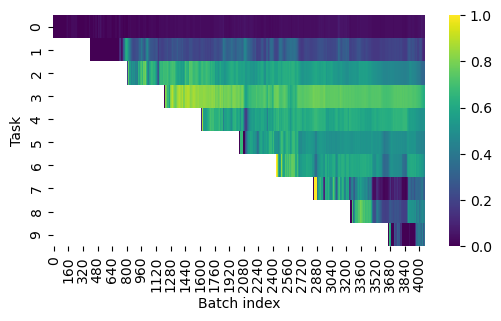}
    \caption{MAS}
    \label{fig:elliptic_ugcn_mas}
\end{subfigure}
\hfill
\begin{subfigure}[b]{0.3\textwidth}
    \includegraphics[width=\textwidth]{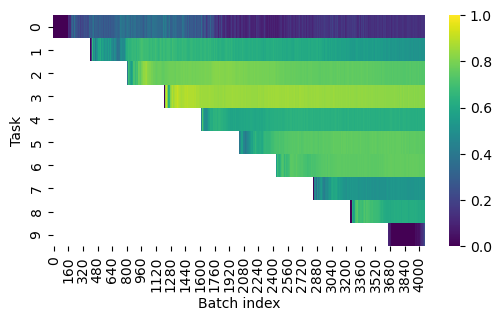}
    \caption{SLDA}
    \label{fig:elliptic_ugcn_slda}
\end{subfigure}
\caption{Anytime evaluation by task for the Elliptic dataset with UGCN backbone.}
\label{fig:heatmaps_elliptic_ugcn}
\end{figure*}

\begin{figure*}[]
\centering
\begin{subfigure}[b]{0.3\textwidth}
    \includegraphics[width=\textwidth]{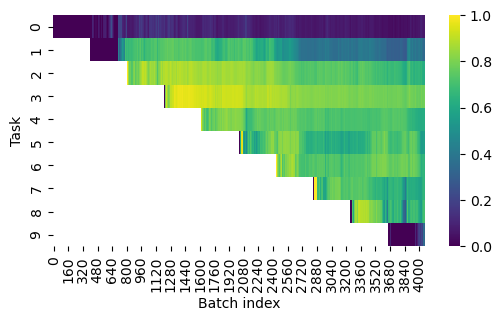}
    \caption{A-GEM}
    \label{fig:elliptic_grnf_agem}
\end{subfigure}
\hfill
\begin{subfigure}[b]{0.3\textwidth}
    \includegraphics[width=\textwidth]{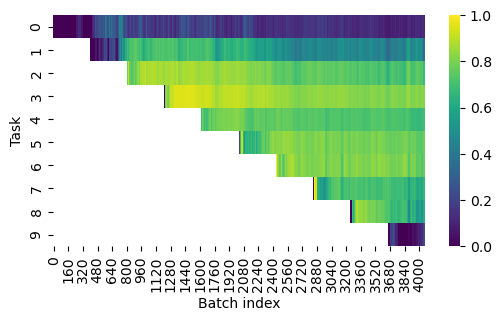}
    \caption{ER}
    \label{fig:elliptic_grnf_er}
\end{subfigure}
\hfill
\begin{subfigure}[b]{0.3\textwidth}
    \includegraphics[width=\textwidth]{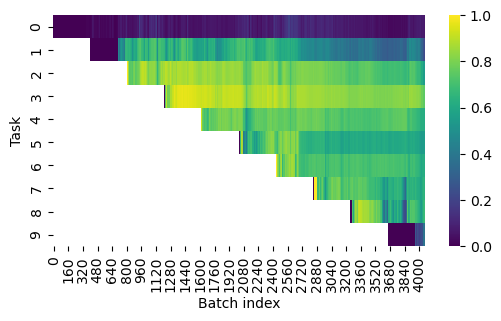}
    \caption{EWC}
    \label{fig:elliptic_grnf_ewc}
\end{subfigure}
\begin{subfigure}[b]{0.3\textwidth}
    \includegraphics[width=\textwidth]{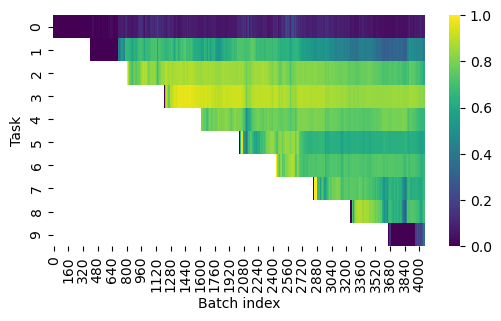}
    \caption{LwF}
    \label{fig:elliptic_grnf_lwf}
\end{subfigure}
\hfill
\begin{subfigure}[b]{0.3\textwidth}
    \includegraphics[width=\textwidth]{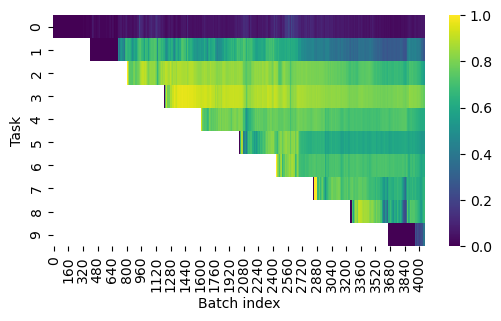}
    \caption{MAS}
    \label{fig:elliptic_grnf_mas}
\end{subfigure}
\hfill
\begin{subfigure}[b]{0.3\textwidth}
    \includegraphics[width=\textwidth]{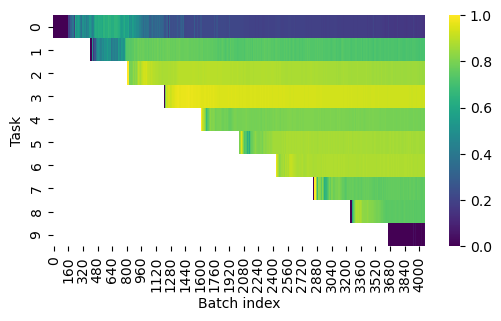}
    \caption{SLDA}
    \label{fig:elliptic_grnf_slda}
\end{subfigure}
\caption{Anytime evaluation by task for the Elliptic dataset with GRNF backbone.}
\label{fig:heatmaps_elliptic_grnf}
\end{figure*}

\begin{figure*}[]
\centering
\begin{subfigure}[b]{0.3\textwidth}
    \includegraphics[width=\textwidth]{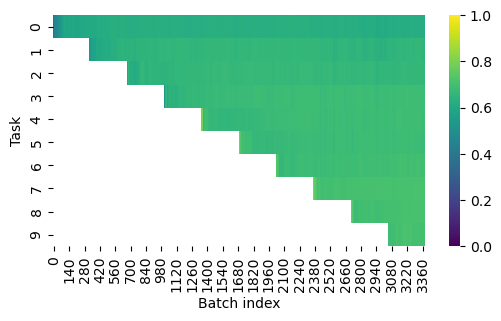}
    \caption{A-GEM}
    \label{fig:arxivTI_ugcn_agem}
\end{subfigure}
\hfill
\begin{subfigure}[b]{0.3\textwidth}
    \includegraphics[width=\textwidth]{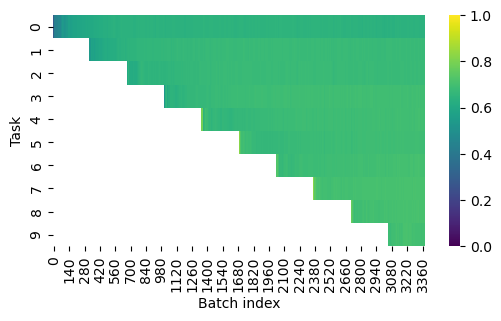}
    \caption{ER}
    \label{fig:arxivTI_ugcn_er}
\end{subfigure}
\hfill
\begin{subfigure}[b]{0.3\textwidth}
    \includegraphics[width=\textwidth]{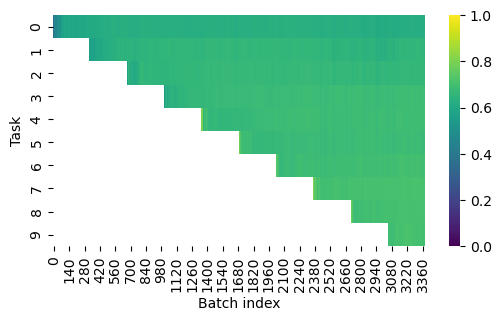}
    \caption{EWC}
    \label{fig:arxivTI_ugcn_ewc}
\end{subfigure}
\begin{subfigure}[b]{0.3\textwidth}
    \includegraphics[width=\textwidth]{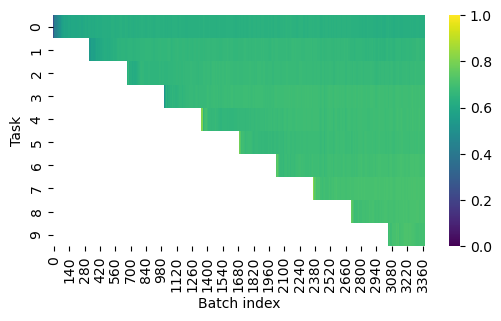}
    \caption{LwF}
    \label{fig:arxivTI_ugcn_lwf}
\end{subfigure}
\hfill
\begin{subfigure}[b]{0.3\textwidth}
    \includegraphics[width=\textwidth]{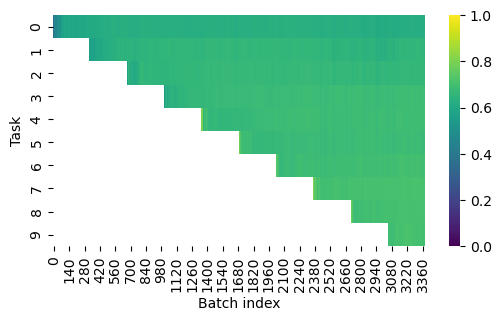}
    \caption{MAS}
    \label{fig:arxivTI_ugcn_mas}
\end{subfigure}
\hfill
\begin{subfigure}[b]{0.3\textwidth}
    \includegraphics[width=\textwidth]{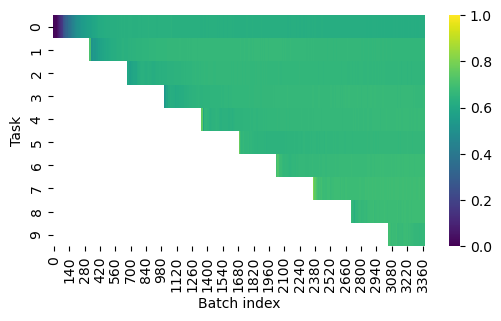}
    \caption{SLDA}
    \label{fig:arxivTI_ugcn_slda}
\end{subfigure}
\caption{Anytime evaluation by task for the Arxiv dataset with UGCN backbone with time-incremental stream. We note how the very homogeneous performance compared to other benchmarks suggests the absence of significant distribution drifts between tasks.}
\label{fig:heatmaps_arxivTI_ugcn}
\end{figure*}

\begin{figure*}[]
\centering
\begin{subfigure}[b]{0.3\textwidth}
    \includegraphics[width=\textwidth]{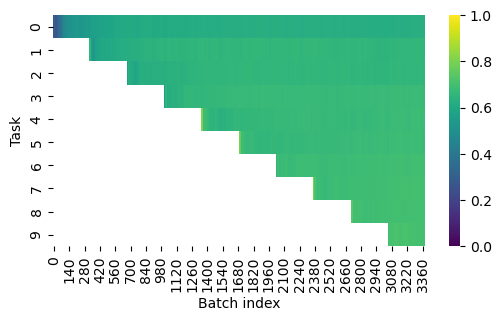}
    \caption{A-GEM}
    \label{fig:arxivTI_grnf_agem}
\end{subfigure}
\hfill
\begin{subfigure}[b]{0.3\textwidth}
    \includegraphics[width=\textwidth]{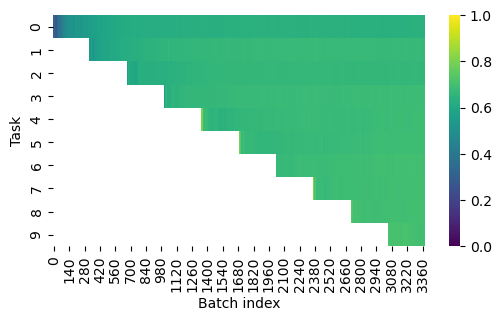}
    \caption{ER}
    \label{fig:arxivTI_grnf_er}
\end{subfigure}
\hfill
\begin{subfigure}[b]{0.3\textwidth}
    \includegraphics[width=\textwidth]{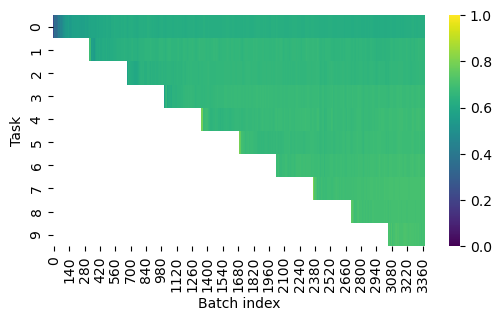}
    \caption{EWC}
    \label{fig:arxivTI_grnf_ewc}
\end{subfigure}
\begin{subfigure}[b]{0.3\textwidth}
    \includegraphics[width=\textwidth]{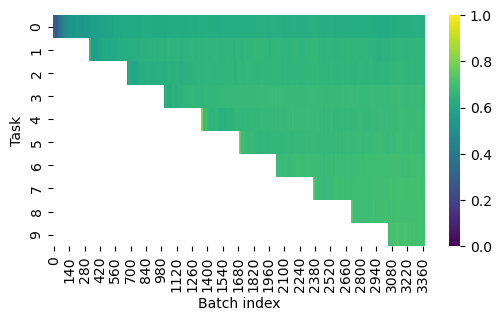}
    \caption{LwF}
    \label{fig:arxivTI_grnf_lwf}
\end{subfigure}
\hfill
\begin{subfigure}[b]{0.3\textwidth}
    \includegraphics[width=\textwidth]{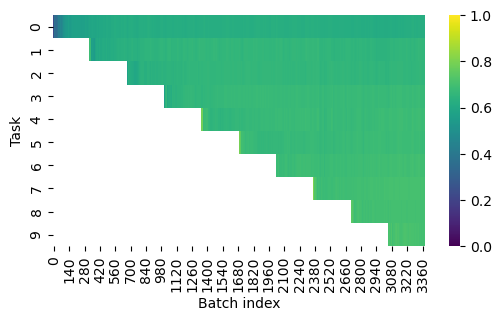}
    \caption{MAS}
    \label{fig:arxivTI_grnf_mas}
\end{subfigure}
\hfill
\begin{subfigure}[b]{0.3\textwidth}
    \includegraphics[width=\textwidth]{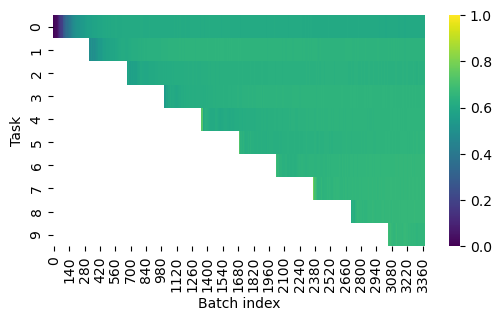}
    \caption{SLDA}
    \label{fig:arxivTI_grnf_slda}
\end{subfigure}
\caption{Anytime evaluation by task for the Arxiv dataset with GRNF backbone with time-incremental stream. We note how the very homogeneous performance compared to other benchmarks suggests the absence of significant distribution drifts between tasks.}
\label{fig:heatmaps_arxivTI_grnf}
\end{figure*}

In Figures \ref{fig:heatmaps_cora_ugcn}-\ref{fig:heatmaps_arxivTI_grnf}, we instead illustrate a more detailed breakdown of model performance: for each benchmark, backbone and considered method, we plot the results of anytime evaluation broken down on individual tasks, allowing a better understanding of when and where forgetting occours.

\begin{thebibliography}{80}
\providecommand{\natexlab}[1]{#1}
\providecommand{\url}[1]{\texttt{#1}}
\expandafter\ifx\csname urlstyle\endcsname\relax
  \providecommand{\doi}[1]{doi: #1}\else
  \providecommand{\doi}{doi: \begingroup \urlstyle{rm}\Url}\fi

\bibitem[Aljundi et~al.(2018)Aljundi, Babiloni, Elhoseiny, Rohrbach, and Tuytelaars]{aljundi_memory_2018}
Rahaf Aljundi, Francesca Babiloni, Mohamed Elhoseiny, Marcus Rohrbach, and Tinne Tuytelaars.
\newblock Memory {Aware} {Synapses}: {Learning} what (not) to forget.
\newblock In \emph{Proceedings of the {European} {Conference} on {Computer} {Vision} ({ECCV})}, pp.\  139--154, 2018.

\bibitem[Aljundi et~al.(2019)Aljundi, Belilovsky, Tuytelaars, Charlin, Caccia, Lin, and Page-Caccia]{aljundi_online_2019}
Rahaf Aljundi, Eugene Belilovsky, Tinne Tuytelaars, Laurent Charlin, Massimo Caccia, Min Lin, and Lucas Page-Caccia.
\newblock Online {Continual} {Learning} with {Maximal} {Interfered} {Retrieval}.
\newblock In \emph{Advances in {Neural} {Information} {Processing} {Systems}}, volume~32, 2019.

\bibitem[Bojchevski \& Günnemann(2018)Bojchevski and Günnemann]{bojchevski_deep_2018}
Aleksandar Bojchevski and Stephan Günnemann.
\newblock Deep {Gaussian} {Embedding} of {Graphs}: {Unsupervised} {Inductive} {Learning} via {Ranking}.
\newblock In \emph{International Conference on Learning Representations}, February 2018.

\bibitem[Caccia et~al.(2021)Caccia, Aljundi, Asadi, Tuytelaars, Pineau, and Belilovsky]{caccia_new_2021}
Lucas Caccia, Rahaf Aljundi, Nader Asadi, Tinne Tuytelaars, Joelle Pineau, and Eugene Belilovsky.
\newblock New {Insights} on {Reducing} {Abrupt} {Representation} {Change} in {Online} {Continual} {Learning}.
\newblock In \emph{International Conference on Learning Representations}, October 2021.

\bibitem[Caroprese et~al.(2025)Caroprese, Pisani, Veloso, Konig, Manco, Hoos, and Gama]{Caroprese2025}
Luciano Caroprese, Francesco~Sergio Pisani, Bruno~Miguel Veloso, Matthias Konig, Giuseppe Manco, Holger Hoos, and Joao Gama.
\newblock Modelling concept drift in dynamic data streams for recommender systems.
\newblock \emph{ACM Trans. Recomm. Syst.}, 3\penalty0 (2), March 2025.
\newblock \doi{10.1145/3707693}.
\newblock URL \url{https://doi.org/10.1145/3707693}.

\bibitem[Chaudhry et~al.(2018)Chaudhry, Ranzato, Rohrbach, and Elhoseiny]{chaudhry_efficient_2018}
Arslan Chaudhry, Marc’Aurelio Ranzato, Marcus Rohrbach, and Mohamed Elhoseiny.
\newblock Efficient {Lifelong} {Learning} with {A}-{GEM}.
\newblock In \emph{ICLR}, September 2018.

\bibitem[Chaudhry et~al.(2019)Chaudhry, Rohrbach, Elhoseiny, Ajanthan, Dokania, Torr, and Ranzato]{chaudhry_tiny_2019}
Arslan Chaudhry, Marcus Rohrbach, Mohamed Elhoseiny, Thalaiyasingam Ajanthan, Puneet~K. Dokania, Philip H.~S. Torr, and Marc'Aurelio Ranzato.
\newblock On {Tiny} {Episodic} {Memories} in {Continual} {Learning}, 2019.
\newblock arXiv:1902.10486.

\bibitem[Chen et~al.(2021)Chen, Wang, and Xie]{ijcai2021p498}
Xu~Chen, Junshan Wang, and Kunqing Xie.
\newblock Trafficstream: A streaming traffic flow forecasting framework based on graph neural networks and continual learning.
\newblock In \emph{Proceedings of the Thirtieth International Joint Conference on Artificial Intelligence, {IJCAI-21}}, pp.\  3620--3626. International Joint Conferences on Artificial Intelligence Organization, 8 2021.
\newblock \doi{10.24963/ijcai.2021/498}.

\bibitem[Cui et~al.(2023)Cui, Wang, Sun, Liu, Jiang, Han, and Hu]{Cui_2023_LifelongEmbeddingLearning}
Yuanning Cui, Yuxin Wang, Zequn Sun, Wenqiang Liu, Yiqiao Jiang, Kexin Han, and Wei Hu.
\newblock Lifelong {Embedding} {Learning} and {Transfer} for {Growing} {Knowledge} {Graphs}.
\newblock \emph{Proceedings of the AAAI Conference on Artificial Intelligence}, 37\penalty0 (4):\penalty0 4217--4224, June 2023.
\newblock \doi{10.1609/aaai.v37i4.25539}.

\bibitem[De~Lange et~al.(2022)De~Lange, Aljundi, Masana, Parisot, Jia, Leonardis, Slabaugh, and Tuytelaars]{de_lange_continual_2022}
Matthias De~Lange, Rahaf Aljundi, Marc Masana, Sarah Parisot, Xu~Jia, Aleš Leonardis, Gregory Slabaugh, and Tinne Tuytelaars.
\newblock A {Continual} {Learning} {Survey}: {Defying} {Forgetting} in {Classification} {Tasks}.
\newblock \emph{IEEE Transactions on Pattern Analysis and Machine Intelligence}, 44\penalty0 (7):\penalty0 3366--3385, July 2022.
\newblock \doi{10.1109/TPAMI.2021.3057446}.

\bibitem[Ding et~al.(2024)Ding, Ji, Wang, and Xu]{Ding_2024_UnderstandingForgettingContinual}
Meng Ding, Kaiyi Ji, Di~Wang, and Jinhui Xu.
\newblock Understanding forgetting in continual learning with linear regression.
\newblock In \emph{Proceedings of the 41st {International} {Conference} on {Machine} {Learning}}, volume 235, pp.\  10978--11001, 2024.

\bibitem[Donghi et~al.(2024)Donghi, Pasa, Oneto, Gallicchio, Micheli, Anguita, Sperduti, and Navarin]{Donghi_2024_InvestigatingOverparameterizedRandomized}
Giovanni Donghi, Luca Pasa, Luca Oneto, Claudio Gallicchio, Alessio Micheli, Davide Anguita, Alessandro Sperduti, and Nicolò Navarin.
\newblock Investigating over-parameterized randomized graph networks.
\newblock \emph{Neurocomputing}, 606, November 2024.
\newblock \doi{10.1016/j.neucom.2024.128281}.

\bibitem[Donghi et~al.(2025)Donghi, Pasa, Zambon, Alippi, and Navarin]{ocgl}
Giovanni Donghi, Luca Pasa, Daniele Zambon, Cesare Alippi, and Nicolò Navarin.
\newblock {Online Continual Graph Learning}, 2025.
\newblock arXiv:2508.03283.

\bibitem[Evron et~al.(2022)Evron, Moroshko, Ward, Srebro, and Soudry]{Evron_2022}
Itay Evron, Edward Moroshko, Rachel Ward, Nathan Srebro, and Daniel Soudry.
\newblock How catastrophic can catastrophic forgetting be in linear regression?
\newblock In \emph{Proceedings of Thirty Fifth Conference on Learning Theory}, volume 178, pp.\  4028--4079. PMLR, Jul 2022.

\bibitem[Evron et~al.(2023)Evron, Moroshko, Buzaglo, Khriesh, Marjieh, Srebro, and Soudry]{Evron_2023_ContinualLearningLinear}
Itay Evron, Edward Moroshko, Gon Buzaglo, Maroun Khriesh, Badea Marjieh, Nathan Srebro, and Daniel Soudry.
\newblock Continual learning in linear classification on separable data.
\newblock In \emph{Proceedings of the 40th {International} {Conference} on {Machine} {Learning}}, volume 202, pp.\  9440--9484, 2023.

\bibitem[Febrinanto et~al.(2023)Febrinanto, Xia, Moore, Thapa, and Aggarwal]{febrinanto_graph_2023}
Falih~Gozi Febrinanto, Feng Xia, Kristen Moore, Chandra Thapa, and Charu Aggarwal.
\newblock Graph {Lifelong} {Learning}: {A} {Survey}.
\newblock \emph{IEEE Computational Intelligence Magazine}, 18\penalty0 (1):\penalty0 32--51, February 2023.
\newblock \doi{10.1109/MCI.2022.3222049}.

\bibitem[Gallicchio \& Micheli(2010)Gallicchio and Micheli]{Gallicchio_2010_GraphEchoState}
Claudio Gallicchio and Alessio Micheli.
\newblock Graph {Echo} {State} {Networks}.
\newblock In \emph{The 2010 {International} {Joint} {Conference} on {Neural} {Networks} ({IJCNN})}, July 2010.
\newblock \doi{10.1109/IJCNN.2010.5596796}.

\bibitem[Gilmer et~al.(2017)Gilmer, Schoenholz, Riley, Vinyals, and Dahl]{gilmer_neural_2017}
Justin Gilmer, Samuel~S. Schoenholz, Patrick~F. Riley, Oriol Vinyals, and George~E. Dahl.
\newblock Neural {Message} {Passing} for {Quantum} {Chemistry}.
\newblock In \emph{Proceedings of the 34th {International} {Conference} on {Machine} {Learning}}, pp.\  1263--1272. PMLR, July 2017.
\newblock ISSN: 2640-3498.

\bibitem[Glorot \& Bengio(2010)Glorot and Bengio]{pmlr-v9-glorot10a}
X.~Glorot and Y.~Bengio.
\newblock Understanding the difficulty of training deep feedforward neural networks.
\newblock In \emph{International Conference on Artificial Intelligence and Statistics}, 2010.

\bibitem[Hamilton et~al.(2017)Hamilton, Ying, and Leskovec]{hamilton_inductive_2017}
Will Hamilton, Zhitao Ying, and Jure Leskovec.
\newblock Inductive {Representation} {Learning} on {Large} {Graphs}.
\newblock In \emph{Advances in {Neural} {Information} {Processing} {Systems}}, volume~30, 2017.

\bibitem[Hayes \& Kanan(2020)Hayes and Kanan]{Hayes_2020_LifelongMachineLearning}
Tyler~L. Hayes and Christopher Kanan.
\newblock Lifelong {Machine} {Learning} with {Deep} {Streaming} {Linear} {Discriminant} {Analysis}.
\newblock In \emph{2020 {IEEE}/{CVF} {Conference} on {Computer} {Vision} and {Pattern} {Recognition} {Workshops} ({CVPRW})}, pp.\  887--896, Seattle, WA, USA, June 2020.
\newblock \doi{10.1109/CVPRW50498.2020.00118}.

\bibitem[Hoang et~al.(2023)Hoang, Tung, Nguyen, Nguyen, Nguyen, and Le]{hoang_universal_2023}
Thanh~Duc Hoang, Do~Viet Tung, Duy-Hung Nguyen, Bao-Sinh Nguyen, Huy~Hoang Nguyen, and Hung Le.
\newblock Universal {Graph} {Continual} {Learning}.
\newblock \emph{Transactions on Machine Learning Research}, 2023.

\bibitem[Hu et~al.(2021)Hu, Fey, Zitnik, Dong, Ren, Liu, Catasta, and Leskovec]{hu_open_2021}
Weihua Hu, Matthias Fey, Marinka Zitnik, Yuxiao Dong, Hongyu Ren, Bowen Liu, Michele Catasta, and Jure Leskovec.
\newblock Open {Graph} {Benchmark}: {Datasets} for {Machine} {Learning} on {Graphs}, February 2021.
\newblock arXiv:2005.00687.

\bibitem[Huang et~al.(2023)Huang, Li, Cao, Fujita, Li, Wu, and Li]{Huang_2023}
C.~Huang, M.~Li, F.~Cao, H.~Fujita, Z.~Li, X.~Wu, and M.~Li.
\newblock {Are Graph Convolutional Networks With Random Weights Feasible?}
\newblock \emph{IEEE Transactions on Pattern Analysis and Machine Intelligence}, 45\penalty0 (3):\penalty0 2751--2768, 2023.

\bibitem[Huang et~al.(2015)Huang, Huang, Song, and You]{Huang_2015_TrendsExtremeLearning}
Gao Huang, Guang-Bin Huang, Shiji Song, and Keyou You.
\newblock Trends in extreme learning machines: {A} review.
\newblock \emph{Neural Networks}, 61:\penalty0 32--48, January 2015.
\newblock \doi{10.1016/j.neunet.2014.10.001}.

\bibitem[Huang et~al.(2006)Huang, Zhu, and Siew]{Huang_2006_ExtremeLearningMachine}
Guang-Bin Huang, Qin-Yu Zhu, and Chee-Kheong Siew.
\newblock Extreme learning machine: {Theory} and applications.
\newblock \emph{Neurocomputing}, 70\penalty0 (1-3):\penalty0 489--501, December 2006.
\newblock \doi{10.1016/j.neucom.2005.12.126}.

\bibitem[Jaeger \& Haas(2004)Jaeger and Haas]{Jaeger2004}
Herbert Jaeger and Harald Haas.
\newblock {Harnessing Nonlinearity: Predicting Chaotic Systems and Saving Energy in Wireless Communication}.
\newblock \emph{Science}, 304\penalty0 (5667):\penalty0 78--80, 2004.

\bibitem[Keriven \& Peyr\'{e}(2019)Keriven and Peyr\'{e}]{Keriven_2019}
Nicolas Keriven and Gabriel Peyr\'{e}.
\newblock Universal invariant and equivariant graph neural networks.
\newblock In \emph{Proceedings of the 33rd International Conference on Neural Information Processing Systems}, 2019.

\bibitem[Kingma \& Ba(2017)Kingma and Ba]{kingma_adam_2017}
Diederik~P. Kingma and Jimmy Ba.
\newblock Adam: {A} {Method} for {Stochastic} {Optimization}, January 2017.
\newblock arXiv:1412.6980.

\bibitem[Kipf \& Welling(2017)Kipf and Welling]{kipf2017semisupervised}
Thomas~N. Kipf and Max Welling.
\newblock Semi-supervised classification with graph convolutional networks.
\newblock In \emph{International Conference on Learning Representations}, 2017.

\bibitem[Kirkpatrick et~al.(2017)Kirkpatrick, Pascanu, Rabinowitz, Veness, Desjardins, Rusu, Milan, Quan, Ramalho, Grabska-Barwinska, Hassabis, Clopath, Kumaran, and Hadsell]{kirkpatrick_overcoming_2017}
James Kirkpatrick, Razvan Pascanu, Neil Rabinowitz, Joel Veness, Guillaume Desjardins, Andrei~A. Rusu, Kieran Milan, John Quan, Tiago Ramalho, Agnieszka Grabska-Barwinska, Demis Hassabis, Claudia Clopath, Dharshan Kumaran, and Raia Hadsell.
\newblock Overcoming catastrophic forgetting in neural networks.
\newblock \emph{Proceedings of the National Academy of Sciences}, 114\penalty0 (13):\penalty0 3521--3526, March 2017.
\newblock \doi{10.1073/pnas.1611835114}.

\bibitem[Koh et~al.(2021)Koh, Kim, Ha, and Choi]{koh_online_2021}
Hyunseo Koh, Dahyun Kim, Jung-Woo Ha, and Jonghyun Choi.
\newblock Online {Continual} {Learning} on {Class} {Incremental} {Blurry} {Task} {Configuration} with {Anytime} {Inference}.
\newblock In \emph{International Conference on Learning Representations}, October 2021.

\bibitem[Le~Baher et~al.(2023)Le~Baher, Azé, Bringay, Poncelet, Rodriguez, and Dunoyer]{LeBaher_2023_PatientElectronicHealth}
Hugo Le~Baher, Jérôme Azé, Sandra Bringay, Pascal Poncelet, Nancy Rodriguez, and Caroline Dunoyer.
\newblock Patient {Electronic} {Health} {Record} as {Temporal} {Graphs} for {Health} {Monitoring}.
\newblock \emph{Studies in Health Technology and Informatics}, 302:\penalty0 561--565, May 2023.
\newblock ISSN 1879-8365.
\newblock \doi{10.3233/SHTI230205}.

\bibitem[Li \& Zeng(2023)Li and Zeng]{Li_2023_CRNetFastContinual}
Depeng Li and Zhigang Zeng.
\newblock {CRNet}: {A} {Fast} {Continual} {Learning} {Framework} {With} {Random} {Theory}.
\newblock \emph{IEEE Transactions on Pattern Analysis and Machine Intelligence}, 45\penalty0 (9):\penalty0 10731--10744, September 2023.
\newblock \doi{10.1109/TPAMI.2023.3262853}.

\bibitem[Li \& Hoiem(2018)Li and Hoiem]{li_learning_2018}
Zhizhong Li and Derek Hoiem.
\newblock Learning without {Forgetting}.
\newblock \emph{IEEE Transactions on Pattern Analysis and Machine Intelligence}, 40\penalty0 (12):\penalty0 2935--2947, December 2018.
\newblock \doi{10.1109/TPAMI.2017.2773081}.

\bibitem[Lin et~al.(2024)Lin, Zhong, Qiu, and Liang]{Lin_2024_EGRACLIoTIntrusion}
Lieqing Lin, Qi~Zhong, Jiasheng Qiu, and Zhenyu Liang.
\newblock E-{GRACL}: an {IoT} intrusion detection system based on graph neural networks.
\newblock \emph{The Journal of Supercomputing}, 81\penalty0 (1):\penalty0 42, October 2024.
\newblock ISSN 1573-0484.
\newblock \doi{10.1007/s11227-024-06471-5}.
\newblock URL \url{https://doi.org/10.1007/s11227-024-06471-5}.

\bibitem[Liu et~al.(2021)Liu, Yang, and Wang]{liu_overcoming_2021}
Huihui Liu, Yiding Yang, and Xinchao Wang.
\newblock Overcoming {Catastrophic} {Forgetting} in {Graph} {Neural} {Networks}.
\newblock \emph{Proceedings of the AAAI Conference on Artificial Intelligence}, 35\penalty0 (10):\penalty0 8653--8661, May 2021.
\newblock \doi{10.1609/aaai.v35i10.17049}.

\bibitem[Liu et~al.(2012)Liu, Gao, and Li]{LIU201258}
Xueyi Liu, Chuanhou Gao, and Ping Li.
\newblock A comparative analysis of support vector machines and extreme learning machines.
\newblock \emph{Neural Networks}, 33:\penalty0 58--66, 2012.
\newblock \doi{https://doi.org/10.1016/j.neunet.2012.04.002}.

\bibitem[Liu et~al.(2023)Liu, Qiu, and Huang]{liu_cat_2023}
Yilun Liu, Ruihong Qiu, and Zi~Huang.
\newblock {CaT}: {Balanced} {Continual} {Graph} {Learning} with {Graph} {Condensation}.
\newblock In \emph{2023 {IEEE} {International} {Conference} on {Data} {Mining} ({ICDM})}, pp.\  1157--1162, 2023.
\newblock \doi{10.1109/ICDM58522.2023.00141}.

\bibitem[Lopez-Paz \& Ranzato(2017)Lopez-Paz and Ranzato]{lopez-paz_gradient_2017}
David Lopez-Paz and Marc'~Aurelio Ranzato.
\newblock Gradient {Episodic} {Memory} for {Continual} {Learning}.
\newblock In \emph{Advances in {Neural} {Information} {Processing} {Systems}}, volume~30, 2017.

\bibitem[Mai et~al.(2022)Mai, Li, Jeong, Quispe, Kim, and Sanner]{mai_online_2022}
Zheda Mai, Ruiwen Li, Jihwan Jeong, David Quispe, Hyunwoo Kim, and Scott Sanner.
\newblock Online continual learning in image classification: {An} empirical survey.
\newblock \emph{Neurocomputing}, 469:\penalty0 28--51, January 2022.
\newblock \doi{10.1016/j.neucom.2021.10.021}.

\bibitem[Maron et~al.(2019)Maron, Fetaya, Segol, and Lipman]{Maron_2019}
Haggai Maron, Ethan Fetaya, Nimrod Segol, and Yaron Lipman.
\newblock On the universality of invariant networks.
\newblock In \emph{Proceedings of the 36th International Conference on Machine Learning}, pp.\  4363--4371, Jun 2019.

\bibitem[McDonnell et~al.(2023)McDonnell, Gong, Parvaneh, Abbasnejad, and van~den Hengel]{McDonnell_2023_RanPACRandomProjections}
Mark~D. McDonnell, Dong Gong, Amin Parvaneh, Ehsan Abbasnejad, and Anton van~den Hengel.
\newblock {RanPAC}: {Random} {Projections} and {Pre}-trained {Models} for {Continual} {Learning}.
\newblock \emph{Advances in Neural Information Processing Systems}, 36:\penalty0 12022--12053, December 2023.

\bibitem[Mehta et~al.(2023)Mehta, Patil, Chandar, and Strubell]{Mehta_2023_EmpiricalInvestigationRole}
Sanket~Vaibhav Mehta, Darshan Patil, Sarath Chandar, and Emma Strubell.
\newblock An {Empirical} {Investigation} of the {Role} of {Pre}-training in {Lifelong} {Learning}.
\newblock \emph{Journal of Machine Learning Research}, 24\penalty0 (214):\penalty0 1--50, 2023.

\bibitem[Micheli(2009)]{micheli_neural_2009}
Alessio Micheli.
\newblock Neural {Network} for {Graphs}: {A} {Contextual} {Constructive} {Approach}.
\newblock \emph{IEEE Transactions on Neural Networks}, 20\penalty0 (3):\penalty0 498--511, March 2009.
\newblock \doi{10.1109/TNN.2008.2010350}.

\bibitem[Mirzadeh et~al.(2022)Mirzadeh, Chaudhry, Yin, Hu, Pascanu, Gorur, and Farajtabar]{Mirzadeh_2022_WideNeuralNetworks}
Seyed~Iman Mirzadeh, Arslan Chaudhry, Dong Yin, Huiyi Hu, Razvan Pascanu, Dilan Gorur, and Mehrdad Farajtabar.
\newblock Wide {Neural} {Networks} {Forget} {Less} {Catastrophically}.
\newblock In \emph{Proceedings of the 39th {International} {Conference} on {Machine} {Learning}}, pp.\  15699--15717, June 2022.

\bibitem[Navarin et~al.(2023{\natexlab{a}})Navarin, Pasa, Gallicchio, and Sperduti]{navarin2023}
N.~Navarin, L.~Pasa, C.~Gallicchio, and A.~Sperduti.
\newblock An untrained neural model for fast and accurate graph classification.
\newblock In \emph{International Conference on Artificial Neural Networks}, 2023{\natexlab{a}}.

\bibitem[Navarin et~al.(2023{\natexlab{b}})Navarin, Pasa, Oneto, and Sperduti]{Navarin_2023_EmpiricalStudyOverParameterized}
Nicolò Navarin, Luca Pasa, Luca Oneto, and Alessandro Sperduti.
\newblock An {Empirical} {Study} of {Over}-{Parameterized} {Neural} {Models} based on {Graph} {Random} {Features}.
\newblock In \emph{{ESANN} 2023 proceesdings}, pp.\  17--22, 2023{\natexlab{b}}.
\newblock \doi{10.14428/esann/2023.ES2023-145}.

\bibitem[Pao \& Takefuji(1992)Pao and Takefuji]{Pao_1992}
Y.-H. Pao and Y.~Takefuji.
\newblock Functional-link net computing: theory, system architecture, and functionalities.
\newblock \emph{Computer}, 25\penalty0 (5):\penalty0 76--79, 1992.
\newblock \doi{10.1109/2.144401}.

\bibitem[Pao et~al.(1994)Pao, Park, and Sobajic]{PAO1994163}
Yoh-Han Pao, Gwang-Hoon Park, and Dejan~J. Sobajic.
\newblock Learning and generalization characteristics of the random vector functional-link net.
\newblock \emph{Neurocomputing}, 6\penalty0 (2):\penalty0 163--180, 1994.
\newblock ISSN 0925-2312.
\newblock \doi{https://doi.org/10.1016/0925-2312(94)90053-1}.
\newblock Backpropagation, Part IV.

\bibitem[Parisi et~al.(2019)Parisi, Kemker, Part, Kanan, and Wermter]{parisi_continual_2019}
German~I. Parisi, Ronald Kemker, Jose~L. Part, Christopher Kanan, and Stefan Wermter.
\newblock Continual lifelong learning with neural networks: {A} review.
\newblock \emph{Neural Networks}, 113:\penalty0 54--71, May 2019.
\newblock \doi{10.1016/j.neunet.2019.01.012}.

\bibitem[Pasa et~al.(2022)Pasa, Navarin, and Sperduti]{pasa2022}
Luca Pasa, Nicolò Navarin, and Alessandro Sperduti.
\newblock Multiresolution reservoir graph neural network.
\newblock \emph{IEEE Transactions on Neural Networks and Learning Systems}, 33\penalty0 (6):\penalty0 2642--2653, 2022.
\newblock \doi{10.1109/TNNLS.2021.3090503}.

\bibitem[Pelosin(2022)]{Pelosin_2022_SimplerBetterOfftheshelf}
Francesco Pelosin.
\newblock Simpler is {Better}: off-the-shelf {Continual} {Learning} {Through} {Pretrained} {Backbones}, May 2022.
\newblock arXiv:2205.01586.

\bibitem[Platonov et~al.(2022)Platonov, Kuznedelev, Diskin, Babenko, and Prokhorenkova]{Platonov_2022_CriticalLookEvaluation}
Oleg Platonov, Denis Kuznedelev, Michael Diskin, Artem Babenko, and Liudmila Prokhorenkova.
\newblock A critical look at the evaluation of {GNNs} under heterophily: {Are} we really making progress?
\newblock In \emph{International Conference on Learning Representations}, 2022.

\bibitem[Prabhu et~al.(2025)Prabhu, Sinha, Kumaraguru, Torr, Sener, and Dokania]{Prabhu_2025_RanDumbRandomRepresentations}
Ameya Prabhu, Shiven Sinha, Ponnurangam Kumaraguru, Philip Torr, Ozan Sener, and Puneet Dokania.
\newblock {RanDumb}: {Random} {Representations} {Outperform} {Online} {Continually} {Learned} {Representations}.
\newblock \emph{Advances in Neural Information Processing Systems}, 37:\penalty0 37988--38006, January 2025.

\bibitem[Rahimi \& Recht(2007)Rahimi and Recht]{NIPS2007_013a006f}
Ali Rahimi and Benjamin Recht.
\newblock Random features for large-scale kernel machines.
\newblock In J.~Platt, D.~Koller, Y.~Singer, and S.~Roweis (eds.), \emph{Advances in Neural Information Processing Systems}, volume~20, 2007.

\bibitem[Rahimi \& Recht(2008)Rahimi and Recht]{rahimi2008weighted}
Ali Rahimi and Benjamin Recht.
\newblock Weighted sums of random kitchen sinks: Replacing minimization with randomization in learning.
\newblock In D.~Koller, D.~Schuurmans, Y.~Bengio, and L.~Bottou (eds.), \emph{Advances in Neural Information Processing Systems}, volume~21. Curran Associates, Inc., 2008.
\newblock URL \url{https://proceedings.neurips.cc/paper_files/paper/2008/file/0efe32849d230d7f53049ddc4a4b0c60-Paper.pdf}.

\bibitem[Rebuffi et~al.(2017)Rebuffi, Kolesnikov, Sperl, and Lampert]{rebuffi_icarl_2017}
Sylvestre-Alvise Rebuffi, Alexander Kolesnikov, Georg Sperl, and Christoph~H. Lampert.
\newblock {iCaRL}: {Incremental} {Classifier} and {Representation} {Learning}.
\newblock In \emph{2017 {IEEE} {Conference} on {Computer} {Vision} and {Pattern} {Recognition} ({CVPR})}, pp.\  5533--5542, Honolulu, HI, July 2017. IEEE.
\newblock \doi{10.1109/CVPR.2017.587}.

\bibitem[Rolnick et~al.(2019)Rolnick, Ahuja, Schwarz, Lillicrap, and Wayne]{rolnick_experience_2019}
David Rolnick, Arun Ahuja, Jonathan Schwarz, Timothy Lillicrap, and Gregory Wayne.
\newblock Experience {Replay} for {Continual} {Learning}.
\newblock In \emph{Advances in {Neural} {Information} {Processing} {Systems}}, volume~32, 2019.

\bibitem[Rudi \& Rosasco(2017{\natexlab{a}})Rudi and Rosasco]{NIPS2017_61b1fb3f}
Alessandro Rudi and Lorenzo Rosasco.
\newblock Generalization properties of learning with random features.
\newblock In \emph{Advances in Neural Information Processing Systems}, volume~30, 2017{\natexlab{a}}.

\bibitem[Rudi \& Rosasco(2017{\natexlab{b}})Rudi and Rosasco]{rudi2017generalization}
Alessandro Rudi and Lorenzo Rosasco.
\newblock Generalization properties of learning with random features.
\newblock In I.~Guyon, U.~Von Luxburg, S.~Bengio, H.~Wallach, R.~Fergus, S.~Vishwanathan, and R.~Garnett (eds.), \emph{Advances in Neural Information Processing Systems}, volume~30. Curran Associates, Inc., 2017{\natexlab{b}}.
\newblock URL \url{https://proceedings.neurips.cc/paper_files/paper/2017/file/61b1fb3f59e28c67f3925f3c79be81a1-Paper.pdf}.

\bibitem[Scardapane \& Wang(2017)Scardapane and Wang]{Scardapane_2017_RandomnessNeuralNetworks}
Simone Scardapane and Dianhui Wang.
\newblock Randomness in neural networks: an overview.
\newblock \emph{WIREs Data Mining and Knowledge Discovery}, 7\penalty0 (2), 2017.
\newblock \doi{10.1002/widm.1200}.

\bibitem[Scarselli et~al.(2009)Scarselli, Gori, Tsoi, Hagenbuchner, and Monfardini]{scarselli_graph_2009}
Franco Scarselli, Marco Gori, Ah~Chung Tsoi, Markus Hagenbuchner, and Gabriele Monfardini.
\newblock The {Graph} {Neural} {Network} {Model}.
\newblock \emph{IEEE Transactions on Neural Networks}, 20\penalty0 (1):\penalty0 61--80, January 2009.
\newblock \doi{10.1109/TNN.2008.2005605}.

\bibitem[Shchur et~al.(2019)Shchur, Mumme, Bojchevski, and Günnemann]{shchur_pitfalls_2019}
Oleksandr Shchur, Maximilian Mumme, Aleksandar Bojchevski, and Stephan Günnemann.
\newblock Pitfalls of {Graph} {Neural} {Network} {Evaluation}, June 2019.
\newblock arXiv:1811.05868.

\bibitem[Soutif–Cormerais et~al.(2023)Soutif–Cormerais, Carta, Cossu, Hurtado, Lomonaco, Van De~Weijer, and Hemati]{soutifcormerais_comprehensive_2023}
Albin Soutif–Cormerais, Antonio Carta, Andrea Cossu, Julio Hurtado, Vincenzo Lomonaco, Joost Van De~Weijer, and Hamed Hemati.
\newblock A {Comprehensive} {Empirical} {Evaluation} on {Online} {Continual} {Learning}.
\newblock In \emph{2023 {IEEE}/{CVF} {International} {Conference} on {Computer} {Vision} {Workshops} ({ICCVW})}, pp.\  3510--3520, October 2023.
\newblock \doi{10.1109/ICCVW60793.2023.00378}.

\bibitem[Su et~al.(2023)Su, Zou, Zhang, and Wu]{Su_2023_RobustGraphIncremental}
Junwei Su, Difan Zou, Zijun Zhang, and Chuan Wu.
\newblock Towards robust graph incremental learning on evolving graphs.
\newblock In \emph{Proceedings of the 40th {International} {Conference} on {Machine} {Learning}}, volume 202, pp.\  32728--32748, 2023.

\bibitem[Su et~al.(2024)Su, Zou, and Wu]{Su_2024_LimitationExperienceReplay}
Junwei Su, Difan Zou, and Chuan Wu.
\newblock On the {Limitation} and {Experience} {Replay} for {GNNs} in {Continual} {Learning}, July 2024.
\newblock arXiv:2302.03534.

\bibitem[Sun et~al.(2023)Sun, Ye, Peng, Wang, and Yu]{Sun_2023_SelfSupervisedContinualGraph}
Li~Sun, Junda Ye, Hao Peng, Feiyang Wang, and Philip~S. Yu.
\newblock Self-{Supervised} {Continual} {Graph} {Learning} in {Adaptive} {Riemannian} {Spaces}.
\newblock \emph{Proceedings of the AAAI Conference on Artificial Intelligence}, 37\penalty0 (4):\penalty0 4633--4642, June 2023.
\newblock \doi{10.1609/aaai.v37i4.25586}.

\bibitem[Van De~Ven et~al.(2022)Van De~Ven, Tuytelaars, and Tolias]{van_de_ven_three_2022}
Gido~M. Van De~Ven, Tinne Tuytelaars, and Andreas~S. Tolias.
\newblock Three types of incremental learning.
\newblock \emph{Nature Machine Intelligence}, 4\penalty0 (12):\penalty0 1185--1197, December 2022.
\newblock \doi{10.1038/s42256-022-00568-3}.

\bibitem[Wang et~al.(2022)Wang, Ciccone, Luise, Yapp, Pontil, and Ciliberto]{Wang_2022_ScheduleRobustOnlineContinual}
Ruohan Wang, Marco Ciccone, Giulia Luise, Andrew Yapp, Massimiliano Pontil, and Carlo Ciliberto.
\newblock Schedule-{Robust} {Online} {Continual} {Learning}, October 2022.
\newblock arXiv:2210.05561 [cs].

\bibitem[Wang et~al.(2025)Wang, Liu, Ma, Li, Zhang, Fu, Li, Yuan, Song, Ma, Zeng, Chen, Zhao, Li, Jiang, Lio, Chawla, Zhang, and Ye]{Wang_2025_GraphFoundationModels}
Zehong Wang, Zheyuan Liu, Tianyi Ma, Jiazheng Li, Zheyuan Zhang, Xingbo Fu, Yiyang Li, Zhengqing Yuan, Wei Song, Yijun Ma, Qingkai Zeng, Xiusi Chen, Jianan Zhao, Jundong Li, Meng Jiang, Pietro Lio, Nitesh Chawla, Chuxu Zhang, and Yanfang Ye.
\newblock Graph {Foundation} {Models}: {A} {Comprehensive} {Survey}, May 2025.
\newblock arXiv:2505.15116 [cs].

\bibitem[Weber et~al.(2019)Weber, Domeniconi, Chen, Weidele, Bellei, Robinson, and Leiserson]{Weber_2019_AntiMoneyLaunderingBitcoin}
Mark Weber, Giacomo Domeniconi, Jie Chen, Daniel Karl~I. Weidele, Claudio Bellei, Tom Robinson, and Charles~E. Leiserson.
\newblock Anti-{Money} {Laundering} in {Bitcoin}: {Experimenting} with {Graph} {Convolutional} {Networks} for {Financial} {Forensics}, 2019.
\newblock arXiv:1908.02591.

\bibitem[Yuan et~al.(2023)Yuan, Guan, Ni, Luo, Man, Wong, and Chang]{yuan_continual_2023}
Qiao Yuan, Sheng-Uei Guan, Pin Ni, Tianlun Luo, Ka~Lok Man, Prudence Wong, and Victor Chang.
\newblock Continual {Graph} {Learning}: {A} {Survey}, 2023.
\newblock arXiv:2301.12230.

\bibitem[Zambon et~al.(2020)Zambon, Alippi, and Livi]{Zambon_2020_GraphRandomNeural}
Daniele Zambon, Cesare Alippi, and Lorenzo Livi.
\newblock Graph {Random} {Neural} {Features} for {Distance}-{Preserving} {Graph} {Representations}.
\newblock In \emph{Proceedings of the 37th {International} {Conference} on {Machine} {Learning}}, pp.\  10968--10977, November 2020.

\bibitem[Zhang et~al.(2022{\natexlab{a}})Zhang, Song, and Tao]{zhang_cglb_2022}
Xikun Zhang, Dongjin Song, and Dacheng Tao.
\newblock {CGLB}: {Benchmark} {Tasks} for {Continual} {Graph} {Learning}.
\newblock \emph{Advances in Neural Information Processing Systems}, 35:\penalty0 13006--13021, December 2022{\natexlab{a}}.

\bibitem[Zhang et~al.(2022{\natexlab{b}})Zhang, Song, and Tao]{zhang_sparsified_2022}
Xikun Zhang, Dongjin Song, and Dacheng Tao.
\newblock Sparsified {Subgraph} {Memory} for {Continual} {Graph} {Representation} {Learning}.
\newblock In \emph{2022 {IEEE} {International} {Conference} on {Data} {Mining} ({ICDM})}, pp.\  1335--1340, 2022{\natexlab{b}}.

\bibitem[Zhang et~al.(2024{\natexlab{a}})Zhang, Song, Chen, and Tao]{Zhang_2024}
Xikun Zhang, Dongjin Song, Yixin Chen, and Dacheng Tao.
\newblock Topology-aware {Embedding} {Memory} for {Continual} {Learning} on {Expanding} {Networks}.
\newblock In \emph{Proceedings of the 30th {ACM} {SIGKDD} {Conference} on {Knowledge} {Discovery} and {Data} {Mining}}, pp.\  4326--4337, 2024{\natexlab{a}}.
\newblock \doi{10.1145/3637528.3671732}.

\bibitem[Zhang et~al.(2024{\natexlab{b}})Zhang, Song, and Tao]{zhang_continual_2024}
Xikun Zhang, Dongjin Song, and Dacheng Tao.
\newblock Continual {Learning} on {Graphs}: {Challenges}, {Solutions}, and {Opportunities}, February 2024{\natexlab{b}}.
\newblock arXiv:2402.11565.

\bibitem[Zhang et~al.(2020)Zhang, Cai, Gong, Liu, and Cai]{Zhang_2020_GraphConvolutionalExtreme}
Zijia Zhang, Yaoming Cai, Wenyin Gong, Xiaobo Liu, and Zhihua Cai.
\newblock Graph {Convolutional} {Extreme} {Learning} {Machine}.
\newblock In \emph{2020 {International} {Joint} {Conference} on {Neural} {Networks} ({IJCNN})}, July 2020.
\newblock \doi{10.1109/IJCNN48605.2020.9206649}.

\bibitem[Zhou \& Cao(2021)Zhou and Cao]{zhou_overcoming_2021}
Fan Zhou and Chengtai Cao.
\newblock Overcoming {Catastrophic} {Forgetting} in {Graph} {Neural} {Networks} with {Experience} {Replay}.
\newblock \emph{Proceedings of the AAAI Conference on Artificial Intelligence}, 35\penalty0 (5):\penalty0 4714--4722, May 2021.
\newblock \doi{10.1609/aaai.v35i5.16602}.

\end{thebibliography}
\end{document}